\documentclass[review]{elsarticle}

\graphicspath{ {./figures/} }

\usepackage[utf8]{inputenc}
\usepackage[T1]{fontenc}
\usepackage{hyperref}
\usepackage{url}
\usepackage{booktabs}
\usepackage{amsfonts}
\usepackage{amsmath,amssymb}
\usepackage{graphicx}
\usepackage{float}
\usepackage{verbatim}
\usepackage{subcaption}
\usepackage{array}
\usepackage{siunitx}
\usepackage{stfloats}
\usepackage{longtable}
\usepackage{pdfpages}
\usepackage{multirow}

\restylefloat{figure}
\floatstyle{plaintop}
\restylefloat{table}

\journal{Applied Soft Computing} 

\biboptions{numbers,sort&compress}

\usepackage{placeins} 

\begin{document}
\begin{frontmatter}


\title{A Progressive Design Study of Visual Encoders and Value Estimation for Replay-Free Parallelized Q-Learning}

\author[unipd]{Taha Shieenavaz\corref{cor1}}
\ead{tahashieenavaz@gmail.com}

\author[unipd]{Shabnam Zareshahraki\fnref{presentaddress}}
\ead{shabnamzaresh@gmail.com}

\author[unipd]{Loris Nanni}
\ead{loris.nanni@unipd.it}

\cortext[cor1]{Corresponding author.}
\address[unipd]{Department of Information Engineering, University of Padua, Via Gradenigo 6/B, 35131 Padova, Italy}
\fntext[presentaddress]{Present address: La Manno Lab,
École polytechnique fédérale de Lausanne (EPFL),
Lausanne, Switzerland.}

\begin{abstract}
Replay-free parallelized Q-learning removes the large experience replay
buffers and target networks used by conventional deep Q-learning, but the
role of network architecture in this training regime remains comparatively
underexplored. We investigate this question through a progressive three-phase
study within the Parallelized Q-Network (PQN) framework. First, we compare
eight convolutional encoder topologies on Atari-57 under a common training
protocol while jointly considering performance and computational complexity.
Second, we integrate Hadamax-style multiplicative feature interactions and
explicit pooling into the selected encoder hierarchy. Third, with the visual
representation fixed, we compare complete categorical-dueling,
ensemble-dueling, and categorical ensemble-dueling value-estimation
configurations. The resulting architecture, Aftab, achieves an interquartile
mean human-normalized score of $6.592$ on Atari-57, compared with $2.715$ for
our independently rerun PQN reference, with a game-level Probability of
Improvement of $0.86$. After completing all architecture selection on
Atari-57, we evaluate Aftab on Procgen Hard. Aftab achieves a terminal IQM
normalized score of $0.418$ compared with $0.382$ for PQN and increases the
normalized area under the learning curve from $0.216$ to $0.541$, although
terminal performance remains heterogeneous across environments. These results
show that visual topology, multiplicative representation, and downstream
value-estimation design can substantially affect replay-free Q-learning, and
that their benefits should be evaluated jointly with computational complexity.
\end{abstract}

\begin{keyword}
Deep reinforcement learning \sep
Q-learning \sep
Visual encoders \sep
Parallelized Q-Network \sep
Hadamax \sep
Categorical regression \sep
Ensemble learning
\end{keyword}

\end{frontmatter}

\section{Introduction}
\label{sec:introduction}

Deep reinforcement learning (DRL) enables agents to learn directly from
high-dimensional visual observations
\citep{mnih2013dqn,mnih2015nature}. In value-based methods, these observations
are typically processed by a convolutional neural network (CNN) whose
representation is used to estimate action values
\citep{lecun1998gradient}. The structure of this visual encoder therefore
directly affects the information available to the value function.

Visual representation learning has evolved considerably since the original
Deep Q-Network (DQN), including deeper residual networks
\citep{He2016resnet}, neural architecture search
\citep{Zoph2017NASNet}, and Vision Transformers
\citep{dosovitskiy2021images}. Reinforcement-learning systems such as IMPALA
\citep{Espeholt2018impala} have likewise benefited from deeper visual
representations. Nevertheless, many influential value-based DRL methods retain
variants of the compact convolutional backbone introduced with DQN. This
motivates a basic architectural question: how much can replay-free value
learning benefit from changes to visual topology without simply increasing
model size or altering the underlying learning algorithm?

This question is particularly relevant to the Parallelized Q-Network (PQN)
\citep{Gallici2025pqn}. PQN removes the large historical replay buffer and
target network used by conventional DQN and instead learns from recently
collected synchronous trajectories across many vectorized environments.
Layer Normalization and multi-step $\lambda$-returns support stable learning
under this replay-free regime, providing a useful setting in which to study
architectural choices without many of the stabilization mechanisms used by
conventional DQN variants.

Hadamax \citep{Kooietal2025hadamax} demonstrates that the visual
representation can have a substantial effect in this setting. It replaces
strided-convolution downsampling with explicit max-pooling and introduces
multiplicative Hadamard interactions between parallel GELU-activated feature
transformations. Under the reported PQN protocol, Hadamax substantially
improves Atari-57 performance without modifying the underlying algorithmic
hyperparameters. However, straightforward five- and seven-layer extensions do
not outperform the original three-layer Hadamax architecture under its
40M-frame ablation protocol. Effective encoder scaling may therefore depend
not only on depth, but also on how convolutional topology, spatial reduction,
and multiplicative interactions are organized.

Motivated by this observation, we study architectural design in PQN through a
progressive three-phase evaluation. The design is intentionally not fully
factorial: each primary Atari configuration is trained on all 57 games for
200 million observed frames using four independent random seeds, making
exhaustive evaluation of every combination computationally prohibitive.
Instead, one representation is selected at each stage and held fixed in the
next. Phase~1 compares eight CNN topologies while retaining the scalar PQN
value head. Phase~2 integrates Hadamax-style multiplicative interactions and
explicit pooling into the selected topology and compares two spatial-reduction
strategies. Phase~3 fixes the resulting visual representation and compares
categorical-dueling, ensemble-dueling, and categorical ensemble-dueling
value-estimation configurations. Because these configurations combine multiple
mechanisms, Phase~3 is interpreted as a comparison of complete architectures
rather than a factorial isolation of their individual components. The final
configuration selected through this process is called \emph{Aftab}.

On Atari-57, Aftab achieves an interquartile mean (IQM)
human-normalized score (HNS) of 6.592, compared with 2.715 for our
independently rerun PQN reference, with a game-level Probability of
Improvement (PoI) of 0.86. After completing all architecture selection on
Atari-57, we evaluate Aftab on Procgen Hard
\citep{cobbe2019procgen}. Aftab obtains a terminal IQM Procgen Normalized
Score of 0.418 versus 0.382 for PQN, while normalized area under the learning
curve increases from 0.216 to 0.541. Terminal performance remains
heterogeneous: PQN retains the higher median and the higher final raw score in
nine of the sixteen environments. We therefore interpret Procgen primarily as
evidence of stronger aggregate learning behavior over the training trajectory
under procedural variation, rather than as evidence of uniform superiority or
formal out-of-distribution generalization.

Our objective is not to compete directly with substantially different
replay-buffer-based or model-based Atari agents. Instead, we ask how far the
replay-free PQN paradigm can be improved through architectural design and how
the resulting gains relate to computational complexity.

The main contributions are:
\begin{itemize}
    \item \textbf{Visual encoder topology in replay-free Q-learning.}
    We compare eight CNN topologies on the complete Atari-57 benchmark under
    a common PQN training protocol, jointly assessing performance, parameter
    count, and computational cost.

    \item \textbf{Hadamax integration into an independently selected visual hierarchy.}
    We examine how multiplicative interactions and explicit pooling behave
    within the selected topology and quantify the resulting
    performance--complexity trade-off.

    \item \textbf{Comparison of complete value-estimation architectures.}
    With the visual representation fixed, we compare categorical-dueling,
    ensemble-dueling, and categorical ensemble-dueling configurations without
    attributing their effects to individual constituent mechanisms.

    \item \textbf{A reproducible final architecture with secondary cross-benchmark evaluation.}
    The resulting Aftab configuration improves our PQN reference on Atari-57
    and is subsequently evaluated on Procgen Hard, which is not used for
    architecture selection.
\end{itemize}

The name \emph{Aftab}, Persian for ``sunshine,'' follows the naming tradition
of Rainbow \citep{Hessel2018rainbow} and denotes the final configuration
obtained through the progressive design process.

\section{Related Work}
\label{sec:related_work}

\subsection{Visual Encoders in Deep Reinforcement Learning}
\label{sec:rw_encoders}

The Deep Q-Network (DQN) established CNNs as effective visual encoders for
value-based reinforcement learning
\citep{mnih2013dqn,mnih2015nature}. Its compact three-layer backbone was
subsequently retained by influential methods including Double DQN
\citep{VanHasselt2016DoubleDQN}, Dueling DQN
\citep{Wang2016dueling}, C51
\citep{bellemare2017distributional}, and Rainbow
\citep{Hessel2018rainbow}, despite substantial changes to learning,
exploration, and value estimation.

Deeper representations have also proved effective in reinforcement learning.
IMPALA \citep{Espeholt2018impala} employs a deeper residual encoder for
large-scale visual learning, while Bigger, Better, Faster (BBF)
\citep{Schwarzer2023BBF} increases the capacity of residual Atari
representations. More recent work on network scaling, including Simba
\citep{Lee2024Simba}, further emphasizes the interaction between architecture,
optimization, and performance. Because such changes often modify depth,
width, spatial resolution, and parameter count simultaneously, improved
performance cannot generally be attributed to scale alone. We therefore
compare complete encoder topologies under a common training protocol and
report their computational complexity alongside performance.

\subsection{Replay-Free and Parallelized Value Learning}
\label{sec:rw_pqn}

Conventional DQN stabilizes nonlinear temporal-difference learning using a
large experience replay buffer and a periodically updated target network
\citep{mnih2013dqn,mnih2015nature}. Alternative approaches have explored
other stabilization mechanisms; for example, CrossQ
\citep{Bhatt2019CrossQ} investigates normalization-based learning without a
target network in continuous control.

PQN \citep{Gallici2025pqn} removes both the large historical replay buffer and
target network, learning instead from recent synchronous trajectories collected
across vectorized environments. Layer Normalization, multi-step
$\lambda$-returns, and repeated minibatch updates over the current rollout
support this training regime. PQN is therefore replay-free in the sense that
it does not maintain and resample from a large buffer of historical
experience, although the current rollout is retained temporarily for return
computation and optimization.

Hadamax \citep{Kooietal2025hadamax} subsequently showed that modifying the
visual representation can substantially improve performance within this
framework. Its parallel multiplicative transformations and explicit
max-pooling outperform the original PQN encoder under the reported Atari-57
protocol, whereas straightforward five- and seven-layer extensions do not
improve upon the original three-layer Hadamax design in the reported
40M-frame ablation. Our Phase~2 study approaches this question differently:
we first select an alternative convolutional hierarchy within PQN and then
integrate Hadamax-style processing into that hierarchy.

\subsection{Value Estimation and Ensemble Exploration}
\label{sec:rw_value_heads}

Value-based reinforcement learning has also been improved through changes to
the value-estimation architecture. Dueling networks
\citep{Wang2016dueling} separate state value from action-dependent advantage,
whereas C51 \citep{bellemare2017distributional} models a distribution over
future returns using a fixed categorical support and a projected
distributional Bellman update.

A distinct line of work formulates scalar value prediction as classification.
\citet{Farebrother2024StopRegressing} show that scalar targets can be encoded
categorically and optimized using cross-entropy. Their HL-Gauss formulation
represents a scalar target as a smooth distribution over fixed value bins.
Our categorical configurations follow this regression-as-classification
approach: the scalar bootstrapped $\lambda$-return is encoded categorically,
rather than using the Bellman-distribution projection of C51.

Ensemble methods provide a complementary mechanism for value estimation and
exploration. Bootstrapped DQN \citep{Osband2016Bootstrapped} uses multiple
independently parameterized Q-heads trained on bootstrap subsets of experience
to promote temporally extended exploration, while subsequent work has examined
related ensemble and optimistic strategies \citep{Nauman2024BRO}. Our
multi-head configurations are inspired by this structure but differ in their
training and data assignment, as detailed in Section~\ref{sec:prelim_bootstrap}.

Rainbow \citep{Hessel2018rainbow} showed that multiple value-learning
mechanisms can be complementary in replay-buffer-based DQN. By contrast, PQN
and Hadamax primarily address the training regime and visual representation.
We therefore compare categorical-dueling, ensemble-dueling, and categorical
ensemble-dueling architectures while holding the Gamma-Hadamax representation
fixed. These experiments compare complete value-estimation configurations and
do not isolate the causal contribution of categorical regression, dueling
decomposition, or ensembling individually.

\subsection{Procedural Evaluation in Reinforcement Learning}
\label{sec:rw_procgen}

Procgen \citep{cobbe2019procgen} provides 16 visually diverse procedurally
generated tasks in which level structure and appearance vary across episodes,
complementing benchmarks such as Atari-57, where each individual game retains
comparatively stable visual and structural regularities. In this study,
Procgen Hard is used only after all architecture selection on Atari-57 and only
for the final Aftab configuration. We therefore use it as a secondary test of
learning under procedural variation, not as evidence of direct transfer from
Atari or as a formal demonstration of out-of-distribution generalization.

\section{Preliminaries}
\label{sec:preliminaries}

This section defines the components needed for the architectural comparisons:
replay-free PQN training, Hadamax representations, categorical and dueling
value estimation, and multi-head value functions.

\subsection{Replay-Free Parallelized Q-Learning}
\label{sec:prelim_pqn}

In value-based reinforcement learning, a network
$Q_{\theta}(s,a)$ estimates the expected return from taking action $a$ in
state $s$. Classical DQN
\citep{mnih2013dqn,mnih2015nature} uses a large replay buffer and a
periodically updated target network to stabilize learning. Throughout this
paper, \emph{Nature DQN} refers to the 2015 DQN formulation
\citep{mnih2015nature} and its standard three-layer convolutional encoder.

PQN \citep{Gallici2025pqn} instead collects short trajectories synchronously
from many vectorized environments and optimizes directly on the resulting
recent batch. It combines this sampling regime with Layer Normalization
\citep{ba2016layernorm}, multi-step $\lambda$-returns, and repeated minibatch
updates over the current rollout.

The recursive $\lambda$-return is

\begin{equation}
R_t^{\lambda}
=
\begin{cases}
r_t, & \text{for a terminal transition}, \\[2mm]
r_t
+
\gamma
\left[
\lambda R_{t+1}^{\lambda}
+
(1-\lambda)
\max_{a'}
Q_{\theta}(s_{t+1},a')
\right],
& \text{otherwise}.
\end{cases}
\label{eq:lambda_return}
\end{equation}

The parameter $\lambda$ interpolates between one-step temporal-difference
bootstrapping and longer-horizon returns. PQN is replay-free in the sense that
it does not maintain and repeatedly sample from a large historical replay
buffer; the current rollout is nevertheless retained temporarily to compute
$\lambda$-returns and perform minibatch optimization. Our experiments preserve
this training structure while modifying the visual representation and
downstream value-estimation architecture.

\begin{figure*}[ht]
    \centering

    \begin{subfigure}[t]{0.43\textwidth}
        \centering
        \includegraphics[
            height=0.27\textheight,
            keepaspectratio
        ]{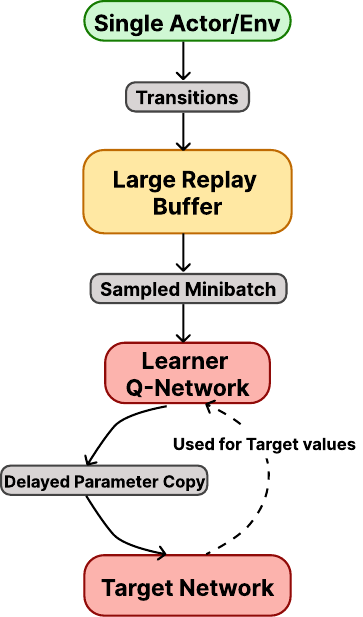}
        \caption{Conventional DQN training.}
        \label{fig:concept_dqn}
    \end{subfigure}
    \hfill
    \begin{subfigure}[t]{0.43\textwidth}
        \centering
        \includegraphics[
            height=0.27\textheight,
            keepaspectratio
        ]{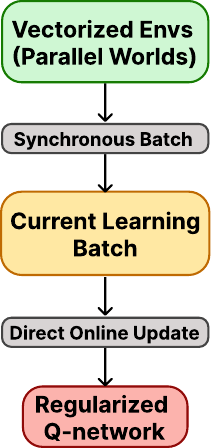}
        \caption{Replay-free parallelized PQN training.}
        \label{fig:concept_pqn}
    \end{subfigure}

    \caption{\textbf{Conceptual comparison of DQN and PQN training.}
    Conventional DQN samples transitions from a large historical replay
    buffer and uses a delayed target network. PQN instead learns from the
    current synchronous batch collected by vectorized environments and does
    not maintain a large historical replay buffer or a separate target
    network.}
    \label{fig:dqn_pqn_training}
\end{figure*}

Figure~\ref{fig:dqn_pqn_training} contrasts the conventional DQN training
loop with the replay-free parallelized PQN regime used throughout this work.

\subsection{Hadamax Representation}
\label{sec:prelim_hadamax}

Hadamax (Hadamard max-pooling) \citep{Kooietal2025hadamax} modifies the PQN
encoder through multiplicative feature interactions and explicit spatial
downsampling. Each Hadamax block applies two parallel stride-one
convolutions, normalizes and activates their outputs, combines them by an
element-wise Hadamard product, and then applies max-pooling.

For input representation $z^{j-1}$, block $j$ is

\begin{equation}
z^{j}
=
\operatorname{MP}
\left(
f\left[
\operatorname{LN}
\left(
z^{j-1}W_{1}^{j-1}
\right)
\right]
\odot
f\left[
\operatorname{LN}
\left(
z^{j-1}W_{2}^{j-1}
\right)
\right]
\right),
\label{eq:hadamax}
\end{equation}

where $W_{1}^{j-1}$ and $W_{2}^{j-1}$ are independently learned convolutional
kernels, $\operatorname{LN}$ denotes Layer Normalization,
$\operatorname{MP}$ denotes max-pooling, $\odot$ is the element-wise Hadamard
product, and $f$ is the GELU activation
\citep{hendrycks2016gelu}. Thus, a Hadamax block changes both feature
interaction and spatial reduction relative to the original PQN encoder:
strided downsampling is replaced by explicit max-pooling and ReLU by GELU.

\begin{figure}[ht]
    \centering

    \begin{subfigure}[b]{0.32\linewidth}
        \centering
        \includegraphics[
            width=\linewidth,
            keepaspectratio
        ]{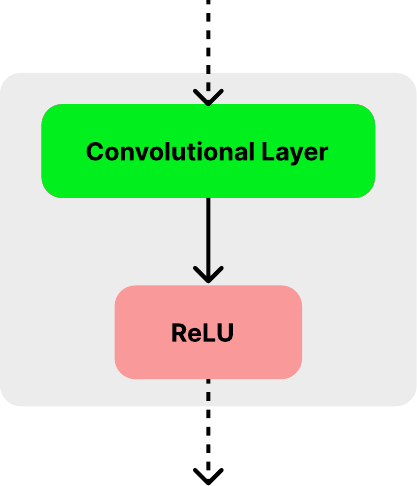}
        \caption{Nature DQN}
        \label{fig:block_dqn}
    \end{subfigure}
    \hfill
    \begin{subfigure}[b]{0.32\linewidth}
        \centering
        \includegraphics[
            width=\linewidth,
            keepaspectratio
        ]{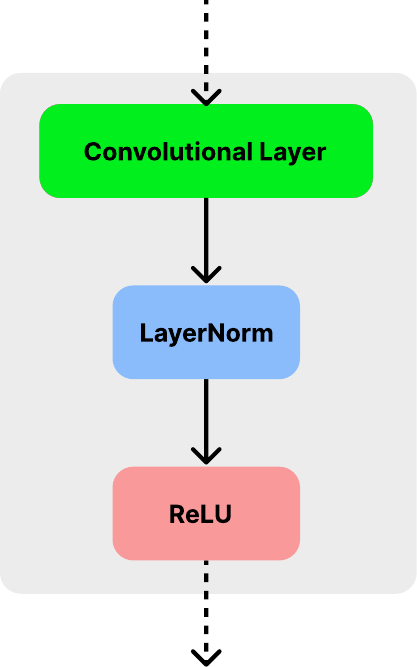}
        \caption{PQN}
        \label{fig:block_pqn}
    \end{subfigure}
    \hfill
    \begin{subfigure}[b]{0.32\linewidth}
        \centering
        \includegraphics[
            width=\linewidth,
            keepaspectratio
        ]{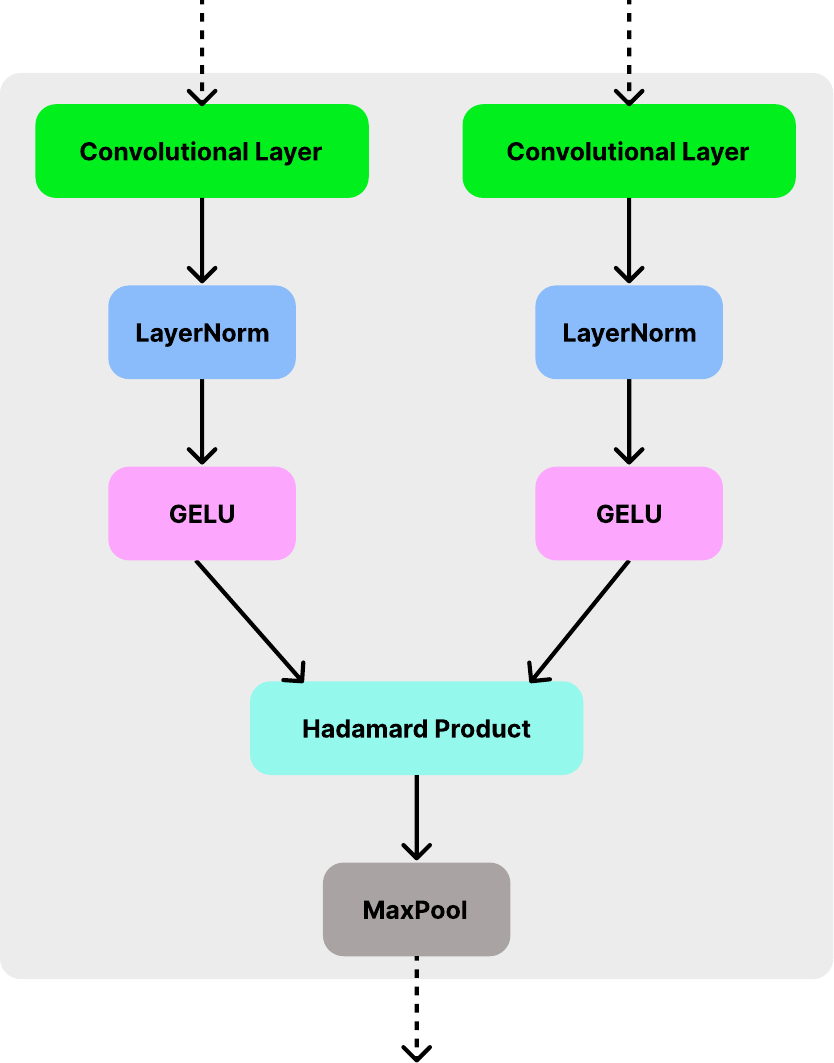}
        \caption{Hadamax}
        \label{fig:block_hadamax}
    \end{subfigure}

    \caption{\textbf{Convolutional processing blocks.}
    Nature DQN uses convolution followed by ReLU; PQN adds Layer
    Normalization; Hadamax combines two LayerNorm--GELU convolutional paths
    by a Hadamard product before explicit max-pooling.}
    \label{fig:blocks_comparison}
\end{figure}

Figure~\ref{fig:blocks_comparison} summarizes the three convolutional
processing blocks.

\subsection{Dueling and Categorical Value Estimation}
\label{sec:prelim_value_estimation}

\subsubsection{Dueling Architecture}
\label{sec:prelim_dueling}

The dueling architecture \citep{Wang2016dueling} decomposes the action-value
function into a state-value term $V(s)$ and an action-dependent advantage
$A(s,a)$:

\begin{equation}
Q(s,a)
=
V(s)
+
A(s,a)
-
\frac{1}{|\mathcal{A}|}
\sum_{a'\in\mathcal{A}} A(s,a').
\label{eq:dueling}
\end{equation}

\begin{figure}[ht]
    \centering
    \includegraphics[
        width=0.90\linewidth,
        keepaspectratio
    ]{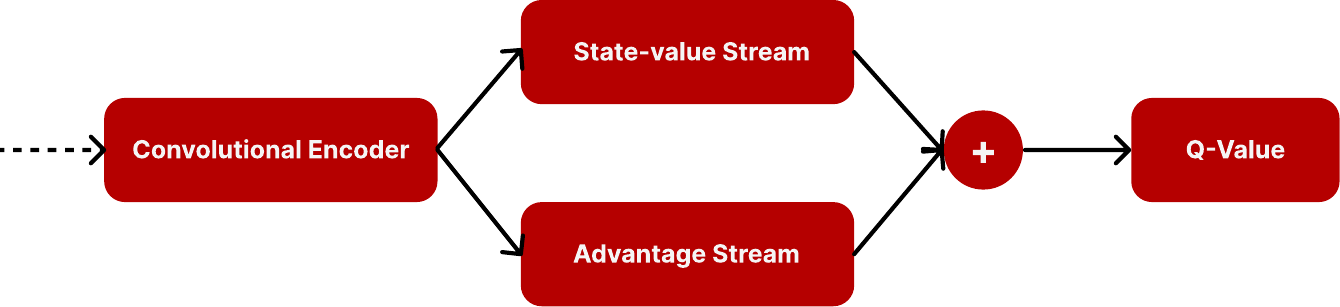}

    \caption{\textbf{Dueling value architecture.}
    The shared visual representation is separated into a state-value stream
    and an action-dependent advantage stream before the two components are
    combined to obtain action values.}
    \label{fig:dueling_architecture}
\end{figure}

As illustrated in Fig.~\ref{fig:dueling_architecture}, mean-centering the
advantage resolves the identifiability ambiguity between the two streams.
This decomposition is used in the Phase~3 value-estimation architectures
while the visual representation is held fixed.

\subsubsection{Categorical Value Estimation}
\label{sec:prelim_distributional}

C51 \citep{bellemare2017distributional} models the distribution of future
returns and applies a distributional Bellman update followed by projection
onto a fixed categorical support. Our categorical configurations instead
follow the regression-as-classification formulation of
\citet{Farebrother2024StopRegressing}. The scalar bootstrapped
$\lambda$-return target $y$ is represented as a smooth categorical target
using Histogram Loss with Gaussian targets (HL-Gauss).

For support point $z_i$ and bin width $\Delta$, the target mass is

\begin{equation}
p_i^*(y)
=
\int_{z_i-\Delta/2}^{z_i+\Delta/2}
\frac{1}{\sigma\sqrt{2\pi}}
\exp
\left(
-\frac{(x-y)^2}{2\sigma^2}
\right)
\,dx,
\label{eq:hl_gauss}
\end{equation}

where $\sigma$ controls target smoothing. Predicted categorical probabilities
$p_i(s,a)$ are optimized by cross-entropy,

\begin{equation}
\mathcal{L}_{\mathrm{CE}}
=
-
\sum_i p_i^*(y)\log p_i(s,a).
\label{eq:categorical_loss}
\end{equation}

Unlike C51, this objective does not estimate a Bellman-updated distribution
over future returns. It provides a categorical representation of the scalar
bootstrapped $\lambda$-return. We therefore use the terms
\emph{categorical value estimation} and \emph{categorical regression}, rather than distributional reinforcement learning in the strict C51 sense.

\begin{figure}[ht]
    \centering
    \includegraphics[
        width=0.78\linewidth,
        keepaspectratio
    ]{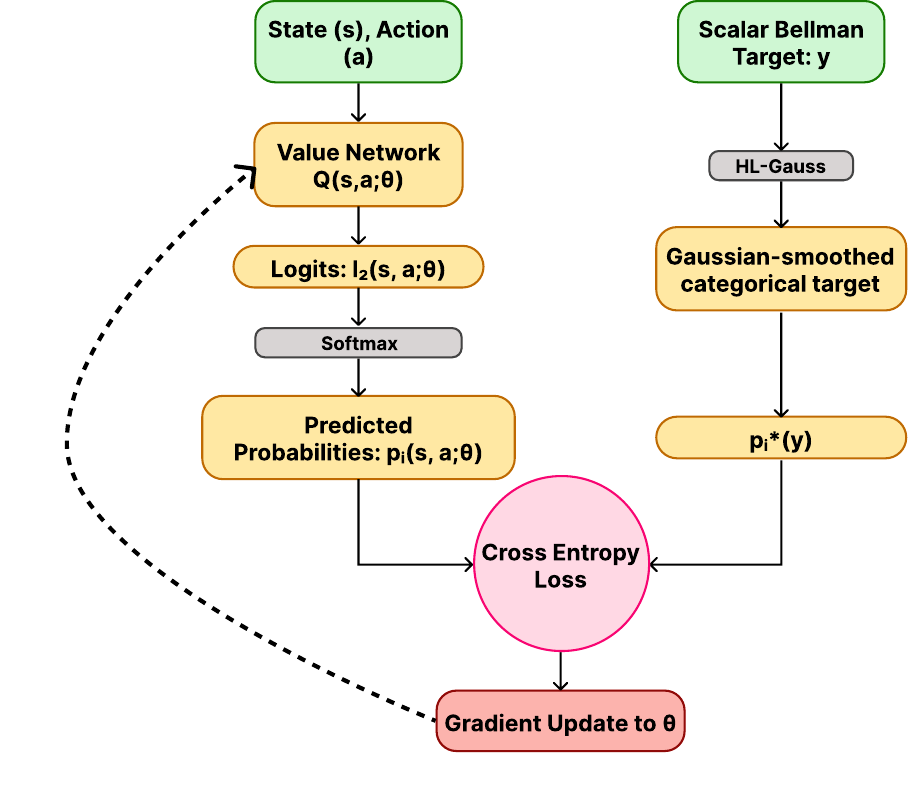}

    \caption{\textbf{Categorical regression with HL-Gauss targets.}
    A scalar bootstrapped target is converted into a Gaussian-smoothed
    categorical target on a fixed support. Cross-entropy is minimized against
    the predicted categorical probabilities, while the categorical prediction
    can subsequently be decoded to a scalar value for behavior and
    bootstrapping.}
    \label{fig:categorical_regression_workflow}
\end{figure}

Figure~\ref{fig:categorical_regression_workflow} summarizes the
regression-as-classification procedure used by the categorical
configurations.

\subsection{Multi-Head Value Functions and Exploration}
\label{sec:prelim_bootstrap}

Bootstrapped DQN \citep{Osband2016Bootstrapped} uses multiple independently
initialized Q-heads, conventionally trained on different bootstrap subsets of
experience, with one head controlling behavior for an extended period to
encourage temporally coherent exploration.

Our ensemble retains the shared-encoder, multi-head structure but differs
from classical Bootstrapped DQN: all $K$ heads receive the same synchronous
PQN training batch, corresponding to bootstrap inclusion probability $p=1$.
The method should therefore not be interpreted as a classical bootstrap
estimator or as posterior sampling. The exact parameter sharing, loss
aggregation, and action-selection procedures are specified in
Section~\ref{sec:methodology}.
\section{Systematic Design Study}
\label{sec:methodology}

We investigate architectural design in replay-free parallelized Q-learning
through a progressive three-phase study. Phase~1 compares alternative
convolutional encoder topologies while retaining the scalar PQN value head.
Phase~2 integrates Hadamax-style processing into the encoder selected in
Phase~1. Phase~3 fixes the resulting visual representation and compares
alternative downstream value-estimation architectures.

The study is progressive rather than fully factorial. Each primary Atari
configuration is evaluated on all 57 games for 200 million observed frames
using four independent training seeds. Exhaustively evaluating every
combination of encoder topology, Hadamax representation, categorical
regression, dueling decomposition, and ensemble structure would therefore be
computationally prohibitive. Instead, one performance--complexity
configuration is selected at each stage and carried forward. Optimization
settings not intrinsic to the evaluated architecture are held fixed within
each phase.

Phase~3 changes both the value architecture and, for categorical
configurations, the training objective: scalar configurations regress the
bootstrapped $\lambda$-return, whereas categorical configurations use
HL-Gauss cross-entropy. Phase~3 is therefore interpreted as a comparison of
complete implemented configurations rather than as a factorial decomposition
of their constituent mechanisms.

\paragraph{Implementation and reference configurations}
All experiments are implemented in PyTorch \citep{pytorch2019}. The scalar
reference follows the PQN training structure described in
Section~\ref{sec:prelim_pqn}, using the Nature-DQN encoder and a normalized
MLP value head. The Hadamax reference is an independent PyTorch
reimplementation of the three-stage architecture of
\citet{Kooietal2025hadamax}. It preserves the multiplicative feature
transformation, GELU activation, and explicit max-pooling design, but uses the
common $4\times84\times84$ Atari input employed throughout this study and
explicit symmetric PyTorch padding rather than JAX-style \nolinkurl{SAME}
padding. Published PQN and Hadamax results are therefore treated only as
external reference points; model selection and statistical analysis use our
independently generated runs. Complete code-level configuration mappings are
provided in the Supplementary Material.

Before entering the encoder, unsigned 8-bit image observations are converted
to floating point and divided by $255$.

\subsection{Phase 1: Convolutional Encoder Design}
\label{sec:method_phase1}

Phase~1 compares eight CNN variants, Alpha through Theta, with the Nature-DQN
encoder used by the PQN reference. The candidates vary in depth, kernel size,
channel progression, spatial downsampling, and final feature-map size; they
were designed to explore alternative visual hierarchies rather than simply
increase parameter count.

For Atari, each encoder receives a $4\times84\times84$ input. Conventional
convolutional stages follow the PQN block in
Fig.~\ref{fig:block_pqn}: convolution, channel-wise
\nolinkurl{LayerNorm2d}, and ReLU. The custom normalization operates over the
channel dimension independently at each spatial location, uses learned
per-channel affine parameters, and sets
$\epsilon_{\mathrm{LN}}=10^{-6}$. Complete layer-by-layer definitions of the
nine encoders are provided in the Supplementary Material.

Because encoder parameter count alone does not characterize the complete
Q-network, we additionally track the final feature-map size, flattened
dimension, and convolutional multiply-accumulate operations (MACs).
Table~\ref{tab:phase1_complexity} summarizes these quantities.

\begin{table*}[ht]
\centering
\caption{\textbf{Phase~1 encoder complexity for a
$4\times84\times84$ input.} MAC counts include convolutional
multiply-accumulates and exclude normalization, bias addition, activations,
and other element-wise operations.}
\label{tab:phase1_complexity}
\resizebox{\textwidth}{!}{%
\begin{tabular}{l r r r r}
\toprule
\textbf{Variant}
& \textbf{Final feature map}
& \textbf{Flattened dim. $F$}
& \textbf{Encoder params}
& \textbf{Encoder MACs (M)} \\
\midrule
PQN     & $64\times7\times7$    & 3,136  & 78,304  & 7.737 \\
Alpha   & $64\times7\times7$    & 3,136  & 174,752 & 27.542 \\
Beta    & $16\times14\times14$  & 3,136  & 89,008  & 61.516 \\
Gamma   & $64\times7\times7$    & 3,136  & 117,168 & 22.902 \\
Delta   & $96\times6\times6$    & 3,456  & 78,552  & 6.143 \\
Epsilon & $64\times8\times8$    & 4,096  & 80,112  & 13.253 \\
Zeta    & $48\times10\times10$  & 4,800  & 77,232  & 25.362 \\
Eta     & $128\times19\times19$ & 46,208 & 78,400  & 28.422 \\
Theta   & $32\times8\times8$    & 2,048  & 76,288  & 9.066 \\
\bottomrule
\end{tabular}%
}
\end{table*}

All Phase~1 configurations use the same scalar value head,
\[
F
\rightarrow
\mathrm{Linear}(F,512)
\rightarrow
\mathrm{LayerNorm}
\rightarrow
\mathrm{ReLU}
\rightarrow
\mathrm{Linear}(512,A),
\]
where $A$ is the environment-specific action count. Its parameter and
linear-MAC counts are
\begin{equation}
P_{\mathrm{scalar}}(F,A)
=
512F+1536+513A,
\qquad
M_{\mathrm{scalar}}(F,A)
=
512F+512A.
\label{eq:scalar_head_complexity}
\end{equation}
Thus, architectures with similar convolutional parameter counts can differ
substantially in complete-model size when their flattened representations
differ.

\subsubsection{Backbone Selection}
\label{sec:method_backbone_selection}

Phase~1 selects a performance--complexity configuration rather than
automatically carrying forward the largest aggregate point estimate. Atari-57
IQM HNS is considered jointly with the paired statistical analysis, encoder
complexity, flattened feature dimension, and complete-model cost.

A non-significant pairwise test is not treated as evidence of equivalence.
When corrected testing does not resolve the ordering of leading configurations
with similar aggregate performance, lower model complexity is used as a
pragmatic secondary criterion. This is a performance--complexity decision
rule, not an equivalence or non-inferiority procedure.

As a sensitivity analysis, the same selection rule is applied to 10 randomly
formed subsets of 15 Atari games using the already completed training runs.
This assesses sensitivity to task composition and is not an independent
validation experiment. Gamma is selected and fixed before Phase~2.

\subsection{Phase 2: Hadamax Integration}
\label{sec:method_phase2}

Phase~2 tests whether the Gamma hierarchy selected in Phase~1 provides an
effective structure for Hadamax processing. This differs from simply making
the original Hadamax encoder deeper: the original study found that
straightforward five- and seven-layer extensions did not improve on its
three-stage architecture under the reported 40-million-frame Atari-57
ablation protocol \citep{Kooietal2025hadamax}.

Our \nolinkurl{HadamaxBlock} implements the two branches using one bias-free
convolution with $2C_{\mathrm{out}}$ output channels, which is split into two
independently parameterized feature banks. Each branch is normalized over
its $C_{\mathrm{out}}$ channels at every spatial location, activated with
GELU, multiplied element-wise with the other branch, and then spatially
reduced by max-pooling, as illustrated in Fig.~\ref{fig:block_hadamax}.

We compare the three-stage Hadamax reference with two five-stage variants
derived from Gamma. Both replace Gamma's strided convolutions with stride-one
Hadamax blocks followed by explicit pooling. They differ only in the pooling
padding of Blocks~3 and~5. Gamma-Hadamax-Valid uses unpadded $3\times3$
pooling and produces the spatial sequence
$84\!\rightarrow\!42\!\rightarrow\!21\!\rightarrow\!19
\!\rightarrow\!9\!\rightarrow\!7$, whereas Gamma-Hadamax-Same uses one
pixel of padding in those blocks and produces
$84\!\rightarrow\!42\!\rightarrow\!21\!\rightarrow\!21
\!\rightarrow\!10\!\rightarrow\!10$.
Complete block specifications are provided in the Supplementary Material.

Figure~\ref{fig:gamma_hadamax_architecture} visualizes the structural change
from the Gamma encoder selected in Phase~1 to the
Gamma-Hadamax-Valid representation carried forward from Phase~2.

\begin{figure*}[ht]
    \centering

    \begin{subfigure}[t]{0.40\textwidth}
        \centering
        \includegraphics[
            height=0.34\textheight,
            keepaspectratio
        ]{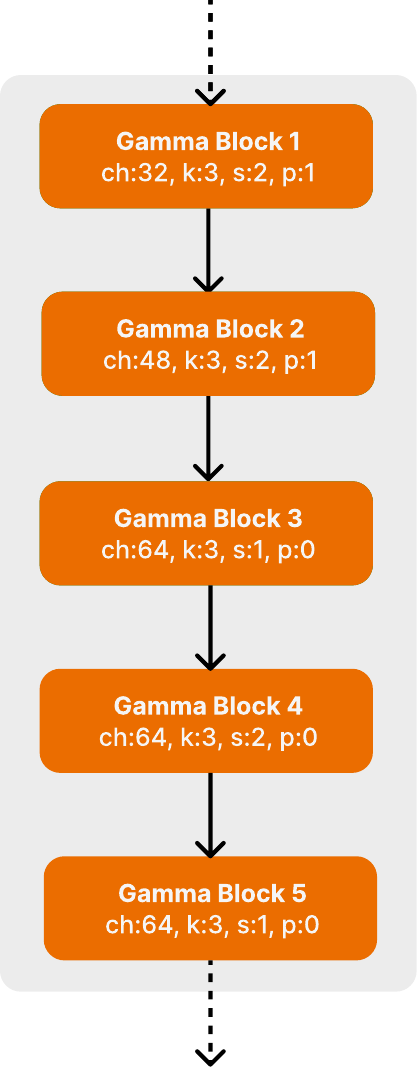}
        \caption{Gamma encoder selected in Phase~1.}
        \label{fig:gamma_encoder_diagram}
    \end{subfigure}
    \hfill
    \begin{subfigure}[t]{0.40\textwidth}
        \centering
        \includegraphics[
            height=0.34\textheight,
            keepaspectratio
        ]{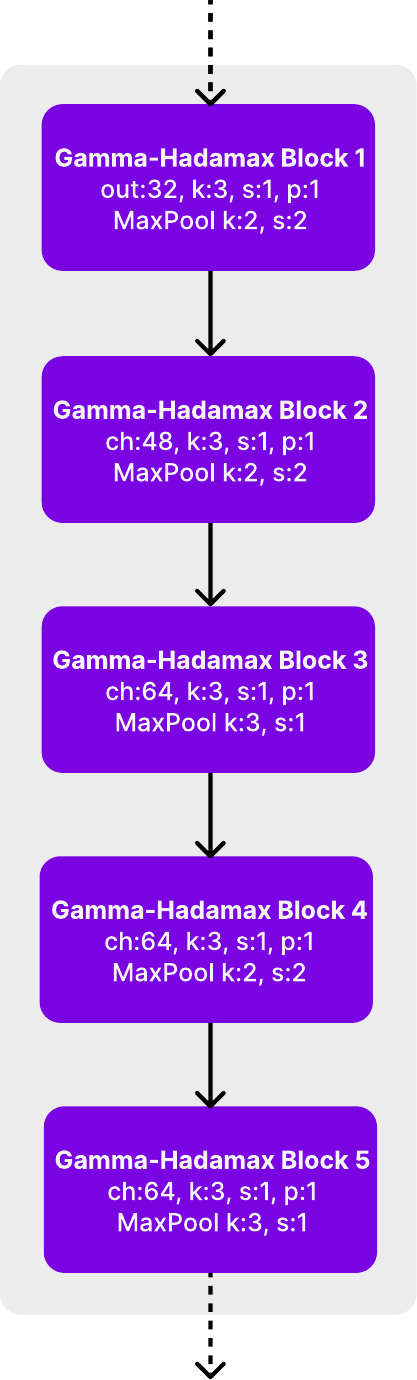}
        \caption{Gamma-Hadamax-Valid.}
        \label{fig:gamma_hadamax_valid_diagram}
    \end{subfigure}

    \caption{\textbf{Phase~2 integration of Hadamax processing into the
    Gamma hierarchy.}
    Gamma uses five conventional convolutional stages.
    Gamma-Hadamax-Valid retains the five-stage channel hierarchy while
    replacing the conventional stages with Hadamax processing and explicit
    max-pooling. Unpadded $3\times3$ pooling in Blocks~3 and~5 produces the
    final $7\times7$ spatial representation carried forward to Phase~3.}
    \label{fig:gamma_hadamax_architecture}
\end{figure*}

\begin{table*}[ht]
\centering
\caption{\textbf{Phase~2 encoder complexity for a
$4\times84\times84$ input.} Hadamax convolutions are bias-free.
MAC counts include convolutional multiply-accumulates only.}
\label{tab:phase2_complexity}
\resizebox{\textwidth}{!}{%
\begin{tabular}{l r r r r}
\toprule
\textbf{Variant}
& \textbf{Final feature map}
& \textbf{$F$}
& \textbf{Encoder params}
& \textbf{Encoder MACs (M)} \\
\midrule
Hadamax Reference
& $64\times11\times11$
& 7,744
& 156,288
& 159.015 \\

Gamma-Hadamax-Valid
& $64\times7\times7$
& 3,136
& 233,792
& 122.001 \\

Gamma-Hadamax-Same
& $64\times10\times10$
& 6,400
& 233,792
& 129.300 \\
\bottomrule
\end{tabular}%
}
\end{table*}

The two Gamma-Hadamax variants have identical encoder parameter counts, but
Gamma-Hadamax-Same produces more than twice the flattened representation of
Gamma-Hadamax-Valid ($6400$ versus $3136$), substantially enlarging the
downstream value head. Selection follows the same performance--complexity rule
as in Phase~1: when corrected testing does not resolve the ordering of the
leading variants, the lower-complexity configuration is preferred.
Gamma-Hadamax-Valid is therefore carried forward as the fixed representation
for Phase~3.

\subsection{Phase 3: Value-Estimation Architectures}
\label{sec:method_phase3}

Phase~3 fixes the Gamma-Hadamax-Valid encoder and compares three downstream
architectures with its scalar value head: Categorical Dueling, Ensemble
Dueling, and the combined categorical ensemble-dueling configuration Aftab.
All downstream streams use a 512-dimensional hidden representation with
Layer Normalization. Categorical Dueling and Ensemble Dueling use ReLU,
whereas Aftab uses GELU. The experiment therefore compares complete
implemented architectures rather than individual components.

\subsubsection{Categorical Dueling}
\label{sec:method_categorical_dueling}

Categorical Dueling combines the dueling decomposition
\citep{Wang2016dueling} with HL-Gauss categorical regression of the scalar
bootstrapped $\lambda$-return. Its value and advantage streams are
\[
F\rightarrow512\rightarrow\mathrm{LayerNorm}
\rightarrow\mathrm{ReLU}\rightarrow51
\]
and
\[
F\rightarrow512\rightarrow\mathrm{LayerNorm}
\rightarrow\mathrm{ReLU}\rightarrow A\times51,
\]
respectively. Value and advantage logits are combined as
\begin{equation}
\ell_i(s,a)
=
\ell_i^V(s)
+
\ell_i^A(s,a)
-
\frac{1}{A}
\sum_{a'=1}^{A}\ell_i^A(s,a').
\label{eq:categorical_dueling}
\end{equation}

The categorical support contains 51 bins over $[-10,10]$. HL-Gauss uses a
sigma ratio of $0.75$, corresponding to $\sigma\approx0.2941$ under the
implemented support discretization. Range clamping is enabled during target
construction, while the framework's optional categorical value-change
clipping is disabled
(\nolinkurl{distributional\_value\_clip}=0).
For each transition, the scalar $\lambda$-return is converted to an HL-Gauss
target and optimized by cross-entropy for the action taken. Categorical
predictions are decoded to scalar action values for behavior and
bootstrapping. As emphasized in Section~\ref{sec:prelim_distributional},
this is categorical regression of a scalar target, not C51-style
distributional Bellman projection.

\subsubsection{Ensemble Dueling}
\label{sec:method_ensemble_dueling}

Ensemble Dueling uses $K=10$ independently parameterized scalar dueling heads
above the shared Gamma-Hadamax-Valid encoder. Each head has a value stream
\[
F\rightarrow512\rightarrow\mathrm{LayerNorm}
\rightarrow\mathrm{ReLU}\rightarrow1
\]
and an advantage stream
\[
F\rightarrow512\rightarrow\mathrm{LayerNorm}
\rightarrow\mathrm{ReLU}\rightarrow A.
\]

At the beginning of a training trajectory, one head is sampled uniformly for
each parallel environment. Actions are selected $\epsilon$-greedily according
to that head until termination or truncation, after which a new head is
sampled. This produces temporally extended head-conditioned behavior.

Unlike classical Bootstrapped DQN \citep{Osband2016Bootstrapped}, the
bootstrap inclusion probability is $p=1$; every head is therefore optimized
on every transition in the synchronous PQN batch. Each head nevertheless
constructs its own $\lambda$-return using its own next-state estimate. For a
mini-batch of size $N$, the loss is
\begin{equation}
\mathcal{L}_{\mathrm{ens}}
=
\frac{1}{NK}
\sum_{j=1}^{N}
\sum_{k=1}^{K}
\frac{1}{2}
\left[
Q_k(s_j,a_j)-R_{j,k}^{\lambda}
\right]^2.
\label{eq:ensemble_loss}
\end{equation}
Because $p=1$, the ensemble is not interpreted as a classical bootstrap
estimator or as posterior sampling.

At evaluation, each head independently selects its greedy action and the final
action is determined by majority vote across the $K$ heads.

\subsubsection{Aftab}
\label{sec:method_aftab}

Aftab combines the $K=10$ ensemble with categorical dueling value estimation.
Each head uses independent value and advantage streams
\[
F\rightarrow512\rightarrow\mathrm{LayerNorm}
\rightarrow\mathrm{GELU}\rightarrow51
\]
and
\[
F\rightarrow512\rightarrow\mathrm{LayerNorm}
\rightarrow\mathrm{GELU}\rightarrow A\times51.
\]

Within each head, value and advantage logits are combined according to
Eq.~\ref{eq:categorical_dueling} and decoded to scalar Q-values for behavior
and bootstrapping. Head-conditioned exploration and majority-vote evaluation
follow Ensemble Dueling. Because $p=1$, all heads receive every transition,
but each constructs its own scalar $\lambda$-return before conversion to an
HL-Gauss target.

Aftab therefore combines categorical regression, dueling decomposition,
multi-head estimation, head-conditioned exploration, and GELU-activated head
streams. Its performance is interpreted as that of the complete configuration;
the experiment does not isolate the causal contribution of each component.

\begin{figure*}[ht]
    \centering
    \includegraphics[
        width=\textwidth,
        keepaspectratio
    ]{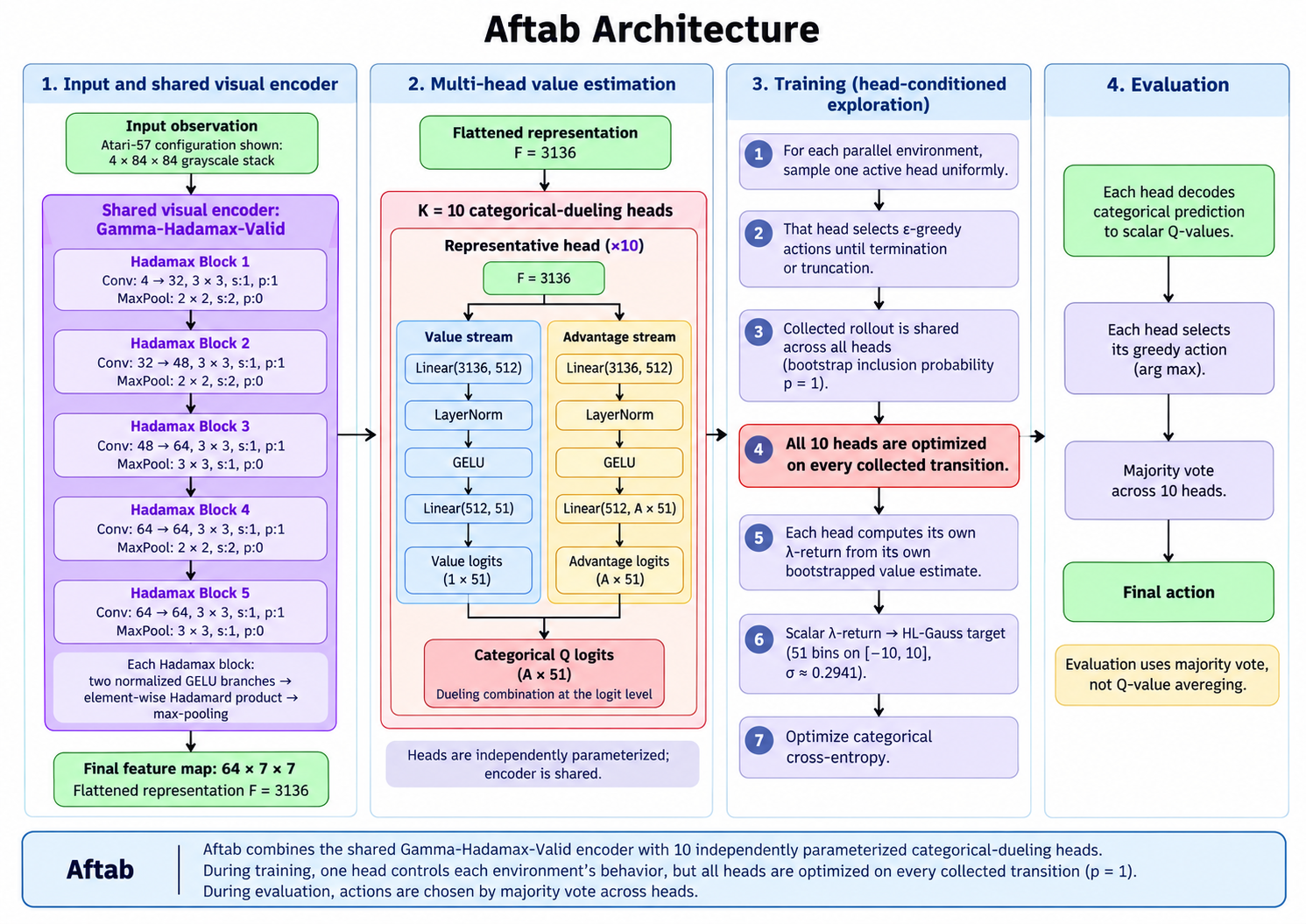}

    \caption{\textbf{Overview of the final Aftab architecture.}
    Aftab uses the shared Gamma-Hadamax-Valid visual encoder followed by
    $K=10$ independently parameterized categorical-dueling heads. During
    training, one head controls the behavior of each parallel environment for
    the duration of a trajectory, while all heads are optimized on every
    collected transition because the bootstrap inclusion probability is
    $p=1$. Each head constructs its own bootstrapped $\lambda$-return and
    converts it into an HL-Gauss categorical target. During evaluation, each
    head independently selects a greedy action and the final action is
    determined by majority vote.}
    \label{fig:aftab_architecture}
\end{figure*}

Figure~\ref{fig:aftab_architecture} summarizes the complete Aftab
architecture and its training and evaluation behavior.

\subsubsection{Phase 3 Model Complexity}
\label{sec:method_phase3_complexity}

The shared Gamma-Hadamax-Valid encoder has $F=3136$, $233{,}792$ trainable
parameters, and $122.001$ million convolutional MACs. Because Atari action
counts vary across environments, Table~\ref{tab:phase3_complexity} reports
complete-model parameter and forward-MAC counts as functions of the
environment-specific action count $A$.

\begin{table*}[ht]
\centering
\caption{\textbf{Phase~3 model complexity.} Counts include the shared
Gamma-Hadamax-Valid encoder. MACs include convolutional and linear
multiply-accumulates but exclude normalization, activations, Hadamard
products, pooling, softmax, and loss computation.}
\label{tab:phase3_complexity}
\resizebox{\textwidth}{!}{%
\begin{tabular}{l l l}
\toprule
\textbf{Configuration}
& \textbf{Trainable parameters}
& \textbf{Forward MACs (M)} \\
\midrule
Scalar Gamma-Hadamax-Valid
& $1{,}840{,}960 + 513A$
& $123.607040 + 0.000512A$ \\

Categorical Dueling
& $3{,}474{,}291 + 26{,}163A$
& $125.238784 + 0.026112A$ \\

Ensemble Dueling ($K=10$)
& $32{,}382{,}282 + 5{,}130A$
& $154.119168 + 0.005120A$ \\

Aftab ($K=10$)
& $32{,}638{,}782 + 261{,}630A$
& $154.375168 + 0.261120A$ \\
\bottomrule
\end{tabular}%
}
\end{table*}

For a representative eight-action Atari environment, the scalar,
Categorical-Dueling, Ensemble-Dueling, and Aftab models contain approximately
$1.845$, $3.684$, $32.423$, and $34.732$ million parameters, respectively,
with approximately $123.611$, $125.448$, $154.160$, and $156.464$ million
counted forward MACs. The ensemble heads therefore impose a much larger
relative cost in parameter storage than in forward arithmetic because the
convolutional representation is shared.
\section{Experimental Setup and Evaluation}
\label{sec:experiments}

Atari-57 is the primary benchmark for the three-phase design study, while
Procgen Hard provides a secondary evaluation of the final Aftab architecture
under procedural variation. All architecture selection is completed on
Atari-57 before Procgen evaluation. The implementation, configurations, and
experimental outputs are available at
\url{https://github.com/tahashieenavaz/aftab}, with a mirror at
\url{https://github.com/lorisnanni/aftab}. An archived version associated
with this study is permanently available through Zenodo at
\url{https://zenodo.org/records/22812154}.

\subsection{Atari-57 Protocol}
\label{sec:atari_protocol}

The primary study uses the complete Atari-57 benchmark
\citep{bellemare2013arcade} through the corresponding EnvPool Atari
\nolinkurl{-v5} environments \citep{weng2022envpool}. Observations consist of
four stacked grayscale frames resized to $84\times84$ and normalized by
dividing unsigned 8-bit pixel values by $255$.

Training uses 128 parallel environments and 32 transitions per environment
per update, producing a rollout batch of 4096 transitions. Each rollout is
randomly permuted into 32 mini-batches of 128 samples and optimized for two
epochs. Atari uses a frame skip of four, four-frame stacking,
\nolinkurl{noop\_max}=30, the minimal action set, and zero repeat-action
probability. Episodic life is enabled during training and disabled during
evaluation.

Learning uses sign-clipped rewards, but reported Atari scores use the original
unclipped ALE rewards retained by EnvPool. Thus, clipped rewards are used to
construct $\lambda$-return targets, whereas raw Atari scores and
human-normalized scores remain on the original game-score scale. Similarly,
life loss may terminate a training transition under episodic life, but reported
game scores are accumulated until true environment termination rather than
treating life-loss pseudo-episodes as separate games.

Eight test environments are maintained alongside training. For training seed
$s$, test environments use seed $s+1000$. Their interactions are excluded from
the training batch and interaction budget. Single-head models act greedily at
evaluation; multi-head models use the majority-vote procedure described in
Section~\ref{sec:method_phase3}.

Each run uses a nominal budget of 200 million observed ALE frames. With frame
skip four, this corresponds to 50 million training action transitions.
Complete rollout collection produces 12,208 updates and 50,003,968 training
transitions, equivalent to 200,015,872 observed frames; this is referred to
throughout as the 200-million-frame budget.

\subsection{Atari Score Aggregation and Human Normalization}
\label{sec:atari_score_aggregation}

All configurations are trained with the same four independent seeds. Let
$G_{m,e,s}$ denote the evaluation score for method $m$, game $e$, and seed
$s$. The game-level score used in the aggregate analysis is the four-seed
mean,
\begin{equation}
\bar{G}_{m,e}
=
\frac{1}{4}\sum_{s=1}^{4}G_{m,e,s}.
\label{eq:seed_average}
\end{equation}


Atari performance is normalized as
\begin{equation}
\mathrm{HNS}_{m,e}
=
\frac{
\bar{G}_{m,e}-G_e^{\mathrm{random}}
}{
G_e^{\mathrm{human}}-G_e^{\mathrm{random}}
},
\label{eq:hns}
\end{equation}
where $G_e^{\mathrm{random}}$ and $G_e^{\mathrm{human}}$ are fixed reference
scores for game $e$. HNS values are not clipped, so values below zero or
above one are retained.


\subsection{Atari Aggregate Metrics and Statistical Analysis}
\label{sec:atari_metrics}

The primary aggregate metric is the interquartile mean (IQM) HNS. The
implementation orders the 57 game-level HNS values, removes 14 values from
each tail, and averages the central 29 values. Median HNS is also reported for
comparison with prior Atari studies.

IQM 95\% confidence intervals are obtained by bootstrap resampling across the
57 game-level values and recomputing IQM. Because each game-level value is
already averaged across four training seeds, these intervals characterize
task-level variation around the seed-averaged estimates rather than the full
between-run uncertainty.


Pairwise comparisons use two-sided Wilcoxon signed-rank tests
\citep{wilcoxon1945individual} on the paired 57-game HNS vectors. The
inferential unit is therefore the Atari game; the four seeds are not treated
as 228 independent observations. Multiple comparisons are corrected using
Bonferroni adjustment within each experimental family. Phase~1 contains
$\binom{9}{2}=36$ comparisons, while the reported four-configuration Phase~2
and Phase~3 families each contain $\binom{4}{2}=6$. Statistical significance
is assessed at family-wise $\alpha=0.05$. A non-significant result is not
interpreted as evidence of equivalence.

We additionally report an empirical game-level Probability of Improvement
(PoI),
\begin{equation}
\mathrm{PoI}(X,Y)
=
\frac{1}{57}
\sum_{e=1}^{57}
\left[
\mathbb{I}
\!\left(
\mathrm{HNS}_{X,e}>\mathrm{HNS}_{Y,e}
\right)
+
\frac{1}{2}
\mathbb{I}
\!\left(
\mathrm{HNS}_{X,e}=\mathrm{HNS}_{Y,e}
\right)
\right].
\label{eq:poi}
\end{equation}
A value of $0.5$ indicates balanced game-level performance. This quantity is
descriptive: it is an empirical proportion over the evaluated Atari suite,
not a Bayesian posterior probability or a probability over hypothetical
independent training runs.

\subsection{Procgen Hard Protocol}
\label{sec:procgen_protocol}

After completing all architecture selection on Atari-57, Aftab and the
independently rerun PQN reference are evaluated on all 16 Procgen Hard
environments \citep{cobbe2019procgen}. Procgen observations are native RGB
images of shape $3\times64\times64$ and are normalized by dividing pixel
values by $255$.

Training uses 64 parallel environments and 256 transitions per environment,
producing 16,384 transitions per rollout. Each rollout is divided into
32 mini-batches of 512 samples and optimized for two epochs. Atari-specific
options such as frame skipping, frame stacking, no-op initialization, reward
clipping, and episodic-life termination are not imposed on Procgen when they
are unsupported by the environment.

Eight test environments use the same $s+1000$ seed offset as Atari. Procgen
has no Atari-style frame skip in these experiments, so one transition
corresponds to one environment interaction. The nominal 200-million-interaction
budget produces 12,208 complete updates, or 200,015,872 training interactions.

\subsection{Procgen Performance Metrics}
\label{sec:procgen_metrics}

Raw episodic scores are retained for all 16 environments. Because reward
scales differ substantially across tasks, aggregate comparisons additionally
use a within-study Procgen Normalized Score (PNS). For each environment $e$,
let
\[
G_e^{\min}
=
\min_{m\in\{\mathrm{PQN},\mathrm{Aftab}\},\,s}G_{m,e,s},
\qquad
G_e^{\max}
=
\max_{m\in\{\mathrm{PQN},\mathrm{Aftab}\},\,s}G_{m,e,s}.
\]
The seed-level normalized score is
\begin{equation}
\mathrm{PNS}_{m,e,s}
=
\frac{
G_{m,e,s}-G_e^{\min}
}{
G_e^{\max}-G_e^{\min}
},
\label{eq:pns}
\end{equation}
and the four seed-level values are averaged within each environment. Median
and IQM are then computed across the resulting 16 environment-level scores,
with task-level bootstrap confidence intervals calculated analogously to
Atari.

Because the normalization bounds are derived from the PQN and Aftab runs in
this experiment, PNS is comparison-dependent and is not an externally
standardized Procgen score. Raw environment-level results are therefore
reported alongside the normalized aggregate.

Learning over the complete trajectory is summarized by normalized area under
the learning curve,
\begin{equation}
\mathrm{nAUC}_m
=
\frac{1}{T}
\int_0^T
\mathrm{PNS}_m(t)\,dt,
\label{eq:nauc}
\end{equation}
where $T=200$ million interactions and the integral is approximated by the
trapezoidal rule. Terminal IQM PNS characterizes end-of-training performance,
whereas nAUC summarizes normalized performance accumulated over the full
training trajectory. Raw-score AUC is additionally computed within each
environment but is not aggregated directly across environments because their
reward scales are not commensurate.


\subsection{Optimization and Reproducibility}
\label{sec:hyperparameters}

The main training settings are summarized in
Table~\ref{tab:hyperparameters_compact}. Both benchmarks use Rectified Adam
(RAdam) \citep{Liu2019radam}. A complete configuration table is provided in
the Supplementary Material.

\begin{table*}[ht]
\centering
\caption{\textbf{Principal training settings.}}
\label{tab:hyperparameters_compact}
\begin{tabular}{l c c}
\toprule
\textbf{Setting} & \textbf{Atari-57} & \textbf{Procgen Hard} \\
\midrule
Observation shape & $4\times84\times84$ & $3\times64\times64$ \\
Training environments & 128 & 64 \\
Rollout length & 32 & 256 \\
Batch size & 4,096 & 16,384 \\
Mini-batch size & 128 & 512 \\
Optimization epochs & 2 & 2 \\
Training budget & 200M observed frames & 200M interactions \\
Learning rate & \multicolumn{2}{c}{$2.5\times10^{-4}$} \\
Optimizer & \multicolumn{2}{c}{RAdam} \\
Discount $\gamma$ & \multicolumn{2}{c}{0.99} \\
$\lambda$ & \multicolumn{2}{c}{0.65} \\
Weight decay & \multicolumn{2}{c}{0} \\
Gradient-norm clipping & \multicolumn{2}{c}{10} \\
Exploration & \multicolumn{2}{c}{linear $\epsilon$: $1.0\rightarrow0.001$
over first 10\%} \\
Training seeds & \multicolumn{2}{c}{4} \\
\bottomrule
\end{tabular}
\end{table*}

Scalar PQN configurations, Phase~1 and Phase~2 variants, and Ensemble Dueling
use half mean-squared error against the bootstrapped $\lambda$-return.
Categorical Dueling and Aftab use the HL-Gauss cross-entropy objective
described in Section~\ref{sec:prelim_distributional}, with 51 support bins
over $[-10,10]$ and sigma ratio $0.75$. Categorical value-change clipping is
disabled.

The four fixed training seeds are
$\{475284,219842,525975,909314\}$ and are shared across competing
configurations to preserve pairing. The released environment pins EnvPool
1.2.0, Gymnasium 1.3.0, PyTorch 2.11.0, and
\nolinkurl{hl-gauss-pytorch} 0.2.3. CUDA training uses float16 automatic mixed
precision with gradient scaling, channels-last storage, and
\nolinkurl{torch.compile} with static shapes and
\nolinkurl{mode="max-autotune"}. Experiments were run on NVIDIA A40 GPUs with
48\,GB of memory; the most demanding configuration required approximately
13 hours of wall-clock training.

Our PQN and Hadamax references are run independently within the same PyTorch
pipeline using the same four study seeds. Published values are used only as
external reproduction references and are not mixed with our experimental
samples or statistical tests. The public repository contains network
definitions, configuration files, processed score tables, learning curves,
statistical matrices, and Procgen AUC outputs.
\section{Results}
\label{sec:results}

We first evaluate the independently implemented PQN and Hadamax references,
then report the three stages of the Atari-57 design study and the secondary
Procgen Hard evaluation. Unless otherwise stated, Atari aggregates are
computed from the 57 game-level HNS values after averaging the four training
seeds, and pairwise inference uses two-sided Wilcoxon signed-rank tests with
Bonferroni correction within the corresponding experimental family.

\subsection{Reproduction of the PQN and Hadamax Reference Baselines}
\label{sec:results_reproduction}

Direct comparison with the original PQN study requires care.
\citet{Gallici2025pqn} report a full Atari-57 median HNS of 2.45 after
400 million frames but do not provide the corresponding full-suite median at
200 million frames. We therefore do not treat the 400-million-frame result as
a numerical reproduction target.

The Hadamax study \citep{Kooietal2025hadamax} reports 200-million-frame
per-game scores for both Hadamax-PQN and its PQN reference over Atari-57.
Using the same human/random normalization as in our analysis, we recomputed
median and IQM HNS from those published score tables and compare them with our
independent four-seed PyTorch runs in
Table~\ref{tab:baseline_reproduction}. The published Hadamax comparison uses
five seeds and differs in implementation and preprocessing, so this is a
diagnostic reproduction check rather than an exact replication.

\begin{table}[ht]
\centering
\caption{\textbf{External reproduction check.} Published-source values are
recomputed from the 200M-frame per-game scores of
\citet{Kooietal2025hadamax}; they are not aggregate statistics explicitly
reported by the original authors.}
\label{tab:baseline_reproduction}
\begin{tabular}{l c c}
\toprule
\textbf{Configuration} & \textbf{Median HNS} & \textbf{IQM HNS} \\
\midrule
Published PQN reference$^\dagger$ & 1.789 & 2.781 \\
Our PQN reference                & 1.894 & 2.715 \\
\midrule
Published Hadamax-PQN$^\dagger$  & 3.071 & 4.751 \\
Our Hadamax reference            & 2.829 & 4.712 \\
\bottomrule
\end{tabular}

\vspace{1mm}
{\footnotesize
$^\dagger$Recomputed in this study from the published per-game raw-score
table.}
\end{table}

Our PQN IQM differs from the recomputed published value by approximately
$-2.4\%$, and our Hadamax IQM by approximately $-0.8\%$. The corresponding
median differences are approximately $+5.9\%$ and $-7.9\%$, respectively.
Given the differences in seeds, implementation framework, Hadamax input
resolution, and padding semantics, the aggregate agreement supports use of
the independent PyTorch implementations as internal references. All
subsequent controlled comparisons use only our four-seed runs.

\subsection{Phase 1: Encoder Topology and Computational Complexity}
\label{sec:results_phase1}

Table~\ref{tab:phase1_aggregate} summarizes Phase~1. Alpha attains the
largest IQM HNS, $3.566$, followed closely by Gamma at $3.508$, compared with
$2.715$ for PQN.

\begin{table}[ht]
\centering
\caption{\textbf{Phase~1 Atari-57 results.} Values are computed from the
four-seed mean for each game.}
\label{tab:phase1_aggregate}
\resizebox{\columnwidth}{!}{%
\begin{tabular}{l c c c}
\toprule
\textbf{Configuration}
& \textbf{Median HNS}
& \textbf{IQM HNS}
& \textbf{IQM 95\% CI} \\
\midrule
PQN     & 1.894 & 2.715 & [1.541, 4.477] \\
\textbf{Alpha} & \textbf{2.618} & \textbf{3.566} & [2.138, 6.292] \\
Beta    & 2.421 & 3.464 & [1.909, 6.242] \\
Gamma   & 2.577 & 3.508 & [2.160, 7.182] \\
Delta   & 1.596 & 2.388 & [1.393, 3.870] \\
Epsilon & 2.266 & 3.341 & [1.851, 6.077] \\
Zeta    & 2.067 & 3.199 & [1.802, 5.791] \\
Eta     & 1.879 & 3.106 & [1.708, 5.692] \\
Theta   & 1.830 & 2.688 & [1.535, 4.504] \\
\bottomrule
\end{tabular}%
}
\end{table}

The Alpha--Gamma IQM difference is only $0.058$, and their paired comparison
is not resolved after correction ($p_{\mathrm{adj}}=1.000$). Both differ from
PQN after Bonferroni correction ($p_{\mathrm{adj}}<0.001$). The descriptive
PoI is approximately $0.85$ for Alpha versus PQN and $0.79$ for Gamma versus
PQN, whereas Alpha versus Gamma is nearly balanced
($\mathrm{PoI}\approx0.51$).

Because Alpha and Gamma both produce $F=3136$, their downstream scalar heads
are identical. Gamma nevertheless reduces encoder parameters from
$174{,}752$ to $117{,}168$ (approximately $33\%$) and convolutional MACs
from $27.542$ to $22.902$ million (approximately $17\%$). Eta illustrates
why convolutional parameter count alone is insufficient: despite only
$78{,}400$ encoder parameters, its $46{,}208$-dimensional flattened
representation produces a complete scalar model of approximately
$23.74$ million parameters for an eight-action environment, compared with
approximately $1.73$ million for Gamma, while attaining lower IQM HNS.

The task-composition sensitivity analysis selects Gamma in all 10 randomly
formed 15-game subsets. Because these subsets reuse the same trained agents,
this is interpreted as selection robustness to sampled task composition, not
as independent validation. Under the predefined performance--complexity rule,
Gamma is therefore carried forward to Phase~2.

Complete game-level results and PoI matrices are provided in the
Supplementary Material.

\subsection{Phase 2: Hadamax Integration}
\label{sec:results_phase2}

The Phase~2 aggregate results are shown in
Table~\ref{tab:phase2_aggregate}. Gamma-Hadamax-Valid and
Gamma-Hadamax-Same increase IQM HNS from $3.508$ for scalar Gamma to
$5.254$ and $5.360$, respectively.

\begin{table}[ht]
\centering
\caption{\textbf{Phase~2 Atari-57 results.}}
\label{tab:phase2_aggregate}
\resizebox{\columnwidth}{!}{%
\begin{tabular}{l c c c}
\toprule
\textbf{Configuration}
& \textbf{Median HNS}
& \textbf{IQM HNS}
& \textbf{IQM 95\% CI} \\
\midrule
Gamma
& 2.577 & 3.508 & [2.182, 7.036] \\

Hadamax Reference
& 2.829 & 4.712 & [2.635, 9.232] \\

\textbf{Gamma-Hadamax-Valid}
& \textbf{3.322} & 5.254 & [2.990, 10.524] \\

Gamma-Hadamax-Same
& 3.143 & \textbf{5.360} & [3.034, 11.111] \\
\bottomrule
\end{tabular}%
}
\end{table}

Both Gamma-Hadamax variants differ from scalar Gamma after correction
(Valid: $p_{\mathrm{adj}}=0.005$; Same:
$p_{\mathrm{adj}}<0.001$), whereas their direct comparison is not resolved
($p_{\mathrm{adj}}=1.000$). Their higher point estimates relative to the
three-stage Hadamax reference are also not resolved after correction
(Valid: $p_{\mathrm{adj}}=0.226$; Same:
$p_{\mathrm{adj}}=0.056$). The Hadamax-reference versus Gamma comparison is
likewise not resolved ($p_{\mathrm{adj}}=0.145$). Thus, the results support
an improvement of both Gamma-Hadamax variants over scalar Gamma, but do not
establish superiority over the independently implemented Hadamax reference.

PoI values are approximately $0.70$ and $0.75$ for Valid and Same versus
Gamma, respectively, while their direct comparison is approximately balanced.
The two variants contain identical encoder parameter counts, but
Gamma-Hadamax-Valid produces $F=3136$ rather than $F=6400$. For a
representative eight-action environment, this corresponds to approximately
$1.85$ versus $3.52$ million total parameters. Valid also requires
approximately $122.0$ million convolutional MACs compared with $129.3$
million for Same and $159.0$ million for the Hadamax reference.

These savings are relative to the Hadamax alternatives, not to plain Gamma:
Gamma requires only approximately $22.9$ million convolutional MACs.
Hadamax integration therefore improves aggregate performance at a substantial
increase in convolutional arithmetic. Because Same provides only a $0.106$
larger IQM point estimate than Valid, is not statistically resolved from it,
and has a substantially larger downstream representation,
Gamma-Hadamax-Valid is selected for Phase~3.

Complete game-level results and selected pairwise statistics are provided in
the Supplementary Material

\subsection{Phase 3: Value-Estimation Architectures}
\label{sec:results_phase3}

Phase~3 fixes the Gamma-Hadamax-Valid representation and compares complete
downstream value-estimation configurations. Because the configurations differ
in objective and, for Aftab, head activation, the results do not constitute a
factorial decomposition of categorical regression, dueling, ensembling, or
activation choice.

\begin{table}[ht]
\centering
\caption{\textbf{Phase~3 Atari-57 results.} Scalar Gamma-Hadamax-Valid is
the representation-level reference carried forward from Phase~2.}
\label{tab:phase3_results_ci}
\resizebox{\columnwidth}{!}{%
\begin{tabular}{l c c c}
\toprule
\textbf{Configuration}
& \textbf{Median HNS}
& \textbf{IQM HNS}
& \textbf{IQM 95\% CI} \\
\midrule
PQN
& 1.894 & 2.715 & [1.533, 4.486] \\

Scalar Gamma-Hadamax-Valid
& 3.322 & 5.254 & [2.990, 10.524] \\

Categorical Dueling
& 4.261 & 6.093 & [3.418, 10.868] \\

Ensemble Dueling
& 3.636 & 5.625 & [3.279, 11.988] \\

\textbf{Aftab}
& \textbf{4.374}
& \textbf{6.592}
& [3.524, 13.536] \\
\bottomrule
\end{tabular}%
}
\end{table}

All three modified heads have higher IQM point estimates than scalar
Gamma-Hadamax-Valid ($5.254$): Categorical Dueling reaches $6.093$,
Ensemble Dueling $5.625$, and Aftab $6.592$. Relative to the original PQN
reference, Aftab increases IQM HNS from $2.715$ to $6.592$.

The currently reported Bonferroni-corrected Phase~3 family contains PQN,
Categorical Dueling, Ensemble Dueling, and Aftab. Within this family, each
modified configuration differs from PQN
($p_{\mathrm{adj}}<0.001$). Categorical Dueling and Ensemble Dueling are not
resolved ($p_{\mathrm{adj}}=0.971$); Aftab differs from Categorical Dueling
($p_{\mathrm{adj}}=0.007$), whereas Aftab and Ensemble Dueling are not
resolved ($p_{\mathrm{adj}}=1.000$). Corresponding PoI values versus PQN are
approximately $0.82$, $0.93$, and $0.86$ for Categorical Dueling, Ensemble
Dueling, and Aftab, respectively.

Because scalar Gamma-Hadamax-Valid is not included in that corrected
four-configuration family, differences between it and the modified Phase~3
heads are interpreted descriptively rather than inferentially. Selected pairwise statistics and the complete PoI matrix are provided in the
Supplementary Material.

Aftab's higher aggregate performance also entails substantially greater model
capacity. For an eight-action environment, scalar Gamma-Hadamax-Valid contains
approximately $1.85$ million parameters and $123.61$ million counted forward
MACs, compared with approximately $34.73$ million parameters and
$156.46$ million MACs for Aftab. The final architecture is therefore
performance-oriented rather than parameter-efficient; the principal
performance--complexity trade-off identified by the study concerns the
Gamma-Hadamax-Valid representation.

\subsection{Performance on Procgen Hard}
\label{sec:results_procgen}

After all architecture selection on Atari-57, Aftab and the independently
rerun PQN reference are trained and evaluated on the 16 Procgen Hard
environments. No Procgen result is used to modify the architecture.

\begin{table}[ht]
\centering
\caption{\textbf{Procgen Hard summary.} PNS is the within-study
normalization defined in Section~\ref{sec:procgen_metrics}.}
\label{tab:procgen_summary}
\begin{tabular}{l c c}
\toprule
\textbf{Metric} & \textbf{PQN} & \textbf{Aftab} \\
\midrule
Median PNS
& \textbf{0.414} & 0.325 \\

IQM PNS
& 0.382 & \textbf{0.418} \\

IQM 95\% CI
& [0.194, 0.541]
& [0.125, 0.741] \\

nAUC
& 0.216 & \textbf{0.541} \\

Higher terminal raw score
& \textbf{9/16} & 7/16 \\

Higher raw-score AUC
& 4/16 & \textbf{12/16} \\
\bottomrule
\end{tabular}
\end{table}

Aftab obtains a terminal IQM PNS of $0.418$ compared with $0.382$ for PQN,
while the median has the opposite ordering ($0.325$ versus $0.414$).
Aftab has the higher final raw score in seven of the sixteen environments and
PQN in nine, indicating substantial task-level heterogeneity and providing no
basis for a claim of uniform terminal superiority.

The learning-trajectory comparison shows a different pattern. Aftab achieves
an nAUC of $0.541$ compared with $0.216$ for PQN and has the higher
environment-specific raw-score AUC in 12 of 16 tasks. Its clearest Procgen
difference therefore concerns performance accumulated over the learning
trajectory rather than consistently higher final performance.

\begin{figure*}[ht]
    \centering
    \includegraphics[width=0.8\textwidth]
    {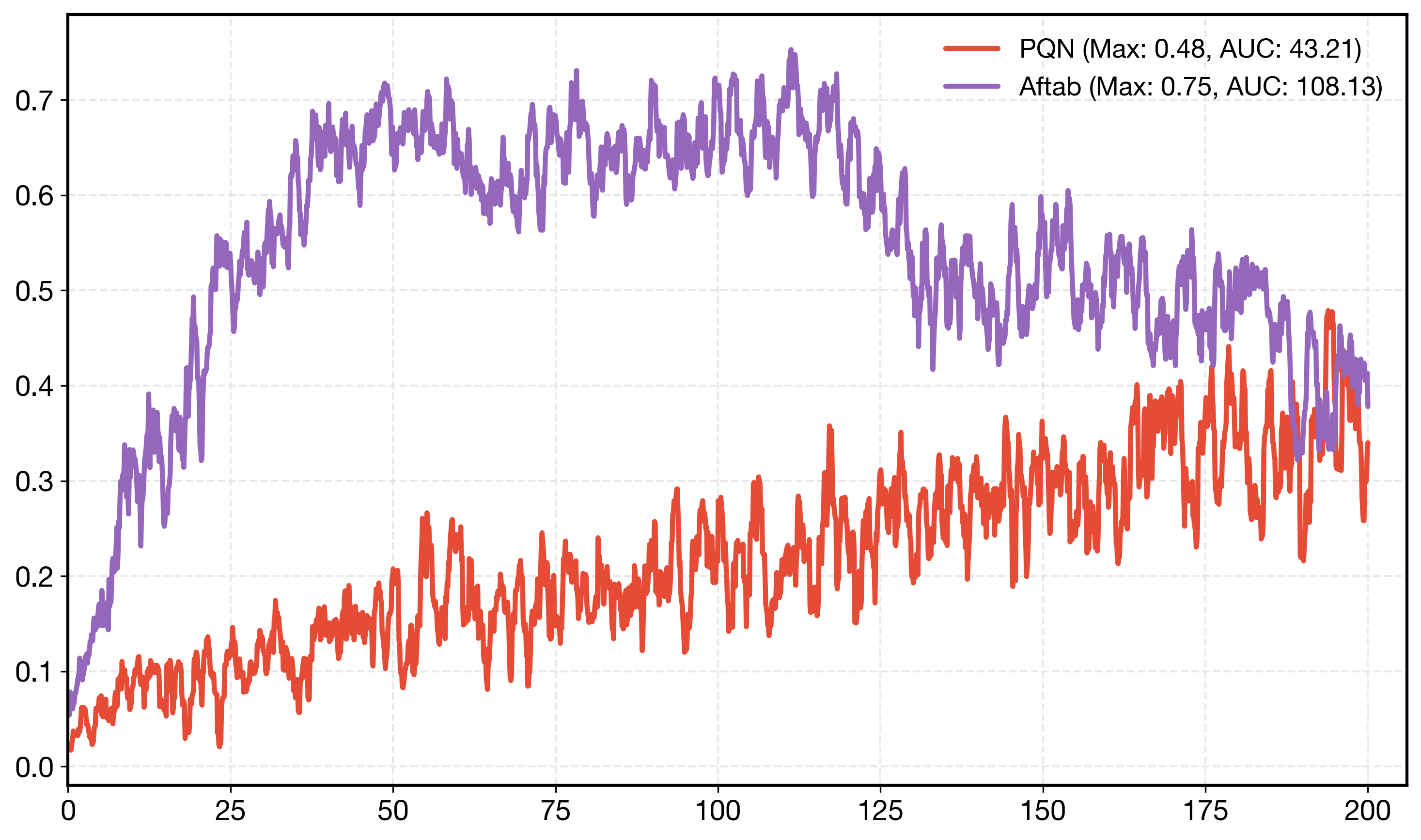}
    \caption{\textbf{Procgen Hard evaluation.}
    Aggregate normalized learning trajectories of PQN and Aftab across the
    16 environments over the 200M-interaction training budget. Terminal IQM
    PNS summarizes end-of-training aggregate performance; nAUC summarizes
    normalized performance accumulated over the training trajectory.}
    \label{fig:procgen}
\end{figure*}

Figure~\ref{fig:procgen} shows the aggregate normalized trajectories.
Environment-level terminal scores, raw-score AUC values, and individual
learning curves are provided in the Supplementary Material.

Overall, the Procgen experiment shows that the architecture selected on
Atari remains competitive when retrained under procedural variation and can
produce stronger aggregate learning-trajectory performance. It does not
establish uniform final superiority, direct transfer of an Atari-trained
representation, or formal out-of-distribution generalization.
\section{Discussion and Limitations}
\label{sec:discussion}

The three-stage study indicates that architectural choices can materially
affect replay-free parallelized Q-learning, but the resulting trade-offs
differ across the visual representation and downstream value-estimation
stages. In particular, the results do not support a simple ``larger is
better'' explanation: model behavior depends on how convolutional topology,
spatial reduction, multiplicative interactions, and value-estimation
components are organized. At the same time, the progressive experimental
design compares complete configurations rather than providing a factorial
decomposition of individual architectural mechanisms.

\subsection{What the Design Study Reveals}
\label{sec:discussion_phases}

\paragraph{Encoder topology matters beyond parameter count}
Phase~1 shows that parameter count alone is insufficient to explain
performance. Alpha attains the highest encoder-level IQM HNS point estimate,
while Gamma achieves a closely matched result with fewer encoder parameters
and fewer convolutional MACs. Their corrected pairwise comparison does not
resolve a performance ordering, and both produce the same
$3136$-dimensional flattened representation. Gamma is therefore selected
under the predefined performance--complexity rule rather than because it
achieves a statistically established performance advantage.

The remaining candidates reinforce the importance of the representation
delivered to the value head. Eta, for example, has a relatively small
convolutional parameter count but produces a much larger flattened feature
vector, which substantially increases complete-model size without yielding a
higher aggregate score. Complexity in visual value-based agents should
therefore be considered at the level of the complete network rather than from
convolutional parameter count alone.

These comparisons do not identify an independent causal effect of depth.
The evaluated encoders simultaneously differ in kernel sizes, channel
progression, stride, padding, depth, and spatial resolution. Phase~1 therefore
provides evidence about complete convolutional topologies rather than about
any single architectural factor. The 15-game subset analysis further suggests
that the selection of Gamma is not highly sensitive to task composition:
Gamma is selected in all 10 sampled subsets. Because the same trained agents
are reused, however, this constitutes a sensitivity analysis rather than
independent validation.

\paragraph{Hadamax integration improves Gamma at increased arithmetic cost}
Phase~2 shows that Hadamax-style processing can substantially improve the
Gamma representation. Both Gamma-Hadamax variants outperform scalar Gamma
under the corrected paired analysis, while their difference from one another
is not statistically resolved. Their higher aggregate point estimates
relative to the independently implemented three-stage Hadamax reference are
also not resolved after multiple-comparison correction. The results therefore
support the compatibility of the Gamma hierarchy with Hadamax-style
multiplicative processing, but do not establish statistical superiority of
the five-stage variants over the original three-stage Hadamax design.

The comparison between Gamma-Hadamax-Valid and Gamma-Hadamax-Same also
illustrates the importance of spatial dimensionality. The two encoders have
identical encoder parameter counts, but Same preserves a substantially larger
flattened representation and therefore enlarges the downstream value head.
Because its small IQM advantage over Valid is not statistically resolved,
Gamma-Hadamax-Valid is preferred as the lower-complexity configuration. This
is a pragmatic performance--complexity decision rather than evidence that
Valid has higher underlying performance.

The efficiency claim must nevertheless be qualified. Gamma-Hadamax-Valid is
compact relative to the other Hadamax configurations, but it requires
substantially more convolutional arithmetic than plain Gamma. Its advantage
should therefore be understood as a favorable trade-off \emph{among the
evaluated Hadamax configurations}, not as a reduction in computational cost
relative to the original Gamma encoder.

\paragraph{Value-estimation configurations provide further gains at substantial
parameter cost}
Phase~3 fixes Gamma-Hadamax-Valid and compares complete downstream
value-estimation architectures. Categorical Dueling, Ensemble Dueling, and
Aftab all produce higher aggregate IQM point estimates than the scalar
Gamma-Hadamax-Valid configuration, with Aftab attaining the largest observed
IQM and median among the evaluated final configurations.

These comparisons require two qualifications. First, the currently reported
Bonferroni-corrected Phase~3 family contains PQN, Categorical Dueling,
Ensemble Dueling, and Aftab rather than the scalar Gamma-Hadamax-Valid
configuration. Consequently, improvements relative to the scalar Phase~2
reference are currently descriptive point-estimate comparisons rather than
corrected inferential claims. Second, Phase~3 is not a factorial ablation:
the evaluated configurations combine categorical regression, dueling
decomposition, ensembling, and, in Aftab, a different head activation.
Aftab's performance therefore cannot be attributed uniquely to any one of
these mechanisms.

Aftab also substantially increases parameter storage through its ten
independently parameterized categorical dueling heads. The increase in
forward arithmetic is proportionally smaller because the visual encoder is
shared, but the complete Aftab model is nevertheless performance-oriented
rather than parameter-efficient. The clearest performance--complexity result
of the study concerns the Gamma-Hadamax-Valid representation rather than the
complete Aftab architecture.

\paragraph{Categorical and ensemble mechanisms require careful terminology}
The categorical objective used by Categorical Dueling and Aftab converts the
scalar bootstrapped $\lambda$-return into an HL-Gauss categorical target. It
does not model a Bellman-updated return distribution or use the categorical
projection of C51. Its effect should therefore be described as categorical
regression rather than distributional reinforcement learning in the strict
C51 sense.

The ensemble similarly differs from classical Bootstrapped DQN. Individual
heads can control parallel environments for temporally extended periods, but
the bootstrap inclusion probability is $p=1$, so all heads are trained on
every collected transition. The heads retain separate parameters and
head-specific bootstrapping targets, and their active policies influence the
trajectories collected during training, but they are not trained on distinct
bootstrap resamples. The present results therefore do not establish posterior
sampling, classical bootstrap uncertainty estimation, or variance reduction.

\subsection{Performance Under Procedural Variation}
\label{sec:discussion_procgen}

Procgen Hard provides a secondary evaluation after completion of the Atari
design process and reveals a clear distinction between learning-trajectory
and terminal performance. Aftab achieves the higher terminal IQM PNS, whereas
PQN retains the higher median and the higher final raw score in nine of the
sixteen environments. Terminal performance is therefore heterogeneous and
does not support a claim of uniformly improved final behavior.

The learning-trajectory comparison is more favorable to Aftab: its normalized
area under the learning curve is substantially higher, and it obtains the
larger environment-specific raw-score AUC in 12 of the 16 tasks. This pattern
is consistent with stronger performance over substantial portions of
training, even when the advantage is not always retained at the final
checkpoint.

The disagreement between AUC and terminal performance may reflect faster
early learning, stronger intermediate performance, late-training degradation,
or a combination of these effects; the present metrics do not distinguish
among them. A high AUC should therefore not be interpreted as evidence of
stable asymptotic superiority. Procgen shows that the selected architecture
can remain competitive and improve aggregate learning-trajectory behavior
when retrained under procedural variation, but it does not demonstrate direct
transfer of an Atari-trained representation or formal out-of-distribution
generalization.

\subsection{Limitations}
\label{sec:limitations}

Several limitations constrain the scope of these conclusions.

First, the experimental design introduces both benchmark-selection and
factorial limitations. Atari-57 is used for architecture selection as well as
for reporting the headline Atari performance, so the final Atari result is an
in-benchmark design outcome and may overestimate improvement on unseen task
distributions. Procgen was not used for architecture selection, but its
different environment distribution and two-agent comparison do not constitute
independent validation of every design decision. Moreover, the progressive
design fixes Gamma before Phase~2 and Gamma-Hadamax-Valid before Phase~3.
This makes full-suite evaluation computationally feasible but prevents
estimation of all main effects and higher-order interactions; Hadamax is not
tested with every encoder, and the Phase~3 heads are not tested across every
visual representation.

Second, uncertainty and optimization are only partially explored. Each Atari
configuration uses four training seeds, which are averaged before the 57 games
are treated as paired inferential units. The reported task-level bootstrap
confidence intervals therefore characterize variation across games around
seed-averaged estimates rather than fully representing training-run
uncertainty. Additional seeds and hierarchical analyses preserving both
task-level and run-level variation would provide a more complete uncertainty
assessment. In addition, major optimization settings are deliberately shared
across configurations to support controlled architectural comparison. This
improves comparability but does not establish the maximum achievable
performance of each architecture under configuration-specific tuning.

Third, several findings remain implementation-specific. The Hadamax reference
is an independent PyTorch reimplementation rather than an exact reproduction
of the published JAX system: the observation resolution and padding semantics
differ, although aggregate baseline performance is closely reproduced. The
categorical configurations likewise use a particular 51-bin support over
$[-10,10]$, HL-Gauss smoothing, and range clamping. Multi-step
$\lambda$-returns are not guaranteed to remain inside this support, and the
present experiments do not quantify the frequency or effect of
support-boundary saturation. Sensitivity to support range, bin count,
smoothing width, and clamping therefore remains to be studied.

Finally, external validity is limited. The Procgen Normalized Score is an
internal min--max normalization whose bounds depend on the PQN and Aftab runs
included in this experiment; it is useful for within-study aggregation but is
not an externally standardized benchmark score. Raw environment-level results
should therefore remain the basis for task-specific interpretation. More
generally, Atari-57 and Procgen Hard are both discrete-action visual domains.
The present experiments do not establish that the same architectural choices
benefit continuous control, partially observed environments,
language-conditioned tasks, or substantially different sensory modalities.

\subsection{Future Directions}
\label{sec:future_work}

The limitations above motivate several direct extensions. A broader factorial
study could test selected combinations of encoder topology, Hadamax
processing, categorical regression, dueling decomposition, ensemble structure,
and activation function while additional seeds would support hierarchical
uncertainty analysis across both tasks and independent runs. The large
parameter cost of Aftab also motivates head sharing, low-rank
parameterization, and smaller ensembles to determine how much of its
performance can be retained with a more compact value architecture.

Further work should examine sensitivity of HL-Gauss to support range, bin
count, smoothing width, and target saturation, and evaluate the selected
architectural principles in additional visual and non-visual reinforcement
learning domains. For Procgen specifically, final-window averages,
peak-to-final degradation, and related stability measures would complement
AUC and help distinguish rapid learning from persistent end-of-training
performance.


\section{Conclusion}
\label{sec:conclusion}

This work investigated architectural design in replay-free parallelized
Q-learning through a progressive three-phase study of convolutional encoder
topology, Hadamax-style representation learning, and downstream
value-estimation architecture. Using PQN as a common training framework, all
architecture selection was conducted on the complete Atari-57 benchmark,
with Procgen Hard used only as a subsequent secondary evaluation under
procedural variation.

The results show that architectural organization matters beyond model size.
Phase~1 identified Gamma as a favorable performance--complexity choice among
the evaluated convolutional topologies, while Phase~2 showed that integrating
Hadamax-style multiplicative processing into this hierarchy can substantially
improve aggregate Atari performance, albeit at greater convolutional cost.
Gamma-Hadamax-Valid was selected as the fixed representation because it
provided a more favorable downstream complexity trade-off than
Gamma-Hadamax-Same without a statistically resolved performance disadvantage.

With this representation fixed, Phase~3 compared complete value-estimation
configurations rather than isolated mechanisms. Aftab achieved the highest
observed Atari-57 aggregate performance, with an IQM human-normalized score of
$6.592$ compared with $2.715$ for our independently rerun PQN reference and
a game-level Probability of Improvement of $0.86$. This performance comes at
substantial parameter cost because Aftab uses ten independently parameterized
categorical dueling heads; the final architecture should therefore be regarded
as performance-oriented rather than parameter-efficient.

On Procgen Hard, Aftab achieved a higher terminal IQM normalized score than
PQN ($0.418$ versus $0.382$) and a higher normalized area under the learning
curve ($0.541$ versus $0.216$), while PQN retained the higher median and the
higher final raw score in nine of the sixteen environments. These results
therefore support stronger aggregate learning-trajectory performance under
procedural variation rather than uniform terminal superiority, direct transfer
from Atari, or formal out-of-distribution generalization.

Overall, the study indicates that visual topology, multiplicative
representation, spatial reduction, and value-estimation design can
substantially influence replay-free Q-learning, and that architecture
selection should consider statistical evidence and computational complexity
alongside aggregate performance. The resulting Aftab implementation and
released experimental artifacts provide a reproducible basis for testing
these design choices with broader factorial comparisons, additional training
seeds, and more parameter-efficient multi-head architectures.


\section*{CRediT authorship contribution statement}

\textbf{Taha Shieenavaz:}
Conceptualization, Methodology, Software, Writing -- original draft.

\textbf{Shabnam Zareshahraki:}
Methodology, Validation, Writing -- original draft.

\textbf{Loris Nanni:}
Supervision, Project administration, Writing -- review \& editing.

\section*{Declaration of Competing Interest}

The authors declare that they have no known competing financial interests or
personal relationships that could have appeared to influence the work
reported in this paper.

\section*{Data Availability Statement}

The complete Aftab framework, including model definitions, training
configurations, reproducibility settings, processed results, and experimental
artifacts, is publicly available at
\url{https://github.com/tahashieenavaz/aftab}.
An archived version associated with this study is permanently available
through Zenodo at
\url{https://zenodo.org/records/22812154}.

\section*{Funding}

This research did not receive any specific grant from funding agencies in the
public, commercial, or not-for-profit sectors.

\section*{Declaration of Generative AI and AI-Assisted Technologies in the
Writing Process}

During the preparation of this work, the authors used ChatGPT (OpenAI) to
assist with language editing, structural condensation, consistency checking,
and journal-format compliance review. After using this tool, the authors
reviewed and edited the content as needed and take full responsibility for the
content of the publication.

\bibliography{references}

@article{weng2022envpool,
    author = {Weng, Jiayi and Huang, Min Lin and Huang, Shengyi and Bo, Hao and Makoviichuk, Denys},
    title = {EnvPool: A Highly Parallel Reinforcement Learning Environment Execution Engine},
    journal = {arXiv preprint arXiv:2206.10558},
    year = {2022}
}

@article{mnih2015nature,
  title     = {Human-level control through deep reinforcement learning},
  author    = {Mnih, Volodymyr and Kavukcuoglu, Koray and Silver, David and Rusu, Andrei A. and Veness, Joel and Bellemare, Marc G. and Graves, Alex and Riedmiller, Martin and Fidjeland, Andreas K. and Ostrovski, Georg and others},
  journal   = {Nature},
  volume    = {518},
  number    = {7540},
  pages     = {529--533},
  year      = {2015},
  publisher = {Nature Publishing Group},
  doi       = {10.1038/nature14236}
}

@article{mnih2013dqn,
  title   = {Playing Atari with Deep Reinforcement Learning},
  author  = {Mnih, Volodymyr and Kavukcuoglu, Koray and Silver, David and Graves, Alex and Antonoglou, Ioannis and Wierstra, Daan and Riedmiller, Martin},
  journal = {arXiv preprint arXiv:1312.5602},
  year    = {2013}
}

@article{Gallici2025pqn,
  title={Simplifying Deep Temporal Difference Learning},
  author={Gallici, Matteo and Fellows, Matt and Ellis, Benjamin and Pou, Arnuad and Masmitja, Enric and Foerster, Jakob and Martin, Mario},
  journal={arXiv preprint arXiv:2407.04811},
  year={2024}
}

@article{Hessel2018rainbow,
  title={Rainbow: Combining improvements in deep reinforcement learning},
  author={Hessel, Matteo and Modayil, Joseph and Van Hasselt, Hado and Schaul, Tom and Ostrovski, Georg and Dabney, Will and Horgan, Dan and Cotton, Bilal and Silver, David},
  journal={Proceedings of the AAAI Conference on Artificial Intelligence},
  volume={32},
  number={1},
  year={2018}
}

@inproceedings{Wang2016dueling,
  title={Dueling network architectures for deep reinforcement learning},
  author={Wang, Ziyu and Schaul, Tom and Hessel, Matteo and Hasselt, Hado and Lanctot, Marc and Freitas, Nando},
  booktitle={International Conference on Machine Learning (ICML)},
  pages={1995--2003},
  year={2016}
}

@inproceedings{Liu2019radam,
  title={On the Variance of the Adaptive Learning Rate and Beyond},
  author={Liu, Liyuan and Jiang, Haoming and He, Pengcheng and Chen, Weizhu and Liu, Xiaodong and Gao, Jianfeng and Han, Jiawei},
  booktitle={International Conference on Learning Representations (ICLR)},
  year={2020}
}

@article{ba2016layernorm,
  title={Layer normalization},
  author={Ba, Jimmy Lei and Kiros, Jamie Ryan and Hinton, Geoffrey E},
  journal={arXiv preprint arXiv:1607.06450},
  year={2016}
}

@inproceedings{He2016resnet,
  title={Deep residual learning for image recognition},
  author={He, Kaiming and Zhang, Xiangyu and Ren, Shaoqing and Sun, Jian},
  booktitle={Proceedings of the IEEE conference on computer vision and pattern recognition},
  pages={770--778},
  year={2016}
}

@inproceedings{Espeholt2018impala,
    author = {Espeholt, Lasse and Soyer, Hubert and Munos, Remi and Simonyan, Karen and Mnih, Volodymyr and Ward, Tom and Doron, Yotam and Firoiu, Vlad and Harley, Tim and Dunning, Iain and Legg, Shane and Kavukcuoglu, Koray},
    title = {IMPALA: Scalable Distributed Deep-RL with Importance Weighted Actor-Learner Architectures},
    booktitle = {Proceedings of the 35th International Conference on Machine Learning},
    pages = {1407--1416},
    year = {2018}
}

@article{Kooietal2025hadamax,
  author  = {Kooi, Jacob E. and Yang, Zhao and François-Lavet, Vincent},
  title   = {Hadamax Encoding: Elevating Performance in Model-Free Atari},
  journal = {arXiv preprint arXiv:2505.15345},
  year    = {2025},
  doi     = {10.48550/arXiv.2505.15345},
  url     = {https://arxiv.org/abs/2505.15345}
}

@article{hendrycks2016gelu,
  title   = {Gaussian Error Linear Units (GELUs)},
  author  = {Hendrycks, Dan and Gimpel, Kevin},
  journal = {arXiv preprint arXiv:1606.08415},
  year    = {2016},
  url     = {https://arxiv.org/abs/1606.08415}
}

@article{lecun1998gradient,
  title     = {Gradient-based learning applied to document recognition},
  author    = {LeCun, Yann and Bottou, L{\'e}on and Bengio, Yoshua and Haffner, Patrick},
  journal   = {Proceedings of the IEEE},
  volume    = {86},
  number    = {11},
  pages     = {2278--2324},
  year      = {1998},
  publisher = {IEEE}
}

@article{bellemare2013arcade,
  title   = {The Arcade Learning Environment: An Evaluation Platform for General Agents},
  author  = {Bellemare, Marc G. and Naddaf, Yavar and Veness, Joel and Bowling, Michael},
  journal = {Journal of Artificial Intelligence Research},
  volume  = {47},
  pages   = {253--279},
  year    = {2013},
  doi     = {10.1613/jair.3912},
  url     = {https://jair.org/index.php/jair/article/view/10819}
}

@inproceedings{bellemare2017distributional,
  title     = {A Distributional Perspective on Reinforcement Learning},
  author    = {Bellemare, Marc G and Dabney, Will and Mnih, Volodymyr},
  booktitle = {International Conference on Machine Learning},
  pages     = {449--458},
  year      = {2017},
  publisher = {PMLR}
}

@article{Zoph2017NASNet,
  title     = {Learning Transferable Architectures for Scalable Image Recognition},
  author    = {Zoph, Barret and Vasudevan, Vijay and Shlens, Jonathon and Le, Quoc V.},
  journal   = {arXiv preprint arXiv:1707.07012},
  year      = {2017},
  url       = {https://arxiv.org/abs/1707.07012}
}

@article{Lee2024Simba,
  title     = {Simba: Simplicity Bias for Scaling Up Parameters in Deep Reinforcement Learning},
  author    = {Lee, Hojoon and Hwang, Dongyoon and Kim, Donghu and Kim, Hyunghee and Tai, Jung-Joo and Subramanian, Kaushik and Wurman, Peter R. and Choo, Jaegul and Stone, Peter and Seno, Takuma},
  journal   = {arXiv preprint arXiv:2410.09754},
  year      = {2024},
  url       = {https://arxiv.org/abs/2410.09754}
}

@article{Bhatt2019CrossQ,
  title     = {CrossQ: Batch Normalization in Deep Reinforcement Learning for Greater Sample Efficiency and Simplicity},
  author    = {Bhatt, Aditya and Palenicek, Daniel and Belousov, Boris and Argus, Max and Amiranashvili, Artemij and Brox, Thomas and Peters, Jan},
  journal   = {arXiv preprint arXiv:1902.05605},
  year      = {2019},
  url       = {https://arxiv.org/abs/1902.05605}
}

@inproceedings{Schwarzer2023BBF,
  title     = {Bigger, Better, Faster: Human-level {A}tari with human-level efficiency},
  author    = {Schwarzer, Max and Obando Ceron, Johan Samir and Courville, Aaron and Bellemare, Marc G. and Agarwal, Rishabh and Castro, Pablo Samuel},
  booktitle = {Proceedings of the 40th International Conference on Machine Learning},
  pages     = {30365--30380},
  year      = {2023},
  publisher = {PMLR},
  url       = {https://proceedings.mlr.press/v202/schwarzer23a.html}
}

@inproceedings{Nauman2024BRO,
  title     = {Bigger, Regularized, Optimistic: scaling for compute and sample-efficient continuous control},
  author    = {Nauman, Michal and Ostaszewski, Mateusz and Jankowski, Krzysztof and Mi{\l}o{\'s}, Piotr and Cygan, Marek},
  booktitle = {Advances in Neural Information Processing Systems},
  year      = {2024},
  url       = {https://arxiv.org/abs/2405.16158}
}

@inproceedings{VanHasselt2016DoubleDQN,
  title     = {Deep Reinforcement Learning with Double Q-Learning},
  author    = {Van Hasselt, Hado and Guez, Arthur and Silver, David},
  booktitle = {Proceedings of the AAAI Conference on Artificial Intelligence},
  volume    = {30},
  number    = {1},
  year      = {2016},
  url       = {https://ojs.aaai.org/index.php/AAAI/article/view/10295}
}

@incollection{pytorch2019,
title = {PyTorch: An Imperative Style, High-Performance Deep Learning Library},
author = {Paszke, Adam and Gross, Sam and Massa, Francisco and Lerer, Adam and Bradbury, James and Chanan, Gregory and Killeen, Trevor and Lin, Zeming and Gimelshein, Natalia and Antiga, Luca and Desmaison, Alban and Kopf, Andreas and Yang, Edward and DeVito, Zachary and Raison, Martin and Tejani, Alykhan and Chilamkurthy, Sasank and Steiner, Benoit and Fang, Lu and Bai, Junjie and Chintala, Soumith},
booktitle = {Advances in Neural Information Processing Systems 32},
editor = {H. Wallach and H. Larochelle and A. Beygelzimer and F. d\textquotesingle Alch\'{e}-Buc and E. Fox and R. Garnett},
pages = {8024--8035},
year = {2019},
publisher = {Curran Associates, Inc.},
url = {http://papers.neurips.cc/paper/9015-pytorch-an-imperative-style-high-performance-deep-learning-library.pdf}
}

@article{wilcoxon1945individual,
  title={Individual comparisons by ranking methods},
  author={Wilcoxon, Frank},
  journal={Biometrics Bulletin},
  volume={1},
  number={6},
  pages={80--83},
  year={1945},
  publisher={JSTOR}
}

@article{Osband2016Bootstrapped,
  title={Deep Exploration via Bootstrapped DQN},
  author={Osband, Ian and Blundell, Charles and Pritzel, Alexander and Van Roy, Benjamin},
  journal={arXiv preprint arXiv:1602.04621},
  year={2016}
}

@article{Farebrother2024StopRegressing,
  title={Stop Regressing: Training Value Functions via Classification for Scalable Deep RL},
  author={Farebrother, Jesse and Orbay, Jordi and Vuong, Quan and Ta{\"i}ga, Adrien Ali and Chebotar, Yevgen and Xiao, Ted and Irpan, Alex and Levine, Sergey and Castro, Pablo Samuel and Faust, Aleksandra and Kumar, Aviral and Agarwal, Rishabh},
  journal={arXiv preprint arXiv:2403.03950},
  year={2024}
}

@misc{dosovitskiy2021images,
      title={An Image is Worth 16x16 Words: Transformers for Image Recognition at Scale}, 
      author={Alexey Dosovitskiy and Lucas Beyer and Alexander Kolesnikov and Dirk Weissenborn and Xiaohua Zhai and Thomas Unterthiner and Mostafa Dehghani and Matthias Minderer and Georg Heigold and Sylvain Gelly and Jakob Uszkoreit and Neil Houlsby},
      year={2021},
      eprint={2010.11929},
      archivePrefix={arXiv},
      primaryClass={cs.CV},
      url={https://arxiv.org/abs/2010.11929}, 
}

@article{cobbe2019procgen,
  title={Leveraging Procedural Generation to Benchmark Reinforcement Learning},
  author={Cobbe, Karl and Hesse, Christopher and Hilton, Jacob and Schulman, John},
  journal={arXiv preprint arXiv:1912.01588},
  year={2019}
}

\vspace{2em}

\noindent
\textbf{TAHA SHIEENAVAZ} received an M.Sc. in Computer Engineering from the
University of Padua. His research interests include deep learning, deep
reinforcement learning, and implicit neural representations. His work focuses
on the development and evaluation of machine-learning methods, with an
emphasis on reproducible experimental frameworks and open-source research
software.

\vspace{1em}

\noindent
\textbf{SHABNAM ZARESHAHRAKI} received an M.Sc. in Computer Engineering from
the University of Padua. She is currently completing a Master Valorization
internship in the La Manno Lab at the École polytechnique fédérale de Lausanne
(EPFL). Her research interests include computational biology, machine
learning, deep learning, and the development of computational methods for
biological applications.

\vspace{1em}

\noindent
\textbf{LORIS NANNI} is an Associate Professor in the Department of
Information Engineering at the University of Padua. His research interests
include pattern recognition, machine learning, biometric systems, image
analysis, and bioinformatics. His work spans the development and evaluation
of computational methods for classification, feature representation, and
biomedical data analysis.



\section*{Supplementary material}

**Supplementary Material for**

# A Progressive Design Study of Visual Encoders and Value Estimation for Replay-Free Parallelized Q-Learning

Taha Shieenavaz[a,*], Shabnam Zareshahraki[a,1], Loris Nanni[a]

[a] Department of Information Engineering, University of Padua, Via Gradenigo 6/B, 35131 Padova, Italy

[1] Present address: La Manno Lab, École polytechnique fédérale de Lausanne (EPFL), Lausanne, Switzerland

[*] Corresponding author: Taha Shieenavaz
Email: tahashieenavaz@gmail.com

---

**Document description.** This Supplementary Material provides detailed architecture specifications, implementation mappings, computational-complexity calculations, reproducibility settings, statistical-analysis outputs, Atari-57 environment-level results, learning curves, receptive-field analyses, and Procgen Hard terminal and learning-trajectory results supporting the main manuscript.

## S1. Detailed Architecture and Implementation Specifications

### *S1.1. Configuration-to-Code Mapping*

Table S1 maps the configuration names used throughout the manuscript to the corresponding encoder and network identifiers used in the released PyTorch implementation. This mapping is provided to make the experimental configurations reported in the paper directly traceable to the public codebase.

The implementation retains the legacy term `distributional` in several network identifiers. In the manuscript and Supplementary Material, however, these configurations are described as *categorical* because they use HL-Gauss regression of a scalar bootstrapped $\lambda$-return rather than the categorical Bellman-distribution projection used by C51.

Table S1: **Mapping between manuscript configurations and released implementation identifiers.** The terms in the *Encoder key* and *Network key* columns correspond to identifiers used by the public implementation. The legacy `distributional` network names refer to HL-Gauss categorical regression in this work.

| **Configuration** | **Encoder key** | **Network key** | **Experimental role** |
|---|---|---|---|
| PQN | `nature` | `q` | Scalar PQN reference |
| Alpha–Theta | variant-specific encoder key | `q` | Phase 1 encoder comparison |
| Hadamax Reference | `hadamax` | `q` | Phase 2 reference |
| Gamma-Hadamax-Valid | `gamma-hadamax-valid` | `q` | Phase 2 selected representation |
| Gamma-Hadamax-Same | `gamma-hadamax-same` | `q` | Phase 2 spatial-preservation variant |
| Categorical Dueling | `gamma-hadamax-valid` | `distributional-dueling` | Phase 3 value-estimation configuration |
| Ensemble Dueling | `gamma-hadamax-valid` | `bootstrapped-dueling` | Phase 3 value-estimation configuration |
| Aftab | `gamma-hadamax-valid` | `distributional-bootstrapped-dueling` | Final selected configuration |

All configurations reported in the controlled comparisons are generated from our PyTorch implementation. The PQN reference uses the Nature-DQN encoder and scalar value head, whereas the Hadamax reference is an independent PyTorch reimplementation of the three-stage Hadamax architecture. Published PQN and Hadamax results are used only as external reproduction references; model selection and statistical analysis use the independently generated runs reported in this study.

The architectural definitions corresponding to the identifiers above are given in the following supplementary sections. Phase 1 encoder specifications are pro-

vided in Section S1.2, Hadamax configurations in Section S1.3, and downstream value-estimation architectures in Section S1.4.

*S1.2. Phase 1 Encoder Architectures*

This section provides the complete layer-by-layer definitions of the convolutional encoders evaluated in Phase 1. These specifications correspond to the encoder comparison summarized in the main manuscript.

All Atari encoders receive observations of shape $4 \times 84 \times 84$. Each conventional convolutional stage consists of a convolution followed by channel-wise `LayerNorm2d` and ReLU.

For an activation tensor $x \in \mathbb{R}^{B\times C\times H\times W}$, the custom `LayerNorm2d` operation normalizes the channel dimension independently at every spatial position. The channel-wise mean is

$$\mu_{b,h,w} = \frac{1}{C}\sum_{c=1}^{C} x_{b,c,h,w}, \tag{S1}$$

and the corresponding variance is

$$\sigma^2_{b,h,w} = \frac{1}{C}\sum_{c=1}^{C} \left(x_{b,c,h,w} - \mu_{b,h,w}\right)^2. \tag{S2}$$

The normalized activation for channel $c$ is then

$$\hat{x}_{b,c,h,w} = \gamma_c \frac{x_{b,c,h,w} - \mu_{b,h,w}}{\sqrt{\sigma^2_{b,h,w} + \epsilon_{\mathrm{LN}}}} + \beta_c, \tag{S3}$$

where $\gamma_c$ and $\beta_c$ are learned per-channel affine parameters and $\epsilon_{\mathrm{LN}} = 10^{-6}$.

Each convolutional layer in Table S2 is represented as

$$(C_{\mathrm{in}}, C_{\mathrm{out}}, k, s, p),$$

where $C_{\mathrm{in}}$ and $C_{\mathrm{out}}$ are the input and output channel counts, $k$ is the kernel size, $s$ is the stride, and $p$ is the symmetric zero-padding width.

The candidates were designed to explore different patterns of spatial reduction and channel allocation rather than simply increasing parameter count. In particular, architectures with similar convolutional parameter counts can produce very different flattened feature dimensions and therefore different downstream value-head sizes.

Table S2: **Complete Phase 1 convolutional encoder specifications.** Each convolution is represented as $(C_{\text{in}}, C_{\text{out}}, k, s, p)$ and is followed by channel-wise `LayerNorm2d` and ReLU. A dash indicates that the corresponding layer is absent.

| **Variant** | **Layer 1** | **Layer 2** | **Layer 3** | **Layer 4** | **Layer 5** |
|---|---|---|---|---|---|
| PQN | $(4, 32, 8, 4, 0)$ | $(32, 64, 4, 2, 0)$ | $(64, 64, 3, 1, 0)$ | – | – |
| Alpha | $(4, 32, 4, 2, 1)$ | $(32, 64, 4, 2, 1)$ | $(64, 64, 3, 2, 1)$ | $(64, 64, 5, 1, 0)$ | – |
| Beta | $(4, 32, 6, 2, 2)$ | $(32, 64, 3, 1, 1)$ | $(64, 32, 4, 2, 1)$ | $(32, 16, 8, 1, 0)$ | – |
| Gamma | $(4, 32, 3, 2, 1)$ | $(32, 48, 3, 2, 1)$ | $(48, 64, 3, 1, 0)$ | $(64, 64, 3, 2, 0)$ | $(64, 64, 3, 1, 0)$ |
| Delta | $(4, 24, 9, 4, 0)$ | $(24, 48, 5, 2, 0)$ | $(48, 96, 3, 1, 0)$ | – | – |
| Epsilon | $(4, 32, 3, 2, 1)$ | $(32, 48, 3, 2, 1)$ | $(48, 64, 3, 2, 0)$ | $(64, 64, 3, 1, 0)$ | – |
| Zeta | $(4, 48, 4, 2, 1)$ | $(48, 48, 4, 2, 1)$ | $(48, 48, 4, 2, 1)$ | – | – |
| Eta | $(4, 64, 4, 4, 0)$ | $(64, 128, 3, 1, 0)$ | – | – | – |
| Theta | $(4, 32, 7, 4, 2)$ | $(32, 64, 5, 2, 1)$ | $(64, 32, 3, 1, 0)$ | – | – |

The corresponding encoder complexity values, including final feature-map dimensions, flattened representation sizes, parameter counts, and convolutional MACs, are reported in the main manuscript. Detailed derivations of complete-model parameter and MAC counts are provided in Section S1.5.

*S1.3. Phase 2 Hadamax Architectures*

This section provides the complete implementation details of the Hadamax configurations evaluated in Phase 2. The experiments compare an independently implemented three-stage Hadamax reference with two five-stage variants derived from the Gamma topology selected in Phase 1.

*S1.3.1. Hadamax Block Implementation*

The `HadamaxBlock` is implemented using a single bias-free convolution with $2C_{\text{out}}$ output channels. These channels represent two independently parameterized filter banks. For an input tensor $x$, the convolution produces

$$z = \text{Conv}(x) \in \mathbb{R}^{B \times 2C_{\text{out}} \times H \times W}. \tag{S4}$$

The output is reshaped as

$$z \to \mathbb{R}^{B \times 2 \times C_{\text{out}} \times H \times W}, \tag{S5}$$

yielding two feature branches, $z^{(1)}$ and $z^{(2)}$.

Normalization is applied independently to each branch over the $C_{\text{out}}$ channel dimension at every spatial location. Learned per-branch, per-channel affine parameters are retained. After normalization, both branches are activated using GELU,

$$u^{(1)} = \text{GELU}\left(\text{Norm}\left(z^{(1)}\right)\right), \qquad u^{(2)} = \text{GELU}\left(\text{Norm}\left(z^{(2)}\right)\right). \tag{S6}$$

The Hadamax representation is then obtained by element-wise multiplication,

$$h = u^{(1)} \odot u^{(2)}, \tag{S7}$$

followed by the max-pooling operation associated with the corresponding block.

Thus, although each block is described in terms of $C_{\text{out}}$ output channels, the internal convolution produces $2C_{\text{out}}$ channels before the two branches are separated.

*S1.3.2. Evaluated Phase 2 Configurations*

The Hadamax reference follows the three-stage architecture introduced by Kooi et al. [1], reimplemented in PyTorch for the common $4\times84\times84$ Atari observation protocol used in this study. The implementation uses explicit symmetric PyTorch padding rather than JAX-style `SAME` padding.

The two Gamma-Hadamax variants retain the five-stage channel hierarchy of Gamma, but replace its strided convolutions with stride-one Hadamax blocks followed by explicit max-pooling.

Gamma-Hadamax-Valid uses unpadded $3 \times 3$ pooling in Blocks 3 and 5. Its stagewise spatial resolutions are

$$84 \to 42 \to 21 \to 19 \to 9 \to 7. \tag{S8}$$

Gamma-Hadamax-Same differs only by using one pixel of pooling padding in Blocks 3 and 5, producing

$$84 \to 42 \to 21 \to 21 \to 10 \to 10. \tag{S9}$$

The complete block specifications are given in Table S3. Convolutions are represented as

$$(C_{\text{in}}, C_{\text{out}}, k, s, p),$$

where $C_{\text{out}}$ denotes the number of channels in *each* Hadamax branch. Max-pooling operations are represented as

$$[k, s, p].$$

The two Gamma-Hadamax variants consequently have identical convolutional parameter counts but produce different final spatial representations. Gamma-

Table S3: **Complete Phase 2 Hadamax architecture specifications.** Convolutions are represented as $(C_{\text{in}}, C_{\text{out}}, k, s, p)$, where $C_{\text{out}}$ is the channel count of each Hadamax branch. The internal bias-free convolution therefore produces $2C_{\text{out}}$ channels before the branch split. Max-pooling operations are represented as $[k, s, p]$.

| **Variant** | **Block 1** | **Block 2** | **Block 3** | **Block 4** | **Block 5** |
|---|---|---|---|---|---|
| **Hadamax Reference** | | | | | |
| Convolution | $(4, 32, 8, 1, 4)$ | $(32, 64, 4, 1, 2)$ | $(64, 64, 3, 1, 1)$ | – | – |
| Max Pool | $[4, 4, 0]$ | $[2, 2, 0]$ | $[3, 1, 1]$ | – | – |
| **Gamma-Hadamax-Valid** | | | | | |
| Convolution | $(4, 32, 3, 1, 1)$ | $(32, 48, 3, 1, 1)$ | $(48, 64, 3, 1, 1)$ | $(64, 64, 3, 1, 1)$ | $(64, 64, 3, 1, 1)$ |
| Max Pool | $[2, 2, 0]$ | $[2, 2, 0]$ | $[3, 1, 0]$ | $[2, 2, 0]$ | $[3, 1, 0]$ |
| **Gamma-Hadamax-Same** | | | | | |
| Convolution | $(4, 32, 3, 1, 1)$ | $(32, 48, 3, 1, 1)$ | $(48, 64, 3, 1, 1)$ | $(64, 64, 3, 1, 1)$ | $(64, 64, 3, 1, 1)$ |
| Max Pool | $[2, 2, 0]$ | $[2, 2, 0]$ | $[3, 1, 1]$ | $[2, 2, 0]$ | $[3, 1, 1]$ |

Hadamax-Valid produces a $64 \times 7 \times 7$ feature map ($F = 3136$), whereas Gamma-Hadamax-Same produces a $64 \times 10 \times 10$ feature map ($F = 6400$). The larger flattened representation of Gamma-Hadamax-Same directly increases the size of the downstream fully connected value head.

For comparison, the independently implemented Hadamax reference produces a $64 \times 11 \times 11$ final feature map ($F = 7744$). Complete encoder and full-model parameter/MAC calculations for these configurations are provided in Section S1.5.

*S1.4. Value-Estimation Architecture Details*

This section provides the complete downstream value-estimation architectures used throughout the study. Phase 1 and Phase 2 use a scalar PQN-style value head, whereas Phase 3 compares Categorical Dueling, Ensemble Dueling, and Aftab while holding the Gamma-Hadamax-Valid visual representation fixed.

Let $F$ denote the flattened encoder dimension and $A = |\mathcal{A}|$ the number of actions in the current environment. For Gamma-Hadamax-Valid, $F = 3136$. All hidden projections use dimension $D = 512$.

*S1.4.1. Scalar PQN Value Head*

The scalar configurations used by PQN, all Phase 1 encoder variants, the Phase 2 Hadamax variants, and the scalar Gamma-Hadamax-Valid reference use

$$F \to \mathrm{Linear}(F, 512) \to \mathrm{LayerNorm}(512) \to \mathrm{ReLU} \to \mathrm{Linear}(512, A).$$

The network outputs one scalar action value for each action. For the action selected at transition $t$, training minimizes half squared error against the corresponding bootstrapped $\lambda$-return,

$$\mathcal{L}_{\mathrm{scalar}} = \frac{1}{N}\sum_{j=1}^{N}\frac{1}{2}\left[Q(s_j, a_j) - R_j^{\lambda}\right]^2, \tag{S10}$$

where $N$ denotes the mini-batch size. The same scalar-head architecture is used across the corresponding controlled comparisons; differences in complete model size therefore arise from the encoder representation dimension $F$ and, to a smaller extent, the environment-specific action count $A$.

*S1.4.2. Categorical Dueling*

Categorical Dueling combines the dueling architecture [2] with HL-Gauss categorical regression of the scalar bootstrapped $\lambda$-return.

The value stream is

$$F \to \text{Linear}(F, 512) \to \text{LayerNorm}(512) \to \text{ReLU} \to \text{Linear}(512, 51),$$

while the advantage stream is

$$F \to \text{Linear}(F, 512) \to \text{LayerNorm}(512) \to \text{ReLU} \to \text{Linear}(512, A \times 51).$$

Let $\ell_i^V(s)$ denote the value-stream logit for categorical bin $i$ and $\ell_i^A(s,a)$ the corresponding advantage-stream logit for action $a$. Dueling combination is performed directly at the logit level:

$$\ell_i(s,a) = \ell_i^V(s) + \left[ \ell_i^A(s,a) - \frac{1}{A} \sum_{a'=1}^{A} \ell_i^A(s,a') \right]. \tag{S11}$$

The categorical support contains 51 bins over $[-10, 10]$. When no explicit Gaussian width is supplied, the implementation derives

$$\sigma = 0.75 \left( \frac{10 - (-10)}{51} \right) \approx 0.2941. \tag{S12}$$

The `HLGaussLoss` object is instantiated with

`clamp_to_range=True`,

so scalar targets outside the configured support are handled using the library's range-clamping behavior before construction of the categorical target. The framework's optional categorical value-change clipping is disabled in all reported experiments,

`distributional_value_clip` $= 0.$

For each collected transition, the scalar $\lambda$-return is transformed into an HL-Gauss target probability vector. Cross-entropy is minimized between this target and the predicted logits for the action that was actually taken.

For behavior and bootstrapping, the categorical prediction is decoded back to a scalar action value using the HL-Gauss decoder. Thus, the categorical

configuration performs regression-as-classification of a scalar $\lambda$-return target. It does not perform the categorical Bellman distribution projection used by C51.

### *S1.4.3. Ensemble Dueling*

Ensemble Dueling uses $K = 10$ independently parameterized scalar dueling heads above the shared Gamma-Hadamax-Valid encoder. For each head $k$, the value stream is

$$F \to \mathrm{Linear}(F, 512) \to \mathrm{LayerNorm}(512) \to \mathrm{ReLU} \to \mathrm{Linear}(512, 1),$$

and the advantage stream is

$$F \to \mathrm{Linear}(F, 512) \to \mathrm{LayerNorm}(512) \to \mathrm{ReLU} \to \mathrm{Linear}(512, A).$$

For head $k$, scalar action values are combined using the mean-centered dueling decomposition,

$$Q_k(s, a) = V_k(s) + A_k(s, a) - \frac{1}{A} \sum_{a'=1}^{A} A_k(s, a'). \tag{S13}$$

*Head-conditioned exploration.* At the beginning of training, one head index is sampled uniformly for each parallel environment. Training actions for that environment are selected $\epsilon$-greedily according to the currently active head. The same head remains active until the environment reports termination or truncation, at which point a new head is sampled uniformly. This produces temporally extended head-conditioned behavior within each parallel trajectory.

*Training-data assignment and head-specific targets.* The implementation is inspired by Bootstrapped DQN [3], but the training-data assignment differs from the classical bootstrap formulation. The bootstrap inclusion probability is

$$p = 1.$$

Consequently, the bootstrap mask is identically one and all ten heads are optimized on every transition in the synchronous PQN mini-batch.

Despite receiving the same transitions, the heads maintain independent parameters and head-specific bootstrapping targets. For head $k$, the $\lambda$-return is recursively constructed using that head's own next-state action-value estimate,

$$\max_a Q_k(s_{t+1}, a).$$

For a mini-batch of size $N$, the ensemble objective is

$$\mathcal{L}_{\text{ens}} = \frac{1}{NK} \sum_{j=1}^{N} \sum_{k=1}^{K} \frac{1}{2} \left[ Q_k(s_j, a_j) - R^{\lambda}_{j,k} \right]^2 . \tag{S14}$$

Because $p = 1$, this configuration should not be interpreted as a classical bootstrap estimator based on different resampled datasets or as an implementation of posterior sampling.

*Evaluation.* Evaluation does not select one fixed head and does not average Q-values across heads. Instead, each head independently selects its greedy action,

$$a_k^* = \arg\max_a Q_k(s, a), \tag{S15}$$

and the final evaluation action is the action receiving the largest number of head-level votes. Thus, multi-head evaluation uses majority voting over the ten greedy actions.

#### *S1.4.4. Aftab*

Aftab combines the $K = 10$ ensemble structure with categorical dueling value estimation above the shared Gamma-Hadamax-Valid encoder. Each head contains independently parameterized categorical value and advantage streams.

Unlike the ReLU-activated streams used by Categorical Dueling and Ensemble Dueling, the final Aftab configuration uses GELU.

For each head, the value stream is

$$F \to \text{Linear}(F, 512) \to \text{LayerNorm}(512) \to \text{GELU} \to \text{Linear}(512, 51),$$

and the advantage stream is

$$F \to \text{Linear}(F, 512) \to \text{LayerNorm}(512) \to \text{GELU} \to \text{Linear}(512, A \times 51).$$

Within each head, categorical value and advantage logits are combined using the same mean-centered dueling operation as Eq. S11. The resulting categorical predictions are decoded to scalar action values for behavior and bootstrapping.

*Training.* Aftab uses the same head-conditioned exploration rule as Ensemble Dueling: one head is sampled independently for each parallel environment and retained until termination or truncation. Training actions are selected $\epsilon$-greedily according to that environment's active head.

All ten heads receive every collected transition because $p = 1$. However, each head constructs its own scalar $\lambda$-return from its own bootstrapped action values. For head $k$, this produces a target $R^{\lambda}_{j,k}$ for transition $j$. Each scalar target is then converted independently into an HL-Gauss categorical target.

Let $\mathbf{p}_{j,k}$ denote the HL-Gauss target vector and $\boldsymbol{\ell}_{j,k}(a_j)$ the predicted logits for the action taken. The categorical ensemble objective can be written as

$$\mathcal{L}_{\text{Aftab}} = \frac{1}{NK} \sum_{j=1}^{N} \sum_{k=1}^{K} \text{CE}\left(\mathbf{p}_{j,k}, \boldsymbol{\ell}_{j,k}(a_j)\right), \tag{S16}$$

where CE denotes categorical cross-entropy.

*Evaluation.* During evaluation, each head decodes its categorical prediction to scalar action values and independently selects its greedy action. As in Ensemble Dueling, the final environment action is determined by majority vote across the ten head-level greedy actions.

#### *S1.4.5. Interpretation of the Phase 3 Comparison*

The Phase 3 configurations should be interpreted as complete implemented architectures rather than as a component-wise ablation. In particular:

- Categorical Dueling changes both the output representation and the training objective relative to the scalar reference;
- Ensemble Dueling introduces ten independently parameterized dueling heads, head-conditioned behavior, and head-specific bootstrapping targets;
- Aftab combines categorical regression, dueling decomposition, multi-head estimation, head-conditioned exploration, and GELU-activated streams.

Accordingly, the observed differences among these configurations do not identify the independent causal contribution of categorical regression, dueling decomposition, ensembling, or activation choice.

Detailed parameter and forward-MAC calculations for each value-estimation configuration are provided in Section S1.5.

*S1.5. Model Complexity Derivations*

This section provides the parameter-count and multiply-accumulate-operation (MAC) derivations underlying the complexity values reported in the main manuscript.

Unless otherwise stated, parameter counts include all trainable weights, biases, and normalization affine parameters. MAC counts include convolutional and linear multiply-accumulates. They exclude normalization operations, activation functions, bias additions, pooling, Hadamard products, softmax, categorical decoding, and loss computation.

Let

$$F = \text{flattened encoder dimension}, \qquad D = 512, \qquad A = |\mathcal{A}|,$$

where $A$ is the environment-specific action count. For categorical configurations, let

$$C = 51$$

denote the number of categorical bins, and for ensemble configurations let

$$K = 10$$

denote the number of independently parameterized heads.

*S1.5.1. Scalar Value Head*

The scalar value head used by PQN, all Phase 1 configurations, the Phase 2 scalar configurations, and the scalar Gamma-Hadamax-Valid reference is

$$F \to \text{Linear}(F, D) \to \text{LayerNorm}(D) \to \text{ReLU} \to \text{Linear}(D, A).$$

The first linear layer contains

$$FD + D$$

trainable parameters. Layer normalization contributes a learned scale and bias for each hidden feature,

$$2D,$$

and the output layer contributes

$$DA + A.$$

The total scalar-head parameter count is therefore

$$P_{\text{scalar}}(F, A) = FD + 3D + DA + A. \tag{S17}$$

For $D = 512$,

$$P_{\text{scalar}}(F, A) = 512F + 1536 + 513A. \tag{S18}$$

Under the MAC-counting convention used throughout the study, only the two linear matrix multiplications contribute to the head MAC count:

$$M_{\text{scalar}}(F, A) = FD + DA. \tag{S19}$$

Thus,

$$M_{\text{scalar}}(F, A) = 512F + 512A. \tag{S20}$$

If an encoder contains $P_{\text{enc}}$ trainable parameters and requires $M_{\text{enc}}$ MACs, the corresponding complete scalar model contains

$$P_{\text{total}} = P_{\text{enc}} + P_{\text{scalar}}(F, A), \tag{S21}$$

and

$$M_{\text{total}} = M_{\text{enc}} + M_{\text{scalar}}(F, A). \tag{S22}$$

This dependence on $F$ explains why encoder parameter count alone can be a poor proxy for complete-model complexity. An encoder that preserves a large

spatial representation can substantially enlarge the first fully connected layer even when its convolutional parameter count is modest.

#### *S1.5.2. Representative Phase 1 Complete-Model Costs*

Table S4 illustrates the complete scalar-model complexity for an Atari environment with $A = 8$ actions. The parameter counts are obtained exactly from Eq. S21. The MAC totals use the encoder MAC values reported in the main manuscript, which are rounded to three decimal places, and are therefore shown approximately.

Table S4: **Representative complete-model complexity for the Phase 1 encoders with $A = 8$ actions.** Parameter counts include the encoder and scalar value head. Forward MACs include convolutional and linear multiply-accumulates.

| **Variant** | $F$ | **Total parameters** | **Forward MACs (M)** |
|---|---:|---:|---:|
| PQN | 3,136 | 1,689,576 | $\approx 9.347$ |
| Alpha | 3,136 | 1,786,024 | $\approx 29.152$ |
| Beta | 3,136 | 1,700,280 | $\approx 63.126$ |
| Gamma | 3,136 | 1,728,440 | $\approx 24.512$ |
| Delta | 3,456 | 1,853,664 | $\approx 7.917$ |
| Epsilon | 4,096 | 2,182,904 | $\approx 15.354$ |
| Zeta | 4,800 | 2,540,472 | $\approx 27.824$ |
| Eta | 46,208 | 23,742,536 | $\approx 52.085$ |
| Theta | 2,048 | 1,130,504 | $\approx 10.119$ |

The Eta configuration illustrates the effect most clearly. Although its encoder contains relatively few convolutional parameters, its 46,208-dimensional flattened representation makes the first fully connected layer substantially larger than in the other configurations.

#### *S1.5.3. Representative Phase 2 Complete-Model Costs*

The same scalar-head equations apply to the Phase 2 Hadamax configurations. For $A = 8$, their representative complete-model costs are shown in

Table S5.

Table S5: **Representative complete-model complexity for the Phase 2 Hadamax configurations with $A = 8$ actions.**

| Configuration | $F$ | Total parameters | Forward MACs (M) |
|---|---|---|---|
| Hadamax Reference | 7,744 | 4,126,856 | $\approx 162.984$ |
| Gamma-Hadamax-Valid | 3,136 | 1,845,064 | $\approx 123.611$ |
| Gamma-Hadamax-Same | 6,400 | 3,516,232 | $\approx 132.581$ |

Although Gamma-Hadamax-Valid and Gamma-Hadamax-Same have identical encoder parameter counts, their different flattened feature dimensions result in substantially different complete-model sizes.

*S1.5.4. Categorical Dueling Complexity*

For Categorical Dueling, the value stream contains

$$F \to D \to C,$$

while the advantage stream contains

$$F \to D \to AC.$$

Both streams include Layer Normalization after their first linear projection. The value-stream parameter count is

$$P_V = FD + 3D + DC + C, \tag{S23}$$

where $FD + D$ corresponds to the first linear layer, $2D$ to Layer Normalization, and $DC + C$ to the categorical output layer.

The advantage-stream parameter count is

$$P_A = FD + 3D + DAC + AC. \tag{S24}$$

Thus the complete categorical-dueling head contains

$$P_{\text{cat}} = 2FD + 6D + DC + C + AC(D + 1). \tag{S25}$$

For the fixed Gamma-Hadamax-Valid representation, $F = 3136$, $D = 512$, and $C = 51$. Including the encoder's 233,792 parameters gives

$$P_{\text{CatDuel}}(A) = 3{,}474{,}291 + 26{,}163A. \tag{S26}$$

The corresponding linear MAC count for the two streams is

$$M_{\text{cat}} = 2FD + DC + DAC. \tag{S27}$$

Including the fixed encoder cost of 122.001408 million convolutional MACs,

$$M_{\text{CatDuel}}(A) = 125.238784 + 0.026112A \quad \text{million MACs.} \tag{S28}$$

*S1.5.5. Ensemble Dueling Complexity*

Each scalar dueling head in Ensemble Dueling has an independent value stream

$$F \to D \to 1$$

and advantage stream

$$F \to D \to A,$$

with separate Layer Normalization in each stream.

The parameter count of one value stream is

$$FD + 4D + 1, \tag{S29}$$

and that of one advantage stream is

$$FD + 3D + DA + A. \tag{S30}$$

One complete scalar dueling head therefore contains

$$P_{\text{duel}} = 2FD + 7D + 1 + A(D + 1). \tag{S31}$$

With $K$ independently parameterized heads and a shared encoder,

$$P_{\text{EnsDuel}} = P_{\text{enc}} + K\left[2FD + 7D + 1 + A(D + 1)\right]. \tag{S32}$$

For $F = 3136$, $D = 512$, $K = 10$, and $P_{\text{enc}} = 233{,}792$,

$$P_{\text{EnsDuel}}(A) = 32{,}382{,}282 + 5{,}130A. \tag{S33}$$

The counted linear MACs of one scalar dueling head are

$$M_{\text{duel}} = 2FD + D + DA. \tag{S34}$$

Adding ten heads to the shared Gamma-Hadamax-Valid encoder yields

$$M_{\text{EnsDuel}}(A) = 154.119168 + 0.005120A \quad \text{million MACs}. \tag{S35}$$

*S1.5.6. Aftab Complexity*

Each Aftab head has the same output dimensionalities as the Categorical Dueling configuration: a categorical value stream with $C$ outputs and a categorical advantage stream with $AC$ outputs. Replacing ReLU by GELU does not alter the trainable parameter count or the linear MAC count under the counting convention used here.

The parameter count of one categorical dueling head is therefore

$$P_{\text{cat-duel}} = 2FD + 6D + DC + C + AC(D + 1). \tag{S36}$$

For $K$ independently parameterized heads above a shared encoder,

$$P_{\text{Aftab}} = P_{\text{enc}} + KP_{\text{cat-duel}}. \tag{S37}$$

With $F = 3136$, $D = 512$, $C = 51$, $K = 10$, and $P_{\text{enc}} = 233{,}792$,

$$P_{\text{Aftab}}(A) = 32{,}638{,}782 + 261{,}630A. \tag{S38}$$

The counted linear MACs for one categorical dueling head are

$$M_{\text{cat-duel}} = 2FD + DC + DAC. \tag{S39}$$

Including all ten heads and the shared encoder gives

$$M_{\text{Aftab}}(A) = 154.375168 + 0.261120A \quad \text{million MACs.} \tag{S40}$$

*S1.5.7. Phase 3 Complexity Summary*

Table S6 collects the resulting exact action-dependent expressions.

Table S6: **Phase 3 complete-model complexity.** All configurations use the shared Gamma-Hadamax-Valid encoder. MAC counts include convolutional and linear multiply-accumulates but exclude normalization, activations, Hadamard products, pooling, softmax, categorical decoding, and loss computation.

| **Configuration** | **Trainable parameters** | **Forward MACs (M)** |
|---|---|---|
| Scalar Gamma-Hadamax-Valid | $1{,}840{,}960 + 513A$ | $123.607040 + 0.000512A$ |
| Categorical Dueling | $3{,}474{,}291 + 26{,}163A$ | $125.238784 + 0.026112A$ |
| Ensemble Dueling ($K = 10$) | $32{,}382{,}282 + 5{,}130A$ | $154.119168 + 0.005120A$ |
| Aftab ($K = 10$, $C = 51$) | $32{,}638{,}782 + 261{,}630A$ | $154.375168 + 0.261120A$ |

For a representative Atari environment with $A = 8$ actions, the resulting complexities are shown in Table S7.

These calculations illustrate why parameter storage and arithmetic cost should be considered separately. The ten independently parameterized value-estimation heads substantially increase the number of trainable parameters in Ensemble Dueling and Aftab, whereas their effect on forward MACs is comparatively smaller because the computationally expensive visual encoder is shared.

Table S7: **Representative Phase 3 complexity for $A = 8$ actions.**

| **Configuration** | **Parameters** | **MACs (M)** |
|---|---:|---:|
| Scalar Gamma-Hadamax-Valid | 1,845,064 | 123.611 |
| Categorical Dueling | 3,683,595 | 125.448 |
| Ensemble Dueling | 32,423,322 | 154.160 |
| Aftab | 34,731,822 | 156.464 |

## S2. Receptive Field Analysis

The receptive field ($RF$) of a convolutional neural network at layer $l$ calculates the spatial region of the input volume that influences a single feature map activation. It is defined recursively based on the kernel size ($k$) and stride ($s$) of the current and preceding layers:

$$RF_l = RF_{l-1} + (k_l - 1)\prod_{i=1}^{l-1} s_i \tag{S41}$$

where we assume an initial receptive field $RF_0 = 1$ and an initial stride product of 1. Below, we detail the step-by-step calculation of the maximum theoretical receptive field for the Nature DQN, Hadamax, Gamma, and Gamma-Hadamax encoders.

| Encoder Architecture | Effective Receptive Field |
|---|---|
| Nature DQN Encoder | $36 \times 36$ |
| Gamma Encoder | $39 \times 39$ |
| Hadamax Encoder | $59 \times 59$ |
| Gamma-Hadamax Encoder (Valid and Same) | $70 \times 70$ |

Table S8: Summary of the maximum theoretical receptive fields for the Nature DQN, standard Gamma, baseline Hadamax, and Gamma-Hadamax encoders.

### *S2.1. Nature DQN Encoder*

The Nature DQN architecture consists of three convolutional layers with aggressive early spatial downsampling (a stride of 4 in the first layer). The receptive field expands as follows:

$$\begin{aligned}
\text{Layer 1 } (k=8, s=4): &\quad RF_1 = 1 + (8-1) \times 1 = 8 \\
\text{Layer 2 } (k=4, s=2): &\quad RF_2 = 8 + (4-1) \times 4 = 20 \\
\text{Layer 3 } (k=3, s=1): &\quad RF_3 = 20 + (3-1) \times (4 \times 2) = 36
\end{aligned}$$

The total effective receptive field for the Nature DQN encoder maps to a $36 \times 36$ spatial patch on the original input.

*S2.2. Gamma Encoder*

The Gamma architecture replaces large filters with a deeper, five-layer stack of $3 \times 3$ convolutions. This preserves spatial resolution longer before downsampling. The calculation is as follows:

$$
\begin{aligned}
\text{Layer 1 } (k=3, s=2): &\quad RF_1 = 1 + (3-1) \times 1 = 3 \\
\text{Layer 2 } (k=3, s=2): &\quad RF_2 = 3 + (3-1) \times 2 = 7 \\
\text{Layer 3 } (k=3, s=1): &\quad RF_3 = 7 + (3-1) \times (2 \times 2) = 15 \\
\text{Layer 4 } (k=3, s=2): &\quad RF_4 = 15 + (3-1) \times (2 \times 2 \times 1) = 23 \\
\text{Layer 5 } (k=3, s=1): &\quad RF_5 = 23 + (3-1) \times (2 \times 2 \times 1 \times 2) = 39
\end{aligned}
$$

The total effective receptive field for the Gamma encoder maps to a $39 \times 39$ spatial patch. This demonstrates that despite using exclusively small $3 \times 3$ kernels, the depth of the Gamma architecture allows it to achieve a marginally larger global receptive field than the Nature DQN baseline, while simultaneously supporting a highly complex, non-linear hypothesis space.

*S2.3. Hadamax Encoder*

The Hadamax variant of the Nature DQN architecture replaces standard convolutional strides with explicit Max Pooling steps for downsampling. Because each block contains both a convolution ($c$) and a pooling ($p$) operation, the receptive field expands in two stages per block:

$$
\begin{aligned}
\text{Block 1 Conv } (k=8, s=1): &\quad RF_{c1} = 1 + (8-1) \times 1 = 8 \\
\text{Block 1 Pool } (k=4, s=4): &\quad RF_{p1} = 8 + (4-1) \times 1 = 11 \\
\text{Block 2 Conv } (k=4, s=1): &\quad RF_{c2} = 11 + (4-1) \times (1 \times 4) = 23 \\
\text{Block 2 Pool } (k=2, s=2): &\quad RF_{p2} = 23 + (2-1) \times 4 = 27 \\
\text{Block 3 Conv } (k=3, s=1): &\quad RF_{c3} = 27 + (3-1) \times (4 \times 2) = 43 \\
\text{Block 3 Pool } (k=3, s=1): &\quad RF_{p3} = 43 + (3-1) \times 8 = 59
\end{aligned}
$$

The total effective receptive field for the Hadamax encoder maps to a $59 \times 59$ spatial patch.

*S2.4. Gamma-Hadamax Encoder (Valid & Same)*

The Gamma-Hadamax architecture utilizes a five-block deep configuration. Note that while Valid and Same differ in their pooling padding, padding does not alter the maximum receptive field calculation of a single unit. Thus, the calculation for both is identical:

$$\begin{aligned}
\text{Block 1 Conv } (k=3, s=1): \quad & RF_{c1} = 1 + (3-1) \times 1 = 3 \\
\text{Block 1 Pool } (k=2, s=2): \quad & RF_{p1} = 3 + (2-1) \times 1 = 4 \\
\text{Block 2 Conv } (k=3, s=1): \quad & RF_{c2} = 4 + (3-1) \times (1 \times 2) = 8 \\
\text{Block 2 Pool } (k=2, s=2): \quad & RF_{p2} = 8 + (2-1) \times 2 = 10 \\
\text{Block 3 Conv } (k=3, s=1): \quad & RF_{c3} = 10 + (3-1) \times (2 \times 2) = 18 \\
\text{Block 3 Pool } (k=3, s=1): \quad & RF_{p3} = 18 + (3-1) \times 4 = 26 \\
\text{Block 4 Conv } (k=3, s=1): \quad & RF_{c4} = 26 + (3-1) \times (4 \times 1) = 34 \\
\text{Block 4 Pool } (k=2, s=2): \quad & RF_{p4} = 34 + (2-1) \times 4 = 38 \\
\text{Block 5 Conv } (k=3, s=1): \quad & RF_{c5} = 38 + (3-1) \times (4 \times 2) = 54 \\
\text{Block 5 Pool } (k=3, s=1): \quad & RF_{p5} = 54 + (3-1) \times 8 = 70
\end{aligned}$$

The total effective receptive field for the Gamma-Hadamax encoder maps to a $70 \times 70$ spatial patch. This deeper, decoupled pooling approach substantially expands the receptive field compared to both the standard Gamma architecture ($39 \times 39$) and the baseline Hadamax ($59 \times 59$).

## S3. Individual Environment Learning Curves (Figures)

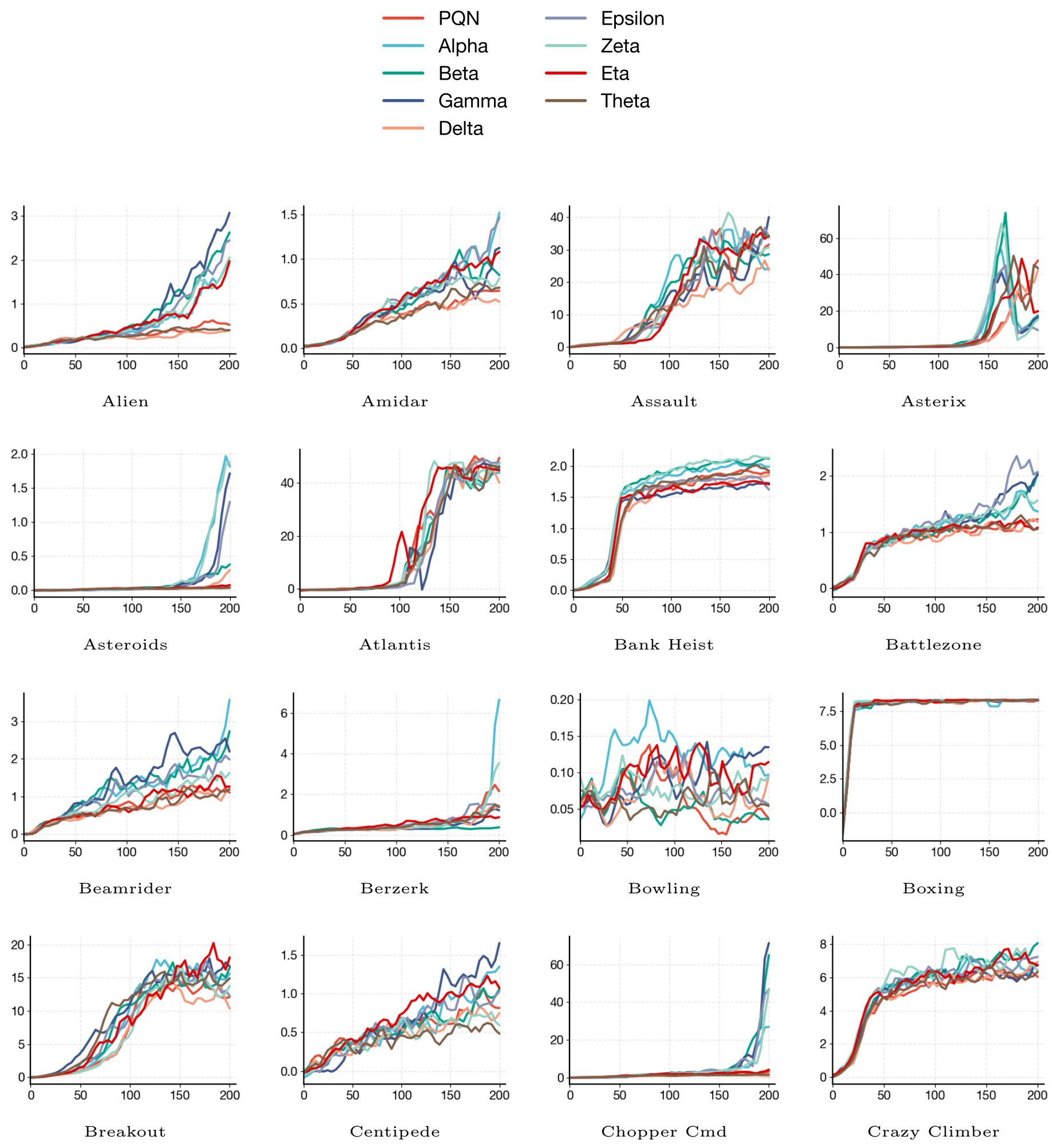


Figure S1: Phase 1 Learning Curves: Human-Normalized Score learning curves over 200 million training frames for the baseline CNN encoders (Alpha through Theta) across all 57 Atari games. (Part 1 of 4)

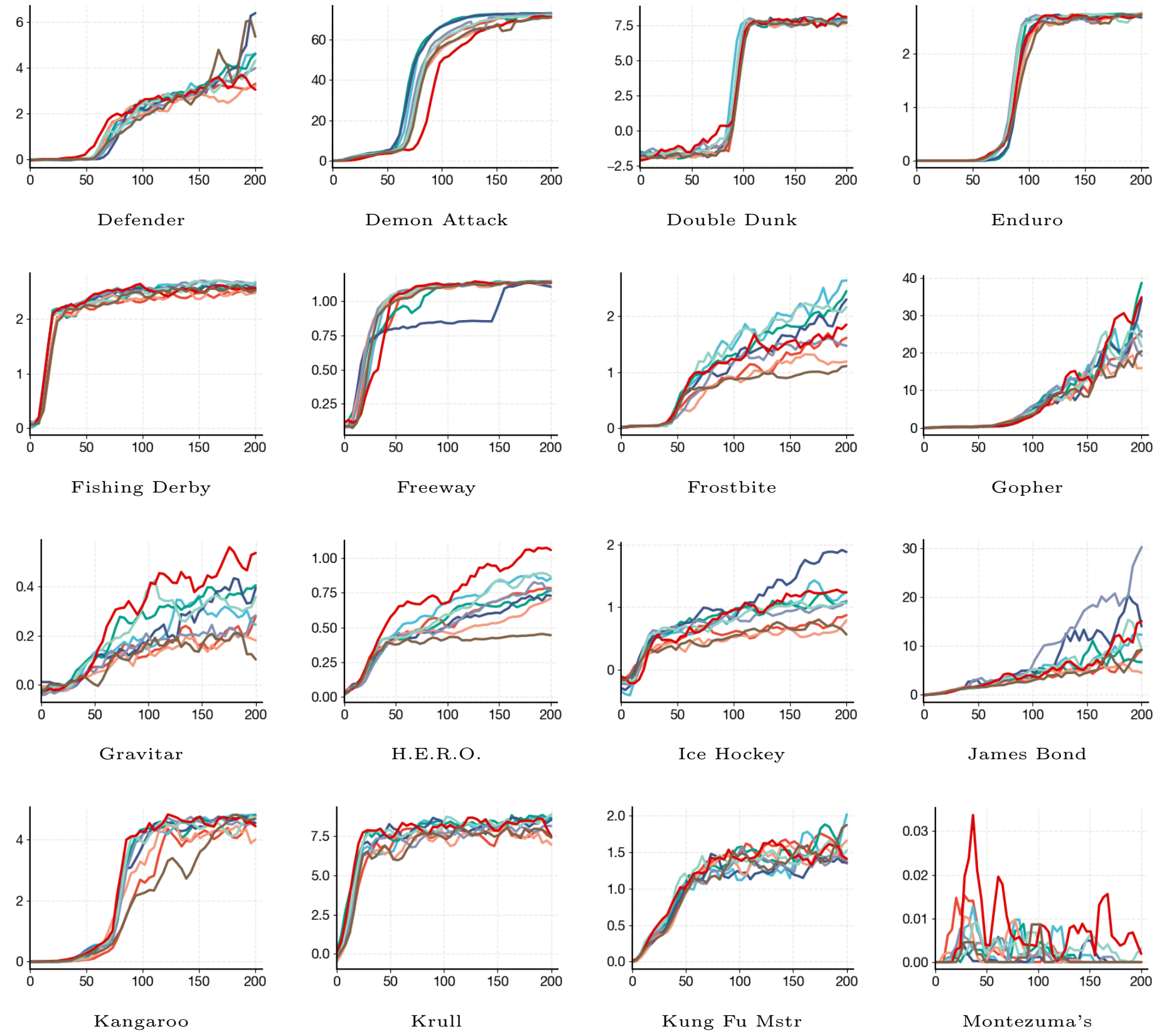


Figure S1: Phase 1 Learning Curves: Human-Normalized Score learning curves over 200 million training frames for the baseline CNN encoders (Alpha through Theta) across all 57 Atari games. (Part 2 of 4)

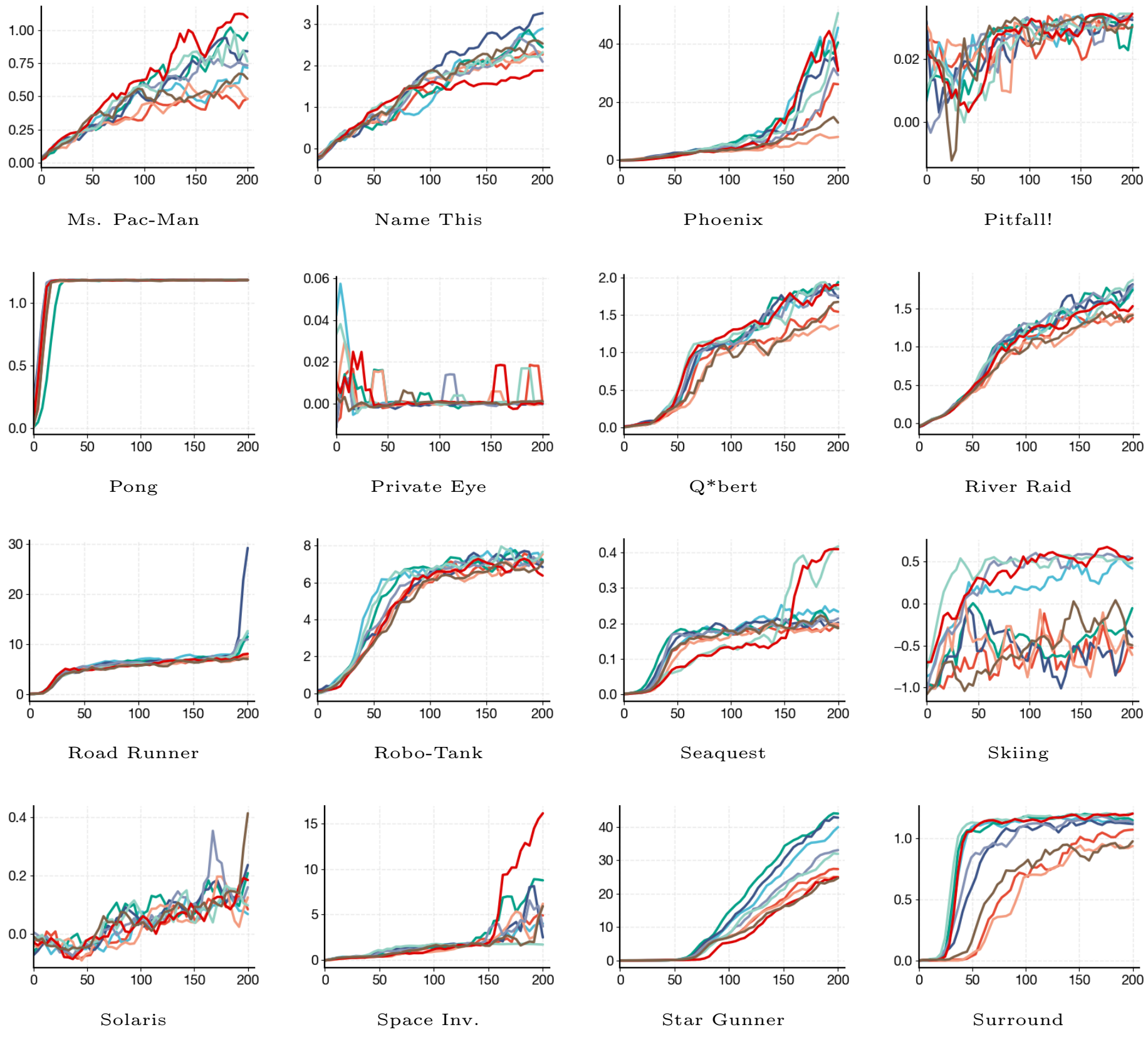


Figure S1: Phase 1 Learning Curves: Human-Normalized Score learning curves over 200 million training frames for the baseline CNN encoders (Alpha through Theta) across all 57 Atari games. (Part 3 of 4)

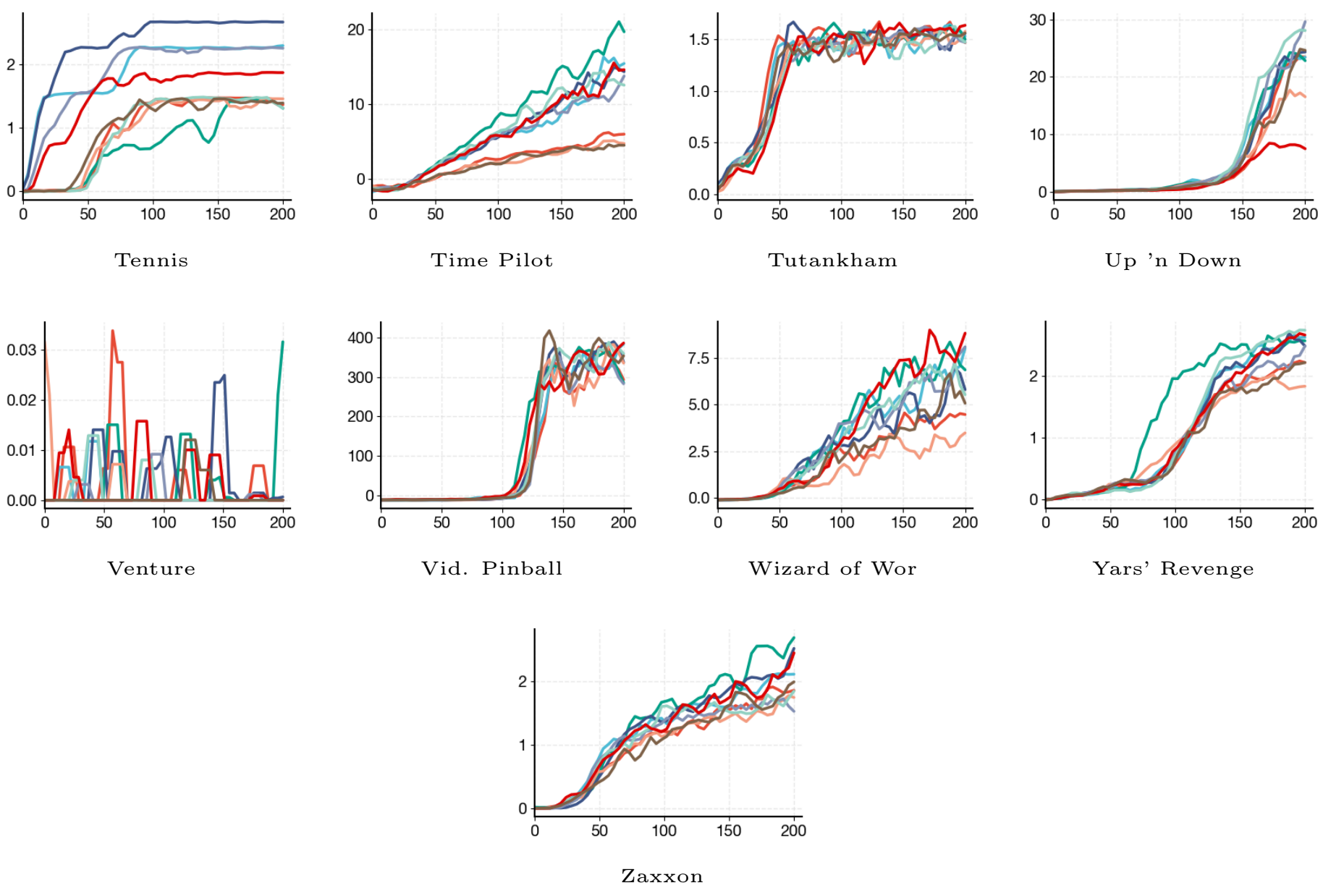


Figure S1: Phase 1 Learning Curves: Human-Normalized Score learning curves over 200 million training frames for the baseline CNN encoders (Alpha through Theta) across all 57 Atari games. (Part 4 of 4)

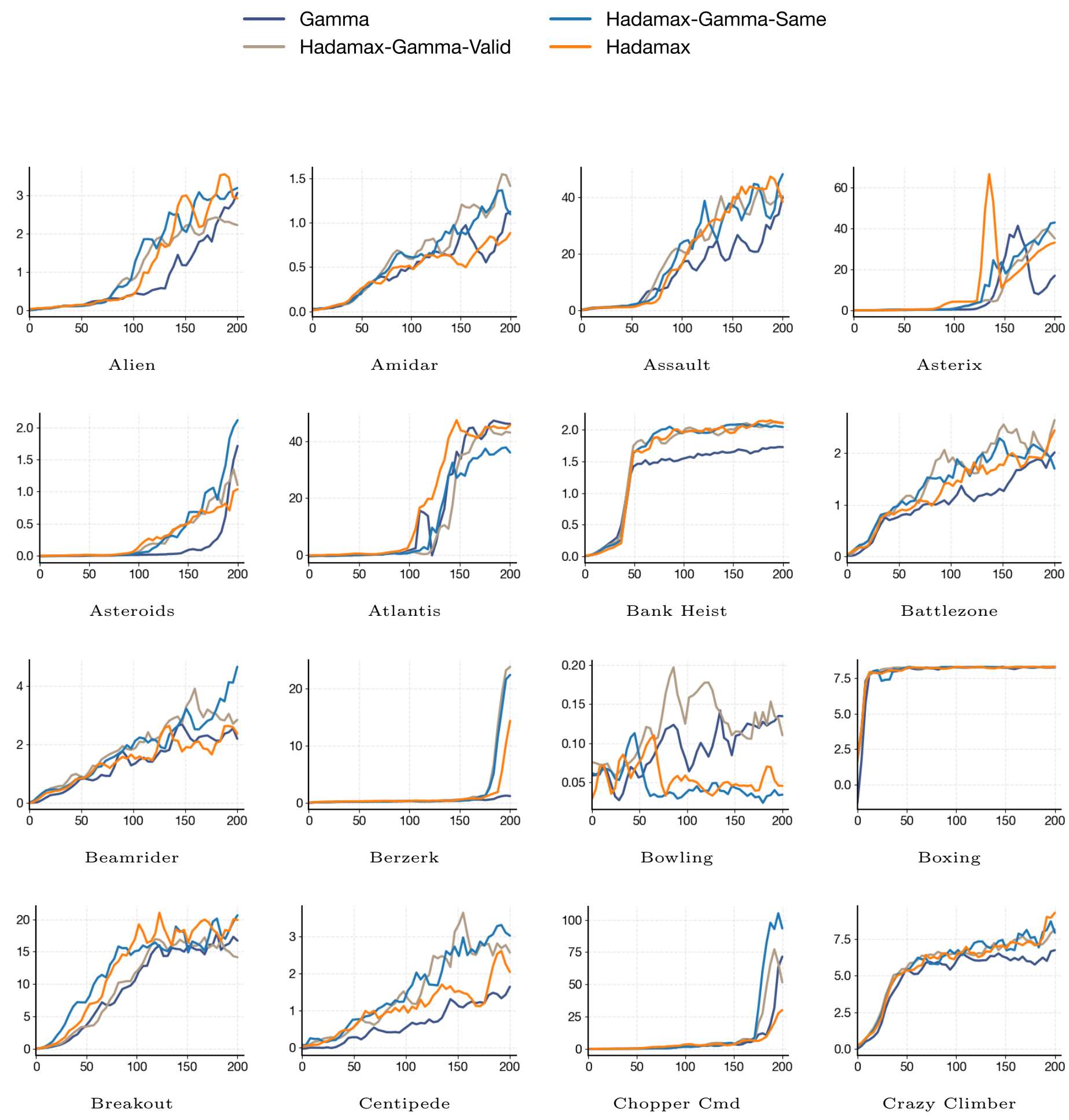


Figure S2: Phase 2 Learning Curves: Human-Normalized Score learning curves over 200 million training frames comparing the standard Gamma baseline to the Hadamax-integrated variants across all 57 Atari games. (Part 1 of 4)

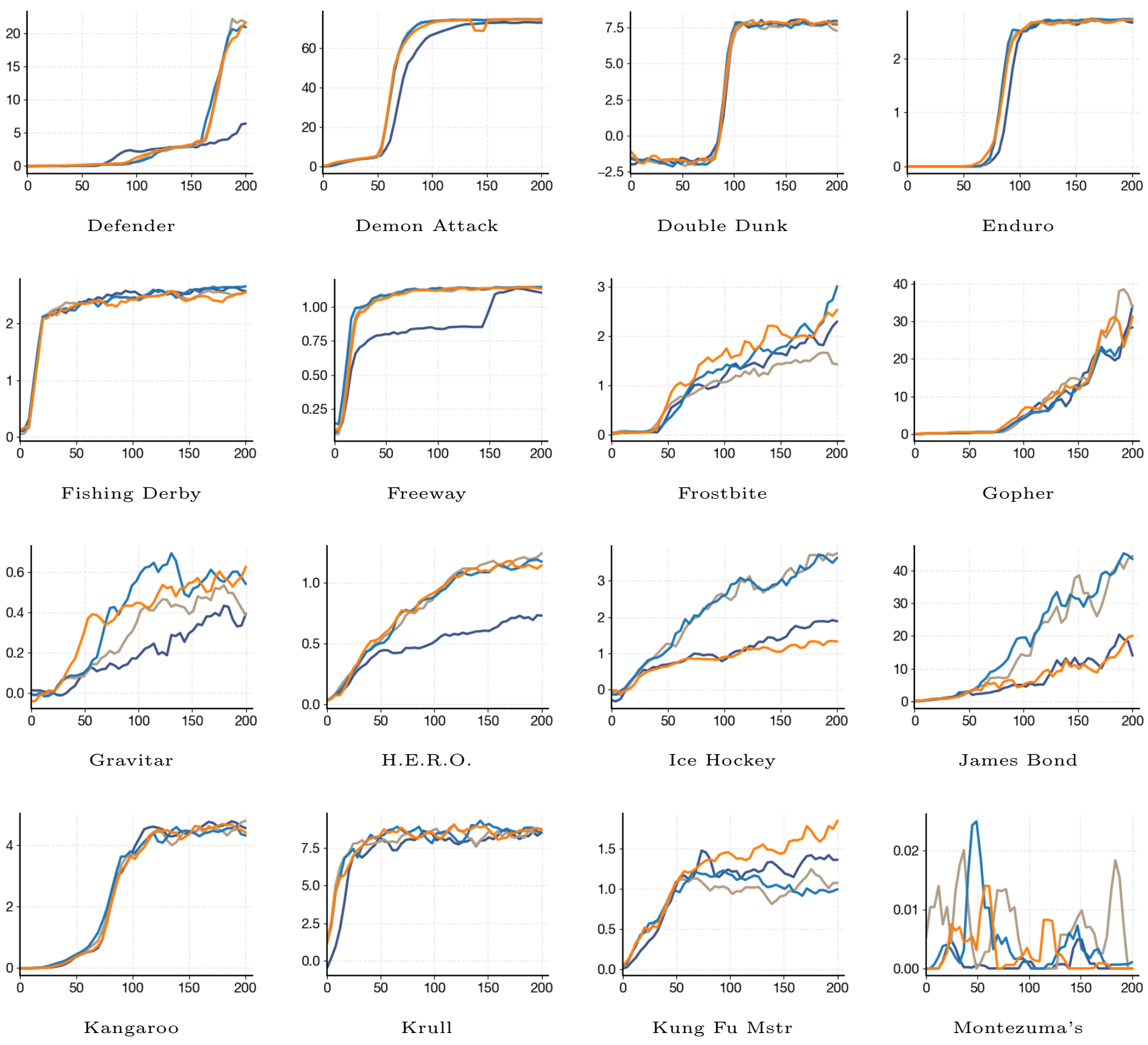


Figure S2: Phase 2 Learning Curves: Human-Normalized Score learning curves over 200 million training frames comparing the standard Gamma baseline to the Hadamax-integrated variants across all 57 Atari games. (Part 2 of 4)

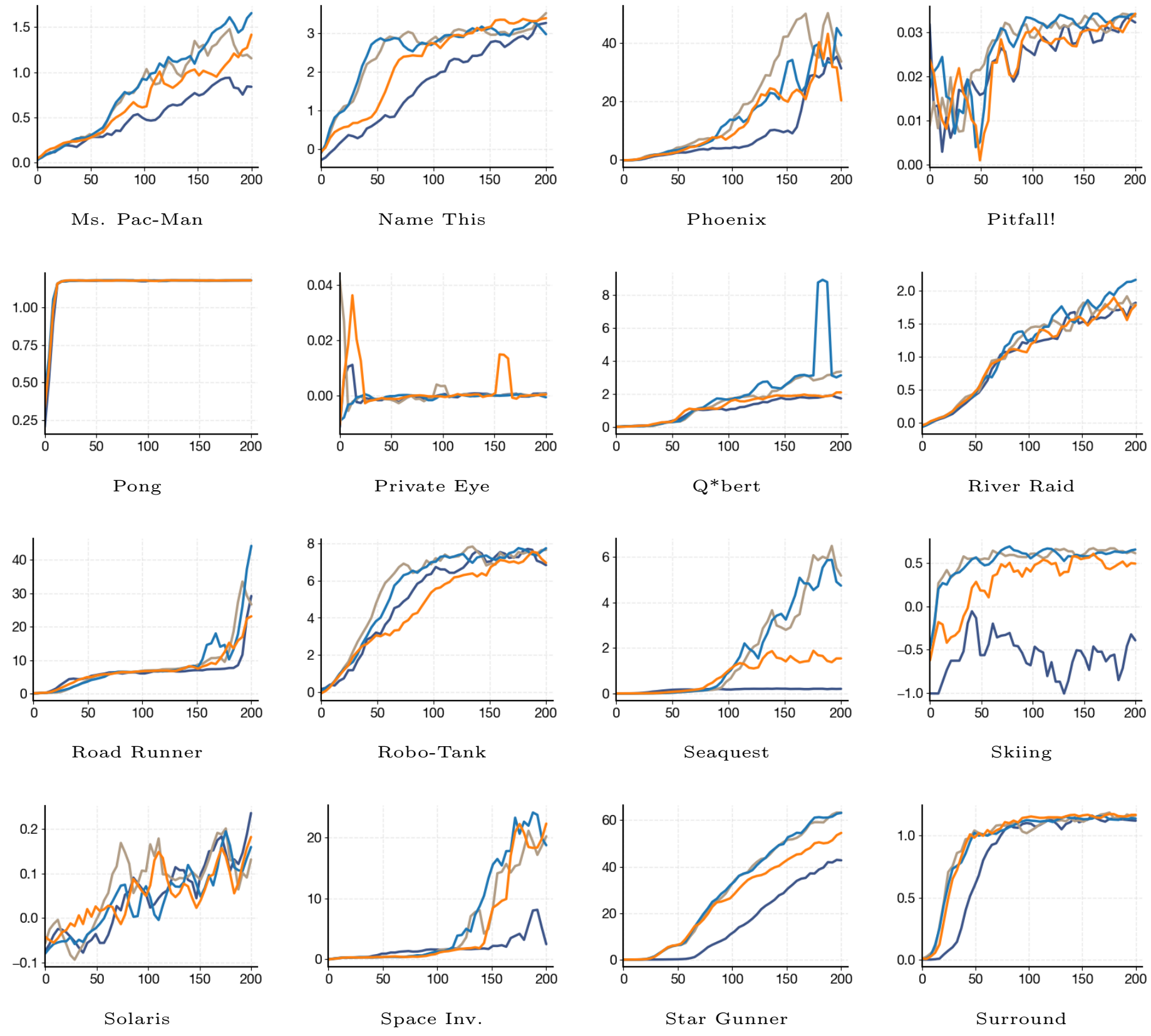


Figure S2: Phase 2 Learning Curves: Human-Normalized Score learning curves over 200 million training frames comparing the standard Gamma baseline to the Hadamax-integrated variants across all 57 Atari games. (Part 3 of 4)

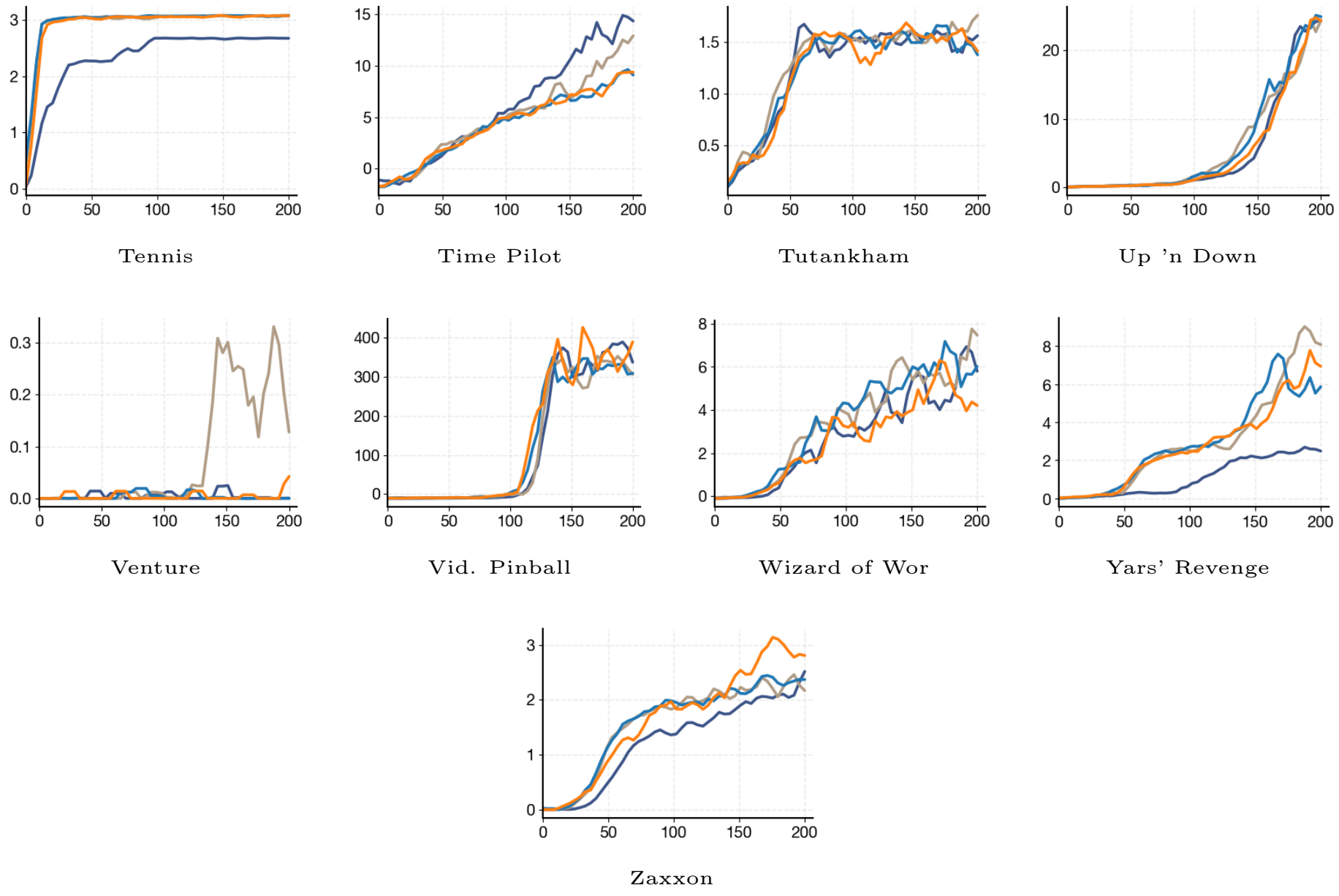


Figure S2: Phase 2 Learning Curves: Human-Normalized Score learning curves over 200 million training frames comparing the standard Gamma baseline to the Hadamax-integrated variants across all 57 Atari games. (Part 4 of 4)

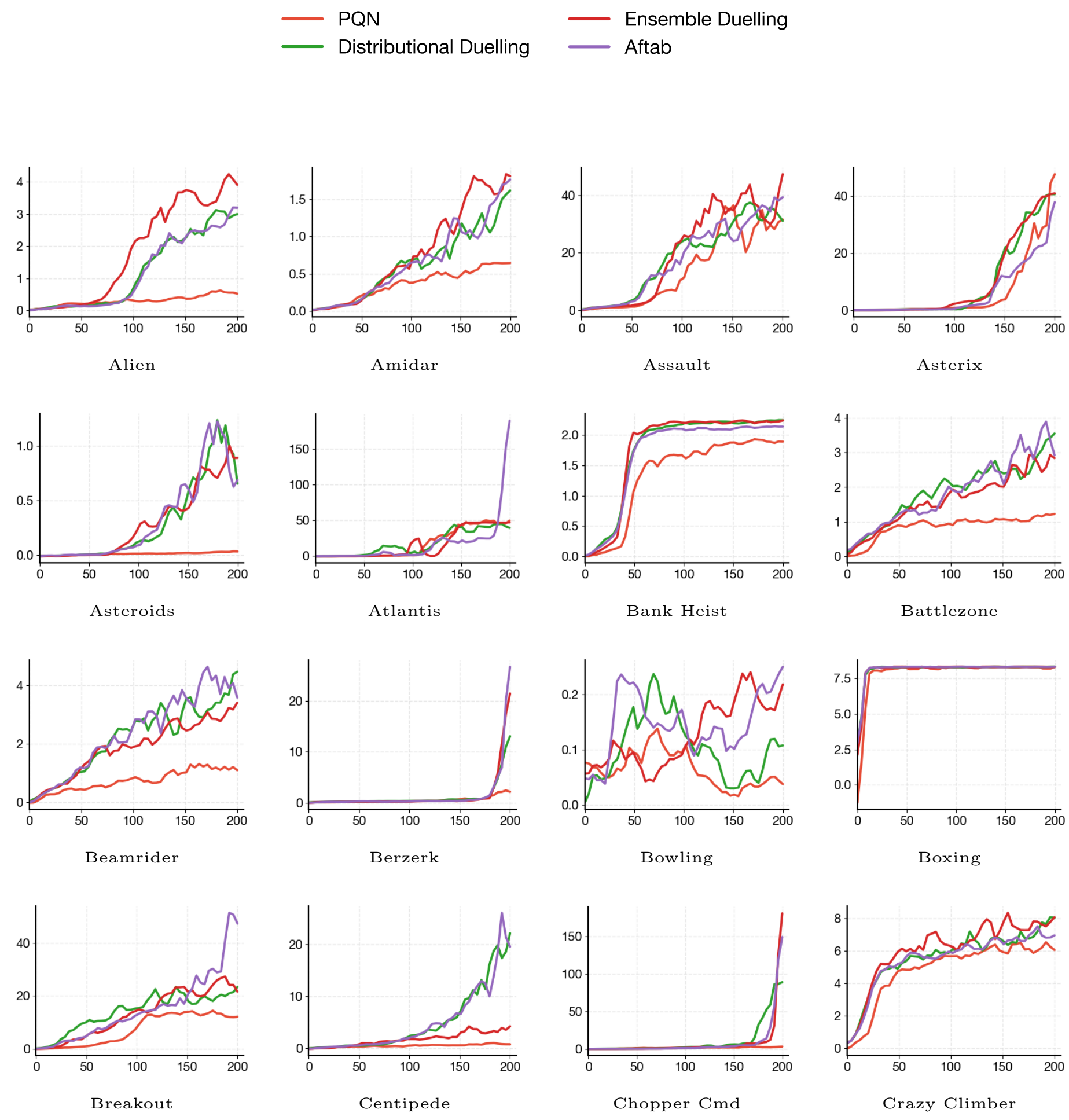


Figure S3: **Phase 3 learning curves.** Human-normalized-score learning curves over 200 million training frames for PQN, Categorical Dueling, Ensemble Dueling, and Aftab across all 57 Atari games. (Part 1 of 4)

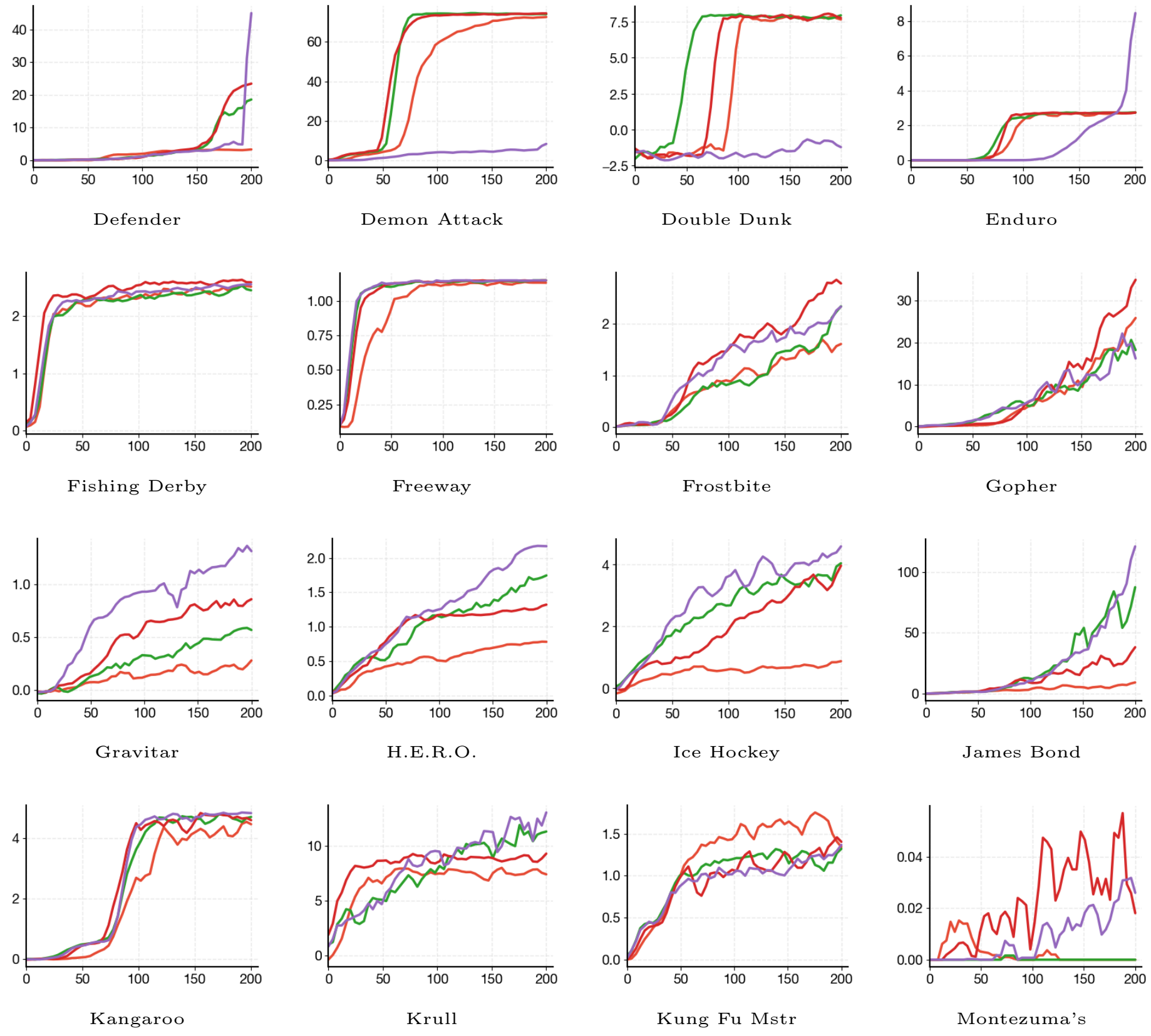


Figure S3: **Phase 3 learning curves.** Human-normalized-score learning curves over 200 million training frames for PQN, Categorical Dueling, Ensemble Dueling, and Aftab across all 57 Atari games. (Part 2 of 4)

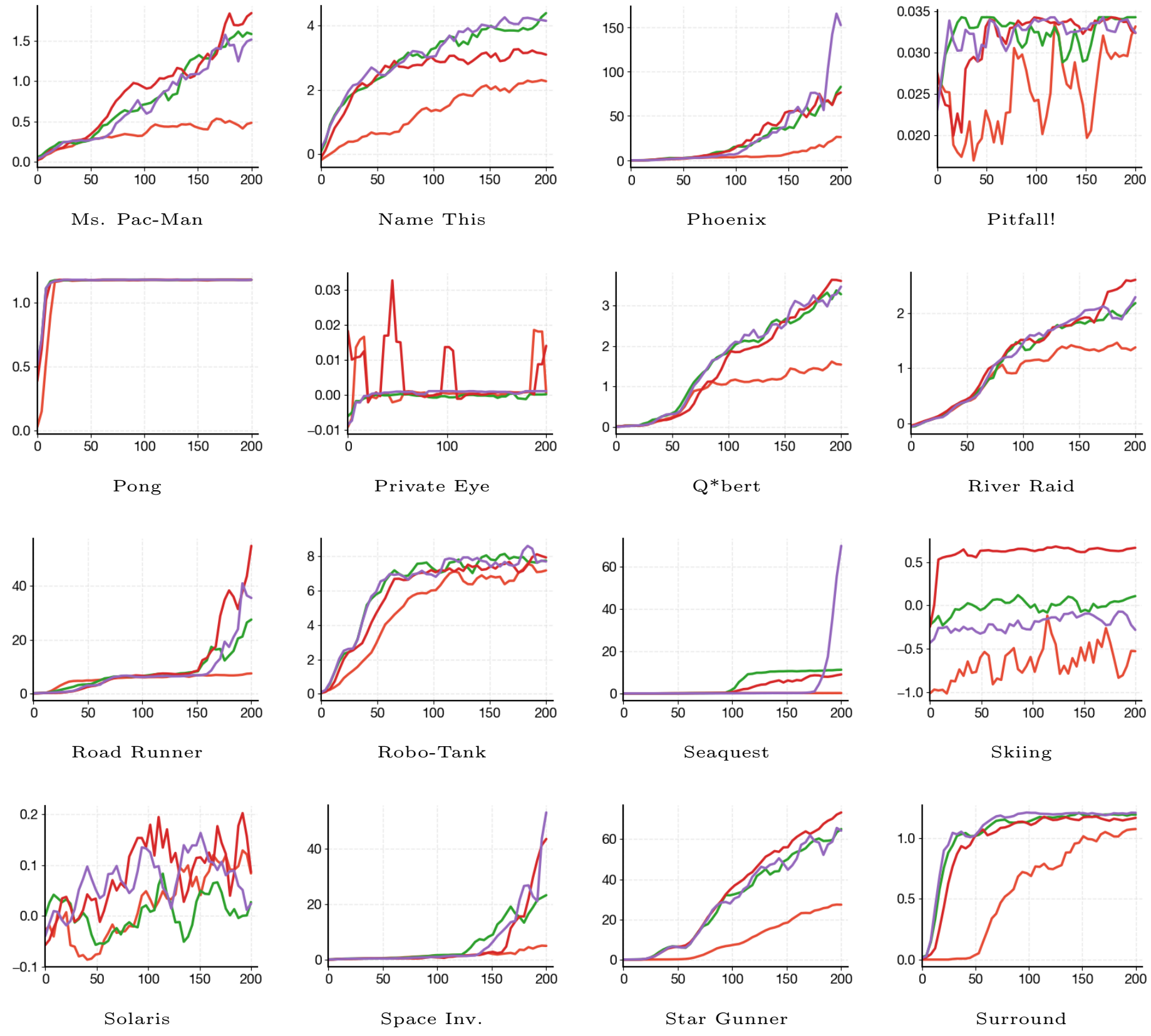


Figure S3: **Phase 3 learning curves.** Human-normalized-score learning curves over 200 million training frames for PQN, Categorical Dueling, Ensemble Dueling, and Aftab across all 57 Atari games. (Part 3 of 4)

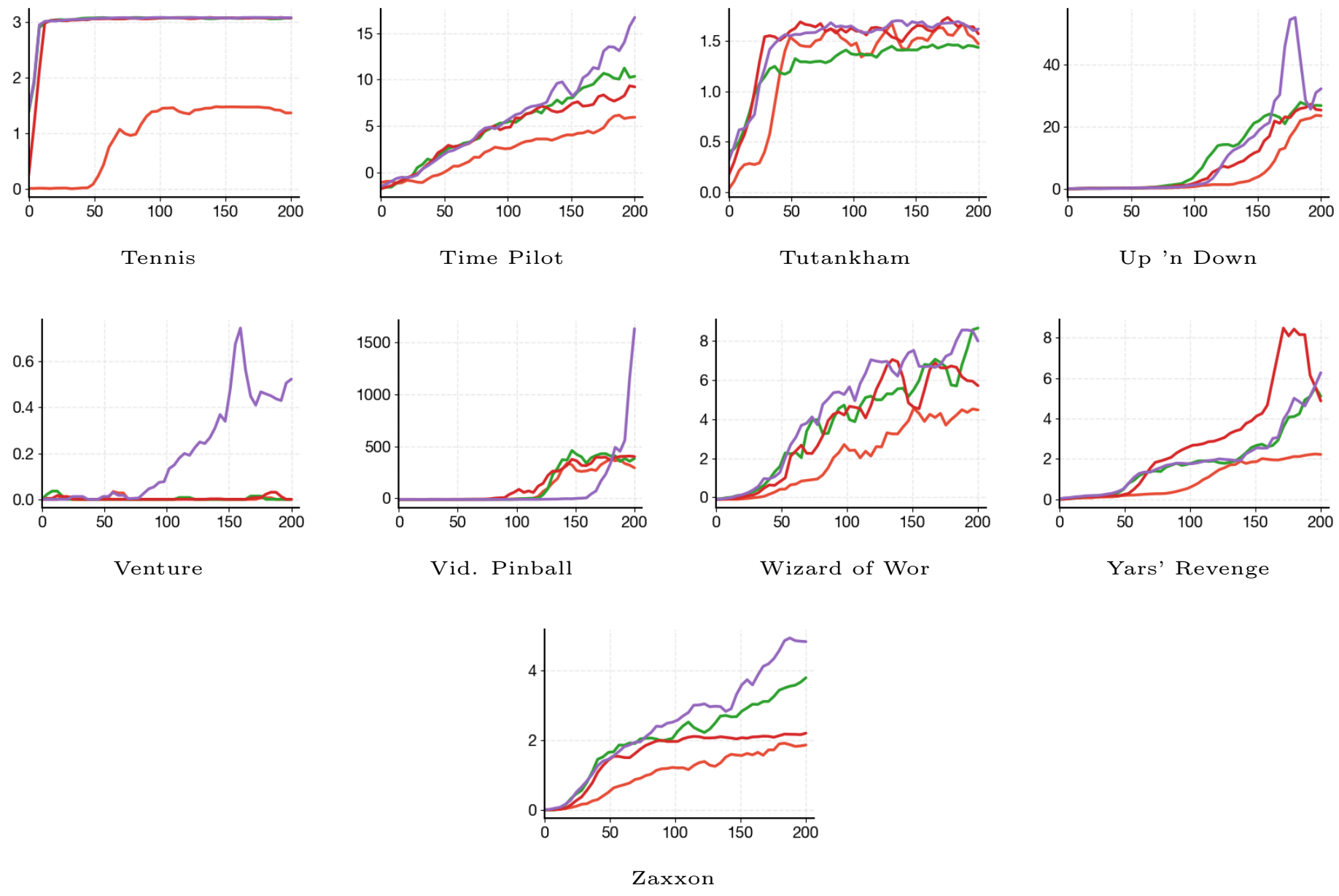


Figure S3: **Phase 3 learning curves.** Human-normalized-score learning curves over 200 million training frames for PQN, Categorical Dueling, Ensemble Dueling, and Aftab across all 57 Atari games. (Part 4 of 4)

Table S9: Full Phase 1 Encoder Results (Part 1: PQN to Delta).

| | PQN | Alpha | Beta | Gamma | Delta |
|---|---|---|---|---|---|
| Alien | 0.551 | 1.821 | 2.421 | **2.806** | 0.370 |
| Amidar | 0.640 | 1.340 | 0.863 | 1.099 | 0.545 |
| Assault | 30.718 | 23.877 | 28.186 | 33.854 | 26.660 |
| Asterix | 44.579 | 15.710 | 15.743 | 14.828 | 35.490 |
| Asteroids | 0.036 | **1.966** | 0.326 | 1.487 | 0.230 |
| Atlantis | 45.579 | 43.774 | 45.780 | 46.256 | 43.365 |
| Bank Heist | 1.894 | 1.990 | 2.123 | 1.725 | 1.828 |
| Battlezone | 1.197 | 1.393 | 1.826 | 1.914 | 1.231 |
| Beamrider | 1.211 | **2.860** | 2.281 | 2.548 | 1.094 |
| Berzerk | 2.456 | **5.368** | 0.343 | 1.282 | 1.306 |
| Bowling | 0.044 | 0.096 | 0.037 | **0.135** | 0.072 |
| Boxing | 8.208 | **8.325** | 8.290 | 8.291 | **8.325** |
| Breakout | 11.918 | 14.731 | 15.050 | **17.307** | 11.621 |
| Centipede | 0.818 | 1.282 | 0.966 | **1.432** | 0.679 |
| Chopper Command | 2.744 | 26.648 | 50.148 | **63.290** | 0.845 |
| Crazy Climber | 6.244 | 6.826 | **7.861** | 6.657 | 6.789 |
| Defender | 3.160 | 4.544 | 4.392 | **6.255** | 3.166 |
| Demon Attack | 72.508 | 72.867 | **73.206** | 72.827 | 70.875 |
| Double Dunk | 7.740 | 7.853 | 7.992 | 7.739 | 7.595 |
| Enduro | 2.730 | **2.737** | 2.705 | 2.685 | 2.728 |
| Fishing Derby | 2.526 | 2.618 | 2.544 | 2.582 | 2.463 |
| Freeway | 1.132 | **1.148** | 1.138 | 1.112 | 1.137 |
| Frostbite | 1.569 | **2.623** | 2.294 | 2.191 | 1.174 |
| Gopher | 24.350 | 27.620 | **34.583** | 28.834 | 15.861 |
| Gravitar | 0.232 | 0.253 | 0.390 | 0.335 | 0.189 |
| H.E.R.O. | 0.786 | 0.836 | 0.756 | 0.733 | 0.694 |
| Ice Hockey | 0.840 | 1.193 | 1.077 | **1.916** | 0.696 |
| James Bond | 8.266 | 12.346 | 6.749 | 18.358 | 4.829 |
| Kangaroo | 4.529 | **4.806** | 4.764 | 4.612 | 3.883 |
| Krull | 7.560 | 8.489 | 8.653 | 8.321 | 7.208 |
| Kung-Fu Master | 1.405 | 1.785 | 1.470 | 1.358 | 1.596 |
| Montezuma's Revenge | 0.000 | 0.000 | 0.000 | 0.000 | 0.000 |
| Ms. Pac-Man | 0.467 | 0.724 | 0.922 | 0.843 | 0.507 |
| Name This Game | 2.302 | 2.843 | 2.538 | **3.242** | 2.344 |
| Phoenix | 26.282 | 39.261 | 36.563 | 35.263 | 7.772 |
| Pitfall! | 0.032 | 0.033 | 0.023 | 0.033 | 0.034 |
| Pong | **1.181** | **1.181** | **1.181** | **1.181** | **1.181** |
| Private Eye | **0.018** | 0.000 | 0.001 | 0.001 | -0.000 |
| Q*bert | 1.555 | 1.871 | 1.853 | 1.805 | 1.332 |
| River Raid | 1.327 | 1.641 | 1.673 | 1.784 | 1.421 |
| Road Runner | 7.353 | 10.694 | 10.317 | **22.764** | 7.498 |
| Robotank | 7.094 | **7.594** | 7.273 | 6.975 | 7.236 |
| Seaquest | 0.190 | 0.240 | 0.192 | 0.197 | 0.199 |
| Skiing | -0.522 | 0.451 | -0.182 | -0.324 | -0.518 |
| Solaris | 0.121 | 0.078 | 0.156 | 0.187 | 0.083 |
| Space Invaders | 4.989 | 4.004 | 8.830 | 5.194 | 4.768 |
| Stargunner | 27.410 | 38.652 | **44.045** | 42.913 | 25.046 |
| Surround | 1.070 | 1.137 | 1.166 | 1.122 | 0.923 |
| Tennis | 1.365 | 2.294 | 1.369 | **2.677** | 1.460 |
| Time Pilot | 5.924 | 15.067 | **21.046** | 14.722 | 4.790 |
| Tutankham | 1.530 | 1.581 | 1.583 | 1.515 | 1.499 |
| Up'n Down | 23.563 | 23.166 | 23.688 | 24.246 | 17.208 |
| Venture | 0.000 | 0.000 | **0.021** | 0.000 | 0.000 |
| Video Pinball | 318.311 | 371.878 | 312.513 | 373.361 | 361.368 |
| Wizard of Wor | 4.518 | 7.059 | 7.129 | 6.668 | 3.270 |

Continued on next page

Table S9 – continued from previous page

| | PQN | Alpha | Beta | Gamma | Delta |
|---|---|---|---|---|---|
| Yars' Revenge | 2.239 | 2.617 | 2.570 | 2.577 | 1.823 |
| Zaxxon | 1.831 | 2.106 | **2.570** | 2.319 | 1.772 |
| Median | 1.894 | **2.618** | 2.421 | 2.577 | 1.596 |
| IQM | 2.715 | **3.566** | 3.464 | 3.508 | 2.388 |
| IQM 95% CI | [1.541, 4.477] | [2.138, 6.292] | [1.909, 6.242] | [2.160, 7.182] | [1.393, 3.870] |

Table S10: Full Phase 1 Encoder Results (Part 2: Epsilon to Theta).

| | Epsilon | Zeta | Eta | Theta |
|---|---|---|---|---|
| Alien | 2.384 | 1.853 | 1.640 | 0.401 |
| Amidar | **1.358** | 0.703 | 1.046 | 0.672 |
| Assault | **36.766** | 30.383 | 33.379 | 34.165 |
| Asterix | 10.611 | 13.262 | 18.969 | **45.196** |
| Asteroids | 1.050 | 1.848 | 0.057 | 0.026 |
| Atlantis | **47.564** | 46.541 | 44.996 | 46.273 |
| Bank Heist | 1.690 | **2.130** | 1.717 | 1.959 |
| Battlezone | **2.019** | 1.506 | 1.035 | 1.039 |
| Beamrider | 2.085 | 1.492 | 1.258 | 1.130 |
| Berzerk | 1.505 | 3.145 | 0.824 | 1.478 |
| Bowling | 0.060 | 0.089 | 0.111 | 0.057 |
| Boxing | **8.325** | 8.302 | **8.325** | 8.320 |
| Breakout | 13.044 | 12.714 | 16.231 | 14.331 |
| Centipede | 0.964 | 0.639 | 1.157 | 0.531 |
| Chopper Command | 44.041 | 33.113 | 3.195 | 1.494 |
| Crazy Climber | 7.191 | 6.711 | 6.851 | 6.090 |
| Defender | 3.866 | 3.957 | 3.192 | 6.064 |
| Demon Attack | 72.666 | 72.110 | 71.404 | 70.828 |
| Double Dunk | 7.819 | 7.731 | **8.113** | 7.719 |
| Enduro | 2.712 | 2.702 | 2.699 | 2.700 |
| Fishing Derby | **2.649** | 2.625 | 2.557 | 2.486 |
| Freeway | 1.138 | 1.140 | 1.140 | 1.135 |
| Frostbite | 1.491 | 2.067 | 1.768 | 1.095 |
| Gopher | 20.647 | 24.558 | 32.248 | 19.611 |
| Gravitar | 0.269 | 0.316 | **0.528** | 0.125 |
| H.E.R.O. | 0.773 | 0.887 | **1.076** | 0.448 |
| Ice Hockey | 1.034 | 1.068 | 1.242 | 0.645 |
| James Bond | **28.346** | 12.110 | 14.549 | 8.973 |
| Kangaroo | 4.520 | 4.761 | 4.570 | 4.647 |
| Krull | 8.128 | **8.653** | 7.968 | 7.693 |
| Kung-Fu Master | 1.426 | 1.447 | 1.445 | **1.830** |
| Montezuma's Revenge | 0.000 | 0.002 | **0.004** | 0.000 |
| Ms. Pac-Man | 0.739 | 0.846 | **1.119** | 0.667 |
| Name This Game | 2.267 | 2.207 | 1.879 | 2.598 |
| Phoenix | 31.549 | **44.392** | 40.795 | 14.803 |
| Pitfall! | 0.033 | **0.034** | 0.033 | 0.030 |
| Pong | 1.180 | **1.181** | **1.181** | **1.181** |
| Private Eye | 0.000 | 0.000 | 0.000 | 0.001 |
| Q*bert | 1.759 | 1.864 | **1.897** | 1.667 |
| River Raid | 1.726 | **1.841** | 1.466 | 1.397 |
| Road Runner | 10.833 | 11.051 | 7.975 | 7.113 |
| Robotank | 7.105 | 7.301 | 6.524 | 6.758 |
| Seaquest | 0.208 | **0.411** | 0.410 | 0.193 |
| Skiing | **0.565** | 0.477 | 0.520 | -0.480 |

Continued on next page

Table S10 – continued from previous page

| | Epsilon | Zeta | Eta | Theta |
|---|---|---|---|---|
| Solaris | 0.118 | 0.105 | 0.191 | **0.295** |
| Space Invaders | 5.956 | 1.728 | **15.470** | 4.586 |
| Stargunner | 32.754 | 32.033 | 24.769 | 23.882 |
| Surround | 1.147 | **1.203** | 1.192 | 0.936 |
| Tennis | 2.266 | 1.366 | 1.874 | 1.388 |
| Time Pilot | 12.470 | 12.749 | 14.516 | 4.490 |
| Tutankham | 1.487 | 1.510 | **1.623** | 1.533 |
| Up'n Down | 27.050 | **28.171** | 7.974 | 24.838 |
| Venture | 0.000 | 0.000 | 0.000 | 0.000 |
| Video Pinball | 298.742 | 372.805 | **374.307** | 335.655 |
| Wizard of Wor | 7.748 | 6.037 | **8.070** | 5.707 |
| Yars' Revenge | 2.328 | **2.740** | 2.683 | 2.194 |
| Zaxxon | 1.604 | 1.731 | 2.208 | 1.933 |
| Median | 2.266 | 2.067 | 1.879 | 1.830 |
| IQM | 3.341 | 3.199 | 3.106 | 2.688 |
| IQM 95% CI | [1.851, 6.077] | [1.802, 5.791] | [1.708, 5.692] | [1.535, 4.504] |

Table S11: Phase 1 Atari-57 Raw Returns: Final unnormalized evaluation scores for the PQN baseline and the eight parameter-constrained CNN encoders (Alpha through Theta). (Part 1 of 2)

| | PQN | Alpha | Beta | Gamma | Delta |
|---|---|---|---|---|---|
| Alien | 4688.197 | 13152.109 | 14752.177 | **18963.333** | 2680.850 |
| Amidar | 1082.500 | 2178.153 | 1139.731 | 1893.639 | 974.150 |
| Assault | 17009.493 | 12128.391 | 14126.837 | 18030.888 | 13932.092 |
| Asterix | 366434.014 | 129816.327 | 154993.197 | 126304.082 | 235632.653 |
| Asteroids | 2312.721 | **103959.762** | 15110.748 | 68774.320 | 11182.687 |
| Atlantis | 742852.041 | 728122.789 | 751796.599 | 742603.741 | 736686.735 |
| Bank Heist | 1402.143 | 1506.395 | **1595.000** | 1288.912 | 1332.789 |
| Battlezone | 44176.871 | 43829.932 | 64149.660 | 64085.034 | 44642.857 |
| Beamrider | 22017.973 | **47062.184** | 39400.667 | 39152.435 | 18069.789 |
| Berzerk | 5877.857 | **15391.463** | 958.571 | 1942.279 | 3578.537 |
| Bowling | 30.000 | 37.830 | 27.724 | 30.959 | 31.333 |
| Boxing | 99.228 | **100.000** | **100.000** | 99.587 | **100.000** |
| Breakout | 352.044 | 436.816 | 451.425 | **504.044** | 326.320 |
| Centipede | 10076.690 | 13596.034 | 11219.793 | **17675.850** | 8680.939 |
| Chopper Command | 17155.442 | 118263.946 | 334721.088 | **360188.095** | 6479.252 |
| Crazy Climber | 161068.367 | 179992.517 | **204894.898** | 179469.728 | 185854.422 |
| Defender | 51787.245 | 75035.374 | 66311.905 | **91159.694** | 53549.320 |
| Demon Attack | 132035.357 | 132607.364 | **133369.286** | 132650.221 | 128967.942 |
| Double Dunk | -1.782 | -1.238 | -1.095 | -1.707 | -1.980 |
| Enduro | **2360.721** | 2357.741 | 2325.959 | 2341.762 | 2353.371 |
| Fishing Derby | 42.286 | 46.748 | 42.823 | 43.272 | 38.197 |
| Freeway | 33.823 | **33.983** | 33.827 | 32.973 | 33.653 |
| Frostbite | 6653.946 | **9596.701** | 9046.803 | 9532.925 | 4842.925 |
| Gopher | 53851.973 | 64661.769 | **72346.054** | 51893.333 | 34766.463 |
| Gravitar | 956.122 | 645.408 | 1319.558 | 1293.367 | 730.102 |
| H.E.R.O. | 24341.207 | 25199.864 | 21372.449 | 24353.571 | 20274.881 |
| Ice Hockey | 0.177 | 2.136 | 0.895 | **14.364** | -3.622 |
| James Bond | 2195.068 | 2500.850 | 1569.048 | 5012.245 | 1360.374 |
| Kangaroo | 13455.442 | 14392.857 | 14387.075 | 14087.755 | 11813.605 |
| Krull | 9707.585 | 10637.381 | 10961.361 | 10724.456 | 9399.592 |
| Kung-Fu Master | 31990.136 | 38857.823 | 31007.823 | 35474.830 | 34184.014 |
| Montezuma's Revenge | **0.000** | **0.000** | **0.000** | **0.000** | **0.000** |
| Ms. Pac-Man | 3365.340 | 5302.415 | 6122.891 | 5855.238 | 3515.544 |
| Name This Game | 15329.014 | 19033.571 | 16205.748 | **21340.510** | 15285.782 |
| Phoenix | 177903.333 | 213915.986 | 235491.803 | 190653.197 | 51112.245 |
| Pitfall! | -5.344 | **0.000** | -21.105 | **0.000** | **0.000** |
| Pong | **21.000** | **21.000** | **21.000** | **21.000** | **21.000** |
| Private Eye | 66.667 | 98.639 | **100.000** | 87.333 | 9.524 |
| Q*bert | 21923.469 | 25177.296 | 25397.874 | 25847.959 | 17267.262 |
| River Raid | 21349.014 | 27019.490 | 27153.299 | 29396.327 | 23497.007 |
| Road Runner | 57719.048 | 72455.442 | 63845.918 | **175828.571** | 56056.463 |
| Robotank | 71.167 | **76.075** | 72.412 | 68.752 | 72.170 |
| Seaquest | 7978.027 | 10092.925 | 7835.646 | 8228.844 | 8466.871 |
| Skiing | -23644.677 | **-9432.129** | -18793.803 | -22986.874 | -23272.214 |
| Solaris | 2605.306 | 2069.932 | 2881.837 | 3081.497 | 2004.762 |
| Space Invaders | 5342.296 | 2655.680 | 11072.245 | 2855.578 | 6468.639 |
| Stargunner | 259520.408 | 371468.027 | **423331.293** | 403316.667 | 241495.918 |
| Surround | 8.350 | 8.857 | 9.248 | 9.146 | 6.071 |
| Tennis | -1.333 | 11.833 | -4.316 | **23.881** | -1.092 |
| Time Pilot | 13860.884 | 27697.619 | **32540.816** | 27330.272 | 11938.095 |
| Tutankham | 249.473 | 255.932 | 264.330 | 254.279 | 259.340 |
| Up'n Down | 263719.048 | 253879.762 | 295438.095 | 266537.007 | 164273.027 |

Continued on next page

Table S11 – continued from previous page

| | PQN | Alpha | Beta | Gamma | Delta |
|---|---|---|---|---|---|
| Venture | **0.000** | **0.000** | **0.000** | **0.000** | **0.000** |
| Video Pinball | 457305.670 | 542947.592 | 448438.116 | 529722.776 | 510972.653 |
| Wizard of Wor | 16897.959 | 28960.544 | 29923.469 | 29785.714 | 13953.401 |
| Yars' Revenge | 115017.827 | 136848.190 | 138392.262 | 140728.207 | 101737.813 |
| Zaxxon | 16136.735 | 17865.306 | **24046.939** | 20938.435 | 14489.116 |

Table S12: Phase 1 Atari-57 Raw Returns: Final unnormalized evaluation scores for the PQN baseline and the eight parameter-constrained CNN encoders (Alpha through Theta). (Part 2 of 2)

| | Epsilon | Zeta | Eta | Theta |
|---|---|---|---|---|
| Alien | 17525.578 | 11592.449 | 11572.415 | 3209.184 |
| Amidar | **2362.010** | 1200.925 | 1726.520 | 1234.459 |
| Assault | **18513.439** | 14149.139 | 17580.704 | 17846.646 |
| Asterix | 86924.830 | 103686.395 | 85679.932 | **374916.327** |
| Asteroids | 28792.415 | 85168.333 | 2549.150 | 1950.884 |
| Atlantis | **761386.735** | 747847.619 | 751528.912 | 758324.490 |
| Bank Heist | 1257.517 | 1588.163 | 1302.075 | 1514.762 |
| Battlezone | **65819.728** | 48993.197 | 38017.007 | 37350.340 |
| Beamrider | 33208.973 | 22814.422 | 20786.871 | 19927.231 |
| Berzerk | 3434.660 | 8512.109 | 2253.231 | 3921.633 |
| Bowling | 30.442 | 34.667 | **38.007** | 30.571 |
| Boxing | **100.000** | **100.000** | **100.000** | **100.000** |
| Breakout | 405.925 | 394.044 | 431.728 | 407.690 |
| Centipede | 11957.031 | 8450.085 | 12807.017 | 7412.735 |
| Chopper Command | 258300.000 | 185923.810 | 19930.612 | 11332.993 |
| Crazy Climber | 191479.252 | 175811.224 | 179710.884 | 153729.252 |
| Defender | 61850.510 | 65859.354 | 50575.850 | 62583.844 |
| Demon Attack | 132331.548 | 131675.442 | 130311.701 | 128636.190 |
| Double Dunk | -1.483 | -1.442 | **-0.850** | -1.449 |
| Enduro | 2343.173 | 2339.864 | 2338.684 | 2326.656 |
| Fishing Derby | **49.184** | 47.177 | 42.537 | 41.000 |
| Freeway | 33.840 | 33.667 | 33.820 | 33.806 |
| Frostbite | 6228.571 | 8996.667 | 8130.306 | 4846.871 |
| Gopher | 39304.626 | 32817.483 | 68673.129 | 45066.190 |
| Gravitar | 1032.483 | 978.061 | **1874.830** | 594.388 |
| H.E.R.O. | 24368.827 | 30249.847 | **34076.003** | 14280.051 |
| Ice Hockey | 0.721 | 1.531 | 2.235 | -3.690 |
| James Bond | **7288.435** | 2657.653 | 3470.918 | 2228.912 |
| Kangaroo | 13774.150 | **14466.667** | 14088.095 | 14328.231 |
| Krull | 10308.878 | **11143.061** | 9998.946 | 10395.680 |
| Kung-Fu Master | 35648.299 | 33834.354 | 32204.762 | **41987.075** |
| Montezuma's Revenge | **0.000** | **0.000** | **0.000** | **0.000** |
| Ms. Pac-Man | 5614.456 | 5822.483 | **7529.932** | 4865.816 |
| Name This Game | 16041.735 | 15393.265 | 12873.435 | 17054.830 |
| Phoenix | 185824.252 | **283574.218** | 255857.245 | 98561.429 |
| Pitfall! | **0.000** | **0.000** | -0.871 | -22.041 |
| Pong | **21.000** | **21.000** | **21.000** | **21.000** |
| Private Eye | 50.000 | 50.000 | 44.983 | **100.000** |
| Q*bert | 24266.241 | 25570.153 | **26119.218** | 22553.486 |
| River Raid | 28564.456 | **30737.823** | 24621.361 | 23529.388 |
| Road Runner | 61521.088 | 81121.769 | 56501.361 | 57602.381 |
| Robotank | 71.554 | 73.391 | 64.133 | 66.609 |

Continued on next page

Table S12 – continued from previous page

| | Epsilon | Zeta | Eta | Theta |
|---|---|---|---|---|
| Seaquest | 8840.816 | **17571.224** | 17243.741 | 8063.129 |
| Skiing | -9766.480 | -10895.473 | -10667.639 | -21527.639 |
| Solaris | 2541.542 | 2247.279 | 3255.102 | **4381.088** |
| Space Invaders | 2966.718 | 2851.633 | **24713.435** | 2864.932 |
| Stargunner | 320142.517 | 315854.082 | 237664.966 | 229308.503 |
| Surround | 9.262 | **9.905** | 9.833 | 7.112 |
| Tennis | 11.633 | -2.500 | -1.000 | -1.303 |
| Time Pilot | 23786.054 | 24746.939 | 26560.544 | 9999.320 |
| Tutankham | 248.612 | 248.173 | **268.939** | 246.932 |
| Up'n Down | 307976.429 | **318144.694** | 91661.429 | 268903.095 |
| Venture | **0.000** | **0.000** | **0.000** | **0.000** |
| Video Pinball | 462849.429 | **544883.133** | 529542.265 | 487644.446 |
| Wizard of Wor | **34776.190** | 23365.646 | 33848.980 | 26704.762 |
| Yars' Revenge | 124140.105 | **144543.561** | 141543.759 | 115265.139 |
| Zaxxon | 15231.293 | 15286.395 | 20634.694 | 17481.633 |

Table S13: Phase 2 Atari-57 Results (HNS).

| | Gamma | Gamma-Hadamax-Valid | Gamma-Hadamax-Same | Hadamax |
|---|---|---|---|---|
| Alien | 2.806 | 2.262 | **3.143** | 2.956 |
| Amidar | 1.099 | **1.538** | 1.170 | 0.814 |
| Assault | 33.854 | 40.430 | **45.000** | 42.586 |
| Asterix | 14.828 | 37.491 | **42.532** | 32.479 |
| Asteroids | 1.487 | 1.346 | **2.006** | 1.003 |
| Atlantis | **46.256** | 43.382 | 37.872 | 44.555 |
| Bank Heist | 1.725 | **2.113** | 2.046 | 2.108 |
| Battlezone | 1.914 | **2.362** | 1.912 | 2.275 |
| Beamrider | 2.548 | 2.709 | **4.126** | 2.594 |
| Berzerk | 1.282 | **23.230** | 21.633 | 10.371 |
| Bowling | **0.135** | 0.131 | 0.034 | 0.046 |
| Boxing | 8.291 | 8.248 | **8.325** | 8.321 |
| Breakout | 17.307 | 14.305 | 19.649 | **20.023** |
| Centipede | 1.432 | 2.772 | **3.109** | 2.261 |
| Chopper Command | 63.290 | 66.259 | **105.041** | 27.477 |
| Crazy Climber | 6.657 | 7.892 | 8.703 | **8.946** |
| Defender | 6.255 | **21.932** | 21.030 | 20.902 |
| Demon Attack | 72.827 | **74.642** | 74.361 | 74.521 |
| Double Dunk | 7.739 | 7.388 | **7.961** | 7.807 |
| Enduro | 2.685 | 2.730 | **2.733** | 2.711 |
| Fishing Derby | 2.582 | 2.513 | **2.634** | 2.523 |
| Freeway | 1.112 | 1.146 | **1.148** | 1.140 |
| Frostbite | 2.191 | 1.446 | **2.729** | 2.408 |
| Gopher | 28.834 | **37.097** | 27.959 | 26.749 |
| Gravitar | 0.335 | 0.422 | 0.575 | **0.577** |
| H.E.R.O. | 0.733 | **1.209** | 1.188 | 1.115 |
| Ice Hockey | 1.916 | **3.701** | 3.498 | 1.342 |
| James Bond | 18.358 | 43.679 | **44.754** | 19.589 |
| Kangaroo | 4.612 | **4.722** | 4.352 | 4.491 |
| Krull | 8.321 | 8.800 | 8.581 | **8.820** |
| Kung-Fu Master | 1.358 | 1.069 | 0.969 | **1.741** |
| Montezuma's Revenge | 0.000 | 0.000 | **0.001** | 0.000 |
| Ms. Pac-Man | 0.843 | 1.195 | **1.600** | 1.272 |
| Name This Game | 3.242 | **3.405** | 3.111 | 3.368 |
| Phoenix | 35.263 | 37.001 | **44.998** | 31.553 |
| Pitfall! | 0.033 | 0.034 | **0.034** | 0.033 |
| Pong | **1.181** | **1.181** | **1.181** | **1.181** |
| Private Eye | **0.001** | 0.001 | -0.000 | 0.000 |
| Q*bert | 1.805 | **3.322** | 3.006 | 2.091 |
| River Raid | 1.784 | 1.787 | **2.140** | 1.712 |
| Road Runner | 22.764 | 29.294 | **37.244** | 22.406 |
| Robotank | 6.975 | **7.643** | 7.565 | 7.174 |
| Seaquest | 0.197 | **5.522** | 4.914 | 1.540 |
| Skiing | -0.324 | 0.636 | **0.649** | 0.501 |
| Solaris | **0.187** | 0.091 | 0.134 | 0.157 |
| Space Invaders | 5.194 | 18.616 | **19.901** | 19.767 |
| Stargunner | 42.913 | **63.307** | 62.754 | 53.536 |
| Surround | 1.122 | 1.155 | 1.140 | **1.167** |
| Tennis | 2.677 | **3.084** | 3.075 | 3.078 |
| Time Pilot | **14.722** | 12.217 | 9.629 | 9.367 |
| Tutankham | 1.515 | **1.701** | 1.460 | 1.477 |
| Up'n Down | 24.246 | 22.753 | **25.159** | 24.811 |
| Venture | 0.000 | **0.203** | 0.000 | 0.028 |
| Video Pinball | **373.361** | 321.522 | 306.502 | 357.007 |
| Wizard of Wor | 6.668 | **7.758** | 5.651 | 4.370 |

Continued on next page

Table S13 – continued from previous page

| | Gamma | Gamma-Hadamax-Valid | Gamma-Hadamax-Same | Hadamax |
|---|---|---|---|---|
| Yars' Revenge | 2.577 | **8.192** | 5.519 | 7.117 |
| Zaxxon | 2.319 | 2.267 | 2.372 | **2.829** |
| Median | 2.577 | **3.322** | 3.143 | 2.829 |
| IQM | 3.508 | 5.254 | **5.360** | 4.712 |
| IQM 95% CI | [2.182, 7.036] | [2.990, 10.524] | [3.034, 11.111] | [2.635, 9.232] |

Table S14: Phase 2 Atari-57 Raw Returns: Final unnormalized evaluation scores comparing the standard Gamma encoder against the Gamma-Hadamax-Valid (V1) and Gamma-Hadamax-Same (V2) architectures.

| | Gamma | Gamma-Hadamax-Valid | Gamma-Hadamax-Same | Hadamax |
|---|---|---|---|---|
| Alien | 18963.333 | 14507.891 | 17811.599 | **19762.347** |
| Amidar | 1893.639 | **2658.752** | 1971.486 | 1357.088 |
| Assault | 18030.888 | 21776.354 | **23277.939** | 22982.881 |
| Asterix | 126304.082 | 320721.088 | **353770.408** | 268579.932 |
| Asteroids | 68774.320 | 68211.905 | **90121.837** | 39592.245 |
| Atlantis | 742603.741 | 706664.286 | 668376.531 | **763619.728** |
| Bank Heist | 1288.912 | **1576.735** | 1533.095 | 1571.020 |
| Battlezone | 64085.034 | 83642.857 | 63918.367 | **89863.946** |
| Beamrider | 39152.435 | 47184.265 | **64156.415** | 44187.966 |
| Berzerk | 1942.279 | **58130.476** | 55552.925 | 31191.837 |
| Bowling | 30.959 | **37.031** | 27.548 | 30.667 |
| Boxing | 99.587 | 99.293 | **100.000** | **100.000** |
| Breakout | 504.044 | 410.425 | 533.874 | **547.340** |
| Centipede | 17675.850 | 23132.677 | **34114.677** | 26424.534 |
| Chopper Command | 360188.095 | 422371.088 | **790388.095** | 141528.912 |
| Crazy Climber | 179469.728 | 192591.837 | **232073.469** | 227088.095 |
| Defender | 91159.694 | **365917.177** | 360621.599 | 351619.898 |
| Demon Attack | 132650.221 | 135620.833 | 135425.476 | **135638.333** |
| Double Dunk | -1.707 | -2.306 | **-1.136** | -1.415 |
| Enduro | 2341.762 | 2348.207 | **2352.299** | 2332.058 |
| Fishing Derby | 43.272 | 40.116 | **47.395** | 42.932 |
| Freeway | 32.973 | **34.000** | **34.000** | 33.806 |
| Frostbite | 9532.925 | 6893.741 | 10607.891 | **10759.048** |
| Gopher | 51893.333 | **83183.061** | 56273.265 | 59860.612 |
| Gravitar | 1293.367 | 1360.374 | **2035.034** | 2000.850 |
| H.E.R.O. | 24353.571 | 36814.575 | **36880.816** | 34529.915 |
| Ice Hockey | 14.364 | **35.327** | 31.551 | 3.262 |
| James Bond | 5012.245 | 11395.918 | **12455.612** | 4558.333 |
| Kangaroo | 14087.755 | **14363.265** | 13030.272 | 13900.340 |
| Krull | 10724.456 | 11037.177 | 11042.143 | **11045.170** |
| Kung-Fu Master | 35474.830 | 23623.810 | 22875.170 | **39016.327** |
| Montezuma's Revenge | **0.000** | **0.000** | **0.000** | **0.000** |
| Ms. Pac-Man | 5855.238 | 8367.099 | **11653.384** | 8421.429 |
| Name This Game | 21340.510 | 21829.626 | 20023.469 | **21919.864** |
| Phoenix | 190653.197 | 237966.701 | **286008.673** | 201750.442 |
| Pitfall! | **0.000** | **0.000** | -0.673 | **0.000** |
| Pong | **21.000** | **21.000** | **21.000** | **21.000** |
| Private Eye | **87.333** | 81.000 | 14.626 | 24.490 |
| Q*bert | 25847.959 | **44113.690** | 39942.517 | 27978.912 |
| River Raid | 29396.327 | 29878.061 | **35587.653** | 29210.612 |
| Road Runner | 175828.571 | 226921.769 | **246640.476** | 168071.088 |
| Robotank | 68.752 | 75.582 | **76.265** | 72.265 |
| Seaquest | 8228.844 | **263118.503** | 208911.565 | 63600.136 |
| Skiing | -22986.874 | -8848.003 | **-8846.616** | -10685.082 |
| Solaris | **3081.497** | 2252.789 | 2734.150 | 2833.537 |
| Space Invaders | 2855.578 | 30371.241 | **30697.704** | 27396.701 |
| Stargunner | 403316.667 | **608855.102** | 606955.442 | 504784.354 |
| Surround | 9.146 | **9.827** | 9.401 | 9.486 |
| Tennis | 23.881 | **24.000** | 23.830 | **24.000** |
| Time Pilot | **27330.272** | 22304.762 | 17815.646 | 18702.721 |
| Tutankham | 254.279 | **275.857** | 240.582 | 257.537 |
| Up'n Down | 266537.007 | 271865.952 | **288032.857** | 269874.490 |

Continued on next page

Table S14 – continued from previous page

| | Gamma | Gamma-Hadamax-Valid | Gamma-Hadamax-Same | Hadamax |
|---|---|---|---|---|
| Venture | 0.000 | **18.367** | 0.000 | 0.000 |
| Video Pinball | **529722.776** | 489526.680 | 450481.476 | 478247.367 |
| Wizard of Wor | 29785.714 | **33751.701** | 26120.408 | 19372.109 |
| Yars' Revenge | 140728.207 | **440586.704** | 260409.735 | 370792.272 |
| Zaxxon | 20938.435 | 20152.721 | 20314.966 | **26114.966** |

Table S15: Phase 3 Atari-57 Results (HNS).

| | PQN | Categorical Dueling | Ensemble Dueling | Aftab |
|---|---|---|---|---|
| Alien | 0.551 | 2.948 | **4.079** | 3.201 |
| Amidar | 0.640 | 1.565 | **1.839** | 1.717 |
| Assault | 30.718 | 33.035 | **40.911** | 38.166 |
| Asterix | **44.579** | 40.605 | 40.652 | 33.011 |
| Asteroids | 0.036 | **0.920** | 0.890 | 0.627 |
| Atlantis | 45.579 | 41.334 | 46.637 | **150.786** |
| Bank Heist | 1.894 | **2.243** | 2.222 | 2.139 |
| Battlezone | 1.197 | **3.424** | 2.927 | 3.353 |
| Beamrider | 1.211 | **4.372** | 3.179 | 4.073 |
| Berzerk | 2.456 | 11.011 | 17.443 | **20.114** |
| Bowling | 0.044 | 0.106 | 0.195 | **0.237** |
| Boxing | 8.208 | **8.325** | 8.317 | 8.312 |
| Breakout | 11.918 | 21.476 | 24.151 | **50.740** |
| Centipede | 0.818 | 18.564 | 3.595 | **21.325** |
| Chopper Command | 2.744 | 86.611 | **126.862** | 119.595 |
| Crazy Climber | 6.244 | **8.057** | 7.761 | 6.812 |
| Defender | 3.160 | 18.042 | 23.064 | **31.197** |
| Demon Attack | 72.508 | 74.118 | **74.480** | 7.215 |
| Double Dunk | 7.740 | 7.853 | **7.856** | -1.085 |
| Enduro | 2.730 | 2.729 | 2.697 | **6.846** |
| Fishing Derby | 2.526 | 2.470 | **2.592** | 2.554 |
| Freeway | 1.132 | **1.149** | **1.149** | 1.148 |
| Frostbite | 1.569 | 2.234 | **2.852** | 2.258 |
| Gopher | 24.350 | 20.596 | **33.025** | 19.747 |
| Gravitar | 0.232 | 0.589 | 0.840 | **1.368** |
| H.E.R.O. | 0.786 | 1.718 | 1.304 | **2.172** |
| Ice Hockey | 0.840 | 3.920 | 3.669 | **4.374** |
| James Bond | 8.266 | 71.457 | 33.223 | **109.931** |
| Kangaroo | 4.529 | 4.650 | 4.655 | **4.825** |
| Krull | 7.560 | 11.239 | 8.875 | **11.900** |
| Kung-Fu Master | 1.405 | 1.268 | **1.456** | 1.295 |
| Montezuma's Revenge | 0.000 | 0.000 | 0.026 | **0.032** |
| Ms. Pac-Man | 0.467 | 1.606 | **1.806** | 1.496 |
| Name This Game | 2.302 | **4.261** | 3.137 | 4.172 |
| Phoenix | 26.282 | 76.027 | 73.512 | **165.021** |
| Pitfall! | 0.032 | **0.034** | 0.033 | 0.033 |
| Pong | **1.181** | **1.181** | **1.181** | 1.179 |
| Private Eye | **0.018** | 0.000 | 0.009 | 0.001 |
| Q*bert | 1.555 | 3.376 | **3.636** | 3.281 |
| River Raid | 1.327 | 2.114 | **2.583** | 2.146 |
| Road Runner | 7.353 | 26.381 | **43.579** | 36.413 |
| Robotank | 7.094 | 7.723 | **8.004** | 7.736 |
| Seaquest | 0.190 | 11.162 | 8.638 | **55.555** |
| Skiing | -0.522 | 0.095 | **0.655** | -0.223 |
| Solaris | 0.121 | 0.001 | **0.157** | 0.012 |
| Space Invaders | 4.989 | 22.334 | 41.280 | **43.490** |
| Stargunner | 27.410 | 63.408 | **72.289** | 65.620 |
| Surround | 1.070 | 1.193 | 1.157 | **1.207** |
| Tennis | 1.365 | 3.073 | **3.078** | 3.078 |
| Time Pilot | 5.924 | 10.247 | 9.330 | **15.709** |
| Tutankham | 1.530 | 1.446 | **1.636** | 1.599 |
| Up'n Down | 23.563 | 26.803 | 25.461 | **30.842** |
| Venture | 0.000 | 0.000 | 0.000 | **0.506** |
| Video Pinball | 318.311 | 355.299 | 405.444 | **1153.948** |
| Wizard of Wor | 4.518 | **8.567** | 5.936 | 8.513 |

Continued on next page

Table S15 – continued from previous page

| | PQN | Categorical Dueling | Ensemble Dueling | Aftab |
|---|---|---|---|---|
| Yars' Revenge | 2.239 | 5.541 | 5.552 | **5.659** |
| Zaxxon | 1.831 | 3.661 | 2.155 | **4.837** |
| Median | 1.894 | 4.261 | 3.636 | **4.374** |
| IQM | 2.715 | 6.093 | 5.625 | **6.592** |
| IQM 95% CI | [1.533, 4.486] | [3.418, 10.868] | [3.279, 11.988] | [3.524, 13.536] |

Table S16: Phase 3 Atari-57 Raw Returns: Final unnormalized evaluation scores detailing the Duelingcomparison of the Phase 3 value-estimation configurations (Categorical Dueling, Ensemble Dueling, and Aftab).

| | PQN | Categorical Dueling | Ensemble Dueling | Aftab |
|---|---|---|---|---|
| Alien | 4688.197 | 20386.327 | **24329.320** | 22953.912 |
| Amidar | 1082.500 | 2634.789 | **3378.616** | 2926.568 |
| Assault | 17009.493 | 15812.541 | **21412.942** | 20397.452 |
| Asterix | **366434.014** | 340450.680 | 330103.741 | 283998.639 |
| Asteroids | 2312.721 | **45602.007** | 43198.027 | 24836.667 |
| Atlantis | 742852.041 | 740343.197 | 773082.653 | **1922416.327** |
| Bank Heist | 1402.143 | **1673.741** | 1657.143 | 1576.259 |
| Battlezone | 44176.871 | **119812.925** | 107248.299 | 111278.912 |
| Beamrider | 22017.973 | **73008.435** | 53364.939 | 71070.449 |
| Berzerk | 5877.857 | 27839.252 | **46650.102** | 39211.531 |
| Bowling | 30.000 | 28.935 | 49.541 | **61.925** |
| Boxing | 99.228 | **100.000** | **100.000** | **100.000** |
| Breakout | 352.044 | 594.007 | 699.381 | **933.010** |
| Centipede | 10076.690 | 202759.650 | 35466.418 | **207815.929** |
| Chopper Command | 17155.442 | 543812.585 | **827556.122** | 568332.993 |
| Crazy Climber | 161068.367 | **216305.102** | 201359.524 | 173999.660 |
| Defender | 51787.245 | 297204.762 | 367071.259 | **505392.857** |
| Demon Attack | 132035.357 | 134858.639 | **135605.238** | 12853.605 |
| Double Dunk | -1.782 | **-1.245** | -1.340 | -21.027 |
| Enduro | 2360.721 | 2345.184 | 2346.575 | **6165.946** |
| Fishing Derby | 42.286 | 36.939 | **45.816** | 43.054 |
| Freeway | 33.823 | **34.000** | **34.000** | **34.000** |
| Frostbite | 6653.946 | 8145.272 | **12135.272** | 9659.218 |
| Gopher | 53851.973 | 44451.429 | **70625.034** | 40949.388 |
| Gravitar | 956.122 | 2012.585 | 2737.075 | **4410.034** |
| H.E.R.O. | 24341.207 | 51736.156 | 37095.714 | **68036.905** |
| Ice Hockey | 0.177 | 35.660 | 31.333 | **40.592** |
| James Bond | 2195.068 | 18633.333 | 7044.728 | **29409.354** |
| Kangaroo | 13455.442 | 14308.844 | 13969.728 | **14474.830** |
| Krull | 9707.585 | **13738.367** | 11224.558 | 13158.673 |
| Kung-Fu Master | **31990.136** | 29379.592 | 26209.184 | 29263.265 |
| Montezuma's Revenge | 0.000 | 0.000 | 123.129 | **148.980** |
| Ms. Pac-Man | 3365.340 | 11182.350 | **11885.439** | 9572.963 |
| Name This Game | 15329.014 | **26781.156** | 19973.878 | 26262.517 |
| Phoenix | 177903.333 | 482720.680 | 483661.565 | **1104620.680** |
| Pitfall! | -5.344 | **0.000** | -5.007 | **0.000** |
| Pong | **21.000** | **21.000** | **21.000** | **21.000** |
| Private Eye | 66.667 | 16.667 | **100.000** | **100.000** |
| Q*bert | 21923.469 | 45021.088 | **48135.969** | 44785.119 |
| River Raid | 21349.014 | 33202.007 | **43151.361** | 35383.163 |
| Road Runner | 57719.048 | 140571.429 | **375416.667** | 257419.048 |
| Robotank | 71.167 | 75.592 | **78.721** | 77.527 |
| Seaquest | 7978.027 | 471696.327 | 365552.687 | **1909299.898** |
| Skiing | -23644.677 | -15831.898 | **-8677.490** | -20213.990 |
| Solaris | **2605.306** | 1216.871 | 2487.279 | 1365.850 |
| Space Invaders | 5342.296 | 34645.204 | **62922.971** | 54110.289 |
| Stargunner | 259520.408 | 619437.075 | **692191.497** | 628127.211 |
| Surround | 8.350 | 9.827 | 9.500 | **9.959** |
| Tennis | -1.333 | 23.833 | **24.000** | **24.000** |
| Time Pilot | 13860.884 | 20004.762 | 19017.347 | **28396.259** |
| Tutankham | 249.473 | 240.306 | **266.622** | 258.588 |
| Up'n Down | 263719.048 | **311849.558** | 271042.823 | 269188.537 |

Continued on next page

Table S16 – continued from previous page

| | PQN | Categorical Dueling | Ensemble Dueling | Aftab |
|---|---|---|---|---|
| Venture | 0.000 | 0.000 | 0.000 | **707.483** |
| Video Pinball | 457305.670 | 478535.429 | 547276.806 | **1700656.636** |
| Wizard of Wor | 16897.959 | **36559.524** | 25731.633 | 36345.578 |
| Yars' Revenge | 115017.827 | 284535.259 | **305711.432** | 293344.007 |
| Zaxxon | 16136.735 | 33449.660 | 19704.422 | **39183.333** |

## S4. Detailed Statistical Analysis

This section provides detailed statistical results supporting the aggregate comparisons reported in the main manuscript. All Atari analyses operate on one human-normalized score (HNS) per method and game after averaging the four independent training seeds. Thus, the inferential unit is the Atari game, yielding 57 paired observations for each pairwise comparison. The four seeds are not treated as 228 independent observations.

Pairwise comparisons use two-sided Wilcoxon signed-rank tests. Multiple comparisons are corrected using the Bonferroni procedure within each predefined experimental family. Phase 1 contains nine configurations and therefore $\binom{9}{2} = 36$ pairwise comparisons. The four-configuration families reported for Phases 2 and 3 each contain $\binom{4}{2} = 6$ comparisons. Statistical significance is assessed at a family-wise level of $\alpha = 0.05$.

A non-significant pairwise comparison is not interpreted as evidence of statistical equivalence. In particular, when corrected testing does not resolve the ordering of leading configurations, the progressive selection procedure additionally considers aggregate performance and computational complexity.

### *S4.1. Probability of Improvement*

In addition to the Wilcoxon analysis, we report a descriptive game-level Probability of Improvement (PoI). For methods $X$ and $Y$,

$$\mathrm{PoI}(X, Y) = \frac{1}{57} \sum_{e=1}^{57} \left[ \mathbb{I}\left(\mathrm{HNS}_{X,e} > \mathrm{HNS}_{Y,e}\right) + \frac{1}{2} \mathbb{I}\left(\mathrm{HNS}_{X,e} = \mathrm{HNS}_{Y,e}\right) \right]. \quad \text{(S42)}$$

Each game on which $X$ has the larger four-seed mean contributes one, a loss contributes zero, and a tie contributes one half. Thus, $\mathrm{PoI} = 0.5$ indicates balanced game-level performance.

This quantity is an empirical proportion over the evaluated Atari-57 task suite. It should not be interpreted as a Bayesian posterior probability, an estimate of posterior sampling behavior, or a probability over hypothetical independent training runs.

### *S4.2. IQM Bootstrap Confidence Intervals*

The 95% confidence intervals reported for interquartile mean (IQM) HNS are obtained by bootstrap resampling across the game-level HNS values and recomputing the IQM for each bootstrap sample. Because the input to this procedure consists of seed-averaged game-level values, the intervals primarily characterize variability across tasks around the four-seed estimates; they do not fully quantify between-training-run uncertainty.

### *S4.3. Phase 1: Encoder Comparison*

Phase 1 compares the PQN reference with the eight alternative convolutional encoders Alpha through Theta. Selected contrasts that determine the interpretation and backbone-selection decision are summarized in Table S17.

Table S17: **Selected Phase 1 pairwise statistics.** Adjusted $p$-values are from two-sided Wilcoxon signed-rank tests with Bonferroni correction over the complete 36-comparison Phase 1 family. PoI is computed from the 57 paired game-level HNS values.

| **Method** $X$ | **Method** $Y$ | $p_{\text{adj}}$ | $\mathbf{PoI}(X, Y)$ |
|---|---|---|---|
| Alpha | PQN | $< 0.001$ | $\approx 0.85$ |
| Gamma | PQN | $< 0.001$ | $\approx 0.79$ |
| Alpha | Gamma | 1.000 | $\approx 0.51$ |

Alpha has the largest Phase 1 IQM HNS point estimate, while Gamma follows closely. Their corrected paired comparison is not resolved and their PoI is approximately balanced. This is not interpreted as statistical equivalence. Instead, Gamma is selected under the study's performance–complexity decision rule because it achieves closely matched aggregate performance with lower encoder parameter count and convolutional cost.

Figure S4 shows the complete Phase 1 PoI matrix.

#### *S4.3.1. Phase 1 Task-Composition Sensitivity*

As an additional sensitivity analysis, the Phase 1 performance–complexity selection procedure is repeated on 10 randomly formed subsets of 15 Atari games

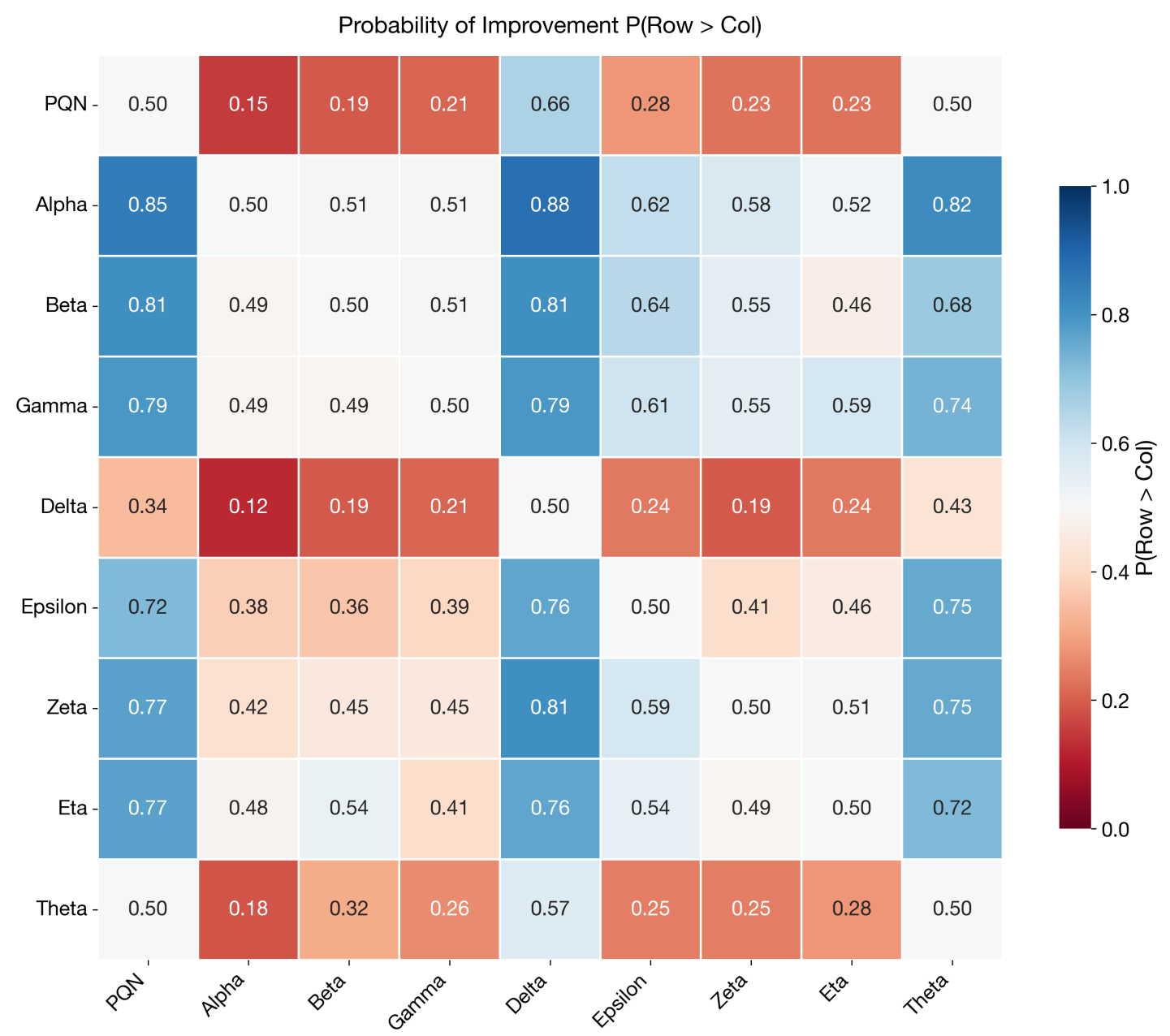


Figure S4: **Complete Phase 1 Probability of Improvement matrix.** Each entry reports PoI(row, column) across the 57 Atari games, with ties contributing one half. Values above 0.5 indicate that the row configuration has the larger four-seed mean on a greater proportion of games.

using the already completed training runs. Gamma is selected in all 10 subsets.

Because no additional agents are trained and the subsets reuse observations from the same Atari-57 experiment, this analysis does not constitute an independent validation set. It instead assesses whether the selection of Gamma is highly sensitive to the particular task composition used by the selection procedure.

### *S4.4. Phase 2: Hadamax Comparison*

Phase 2 compares scalar Gamma, the independently implemented three-stage Hadamax reference, Gamma-Hadamax-Valid, and Gamma-Hadamax-Same. The six pairwise comparisons form one Bonferroni family.

Table S18 reports the pairwise results used in the interpretation and selection decision.

Table S18: **Selected Phase 2 pairwise statistics.** Adjusted $p$-values are from two-sided Wilcoxon signed-rank tests with Bonferroni correction over the six-comparison Phase 2 family.

| **Method** $X$ | **Method** $Y$ | $p_{\text{adj}}$ | $\textbf{PoI}(X, Y)$ |
|---|---|---|---|
| Gamma-Hadamax-Valid | Gamma | 0.005 | $\approx 0.70$ |
| Gamma-Hadamax-Same | Gamma | $< 0.001$ | $\approx 0.75$ |
| Gamma-Hadamax-Valid | Gamma-Hadamax-Same | 1.000 | $\approx 0.50$ |
| Gamma-Hadamax-Valid | Hadamax Reference | 0.226 | $\approx 0.65$ |
| Gamma-Hadamax-Same | Hadamax Reference | 0.056 | $\approx 0.62$ |
| Hadamax Reference | Gamma | 0.145 | – |

Both Gamma-Hadamax variants differ from scalar Gamma after correction. However, their direct comparison is not resolved, and neither five-stage variant is statistically resolved from the independently implemented three-stage Hadamax reference after correction. The statistical analysis therefore supports an improvement over scalar Gamma but does not establish superiority of either five-stage Gamma-Hadamax variant over the Hadamax reference.

The PoI for Gamma-Hadamax-Valid versus Gamma-Hadamax-Same is approximately balanced. Combined with the substantially smaller flattened representation of Gamma-Hadamax-Valid, this motivates carrying Valid forward under the performance–complexity decision rule.

Figure S5 shows the complete Phase 2 PoI matrix.

### *S4.5. Phase 3: Value-Estimation Comparison*

The currently reported Phase 3 corrected statistical family contains PQN, Categorical Dueling, Ensemble Dueling, and Aftab. This detail is important because scalar Gamma-Hadamax-Valid, although it is the representation-level reference carried forward from Phase 2, is not a member of the currently available four-configuration corrected matrix.

Table S19 therefore reproduces only the inferential claims supported by the currently reported comparison family.

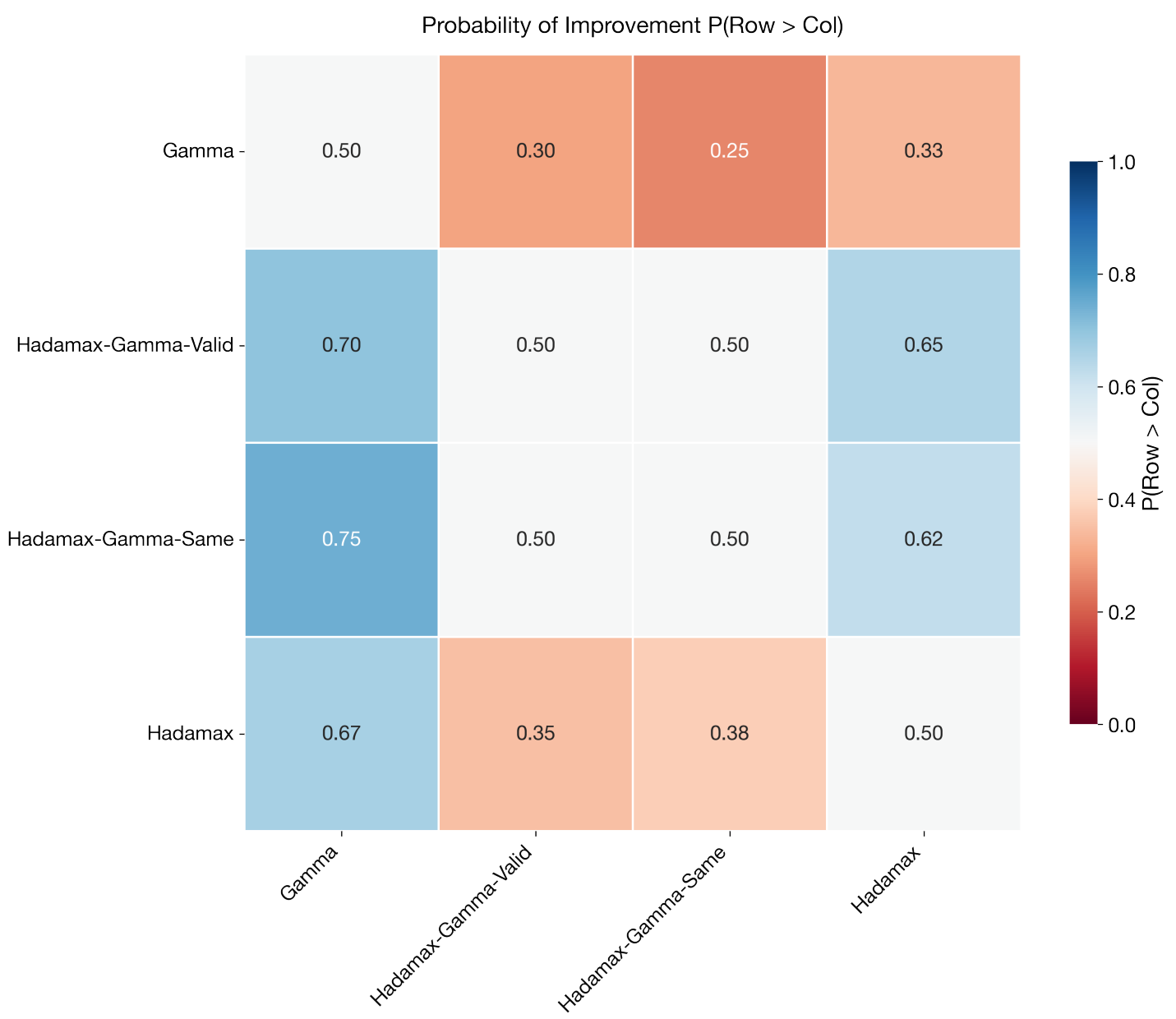


Figure S5: **Complete Phase 2 Probability of Improvement matrix.** Pairwise game-level comparison of Gamma, the Hadamax reference, Gamma-Hadamax-Valid, and Gamma-Hadamax-Same across Atari-57.

Table S19: **Selected statistics from the currently reported Phase 3 comparison family.** The Bonferroni family consists of PQN, Categorical Dueling, Ensemble Dueling, and Aftab, yielding six pairwise comparisons.

| **Method** $X$ | **Method** $Y$ | $p_{\text{adj}}$ | $\textbf{PoI}(X, Y)$ |
|---|---|---|---|
| Categorical Dueling | PQN | $< 0.001$ | $\approx 0.82$ |
| Ensemble Dueling | PQN | $< 0.001$ | $\approx 0.93$ |
| Aftab | PQN | $< 0.001$ | 0.86 |
| Categorical Dueling | Ensemble Dueling | 0.971 | – |
| Aftab | Categorical Dueling | 0.007 | $\approx 0.68$ |
| Aftab | Ensemble Dueling | 1.000 | $\approx 0.51$ |

Within this family, all three modified value-estimation configurations differ from PQN after correction. Aftab differs from Categorical Dueling, whereas the

Aftab–Ensemble Dueling comparison is not statistically resolved.

The comparison between scalar Gamma-Hadamax-Valid and the three modified Phase 3 heads is therefore currently descriptive. The higher IQM point estimates of Categorical Dueling, Ensemble Dueling, and Aftab relative to scalar Gamma-Hadamax-Valid should not be presented as corrected inferential differences unless a statistical family containing those four configurations is computed directly.

Figure S6 shows the complete PoI matrix for the currently reported Phase 3 family.

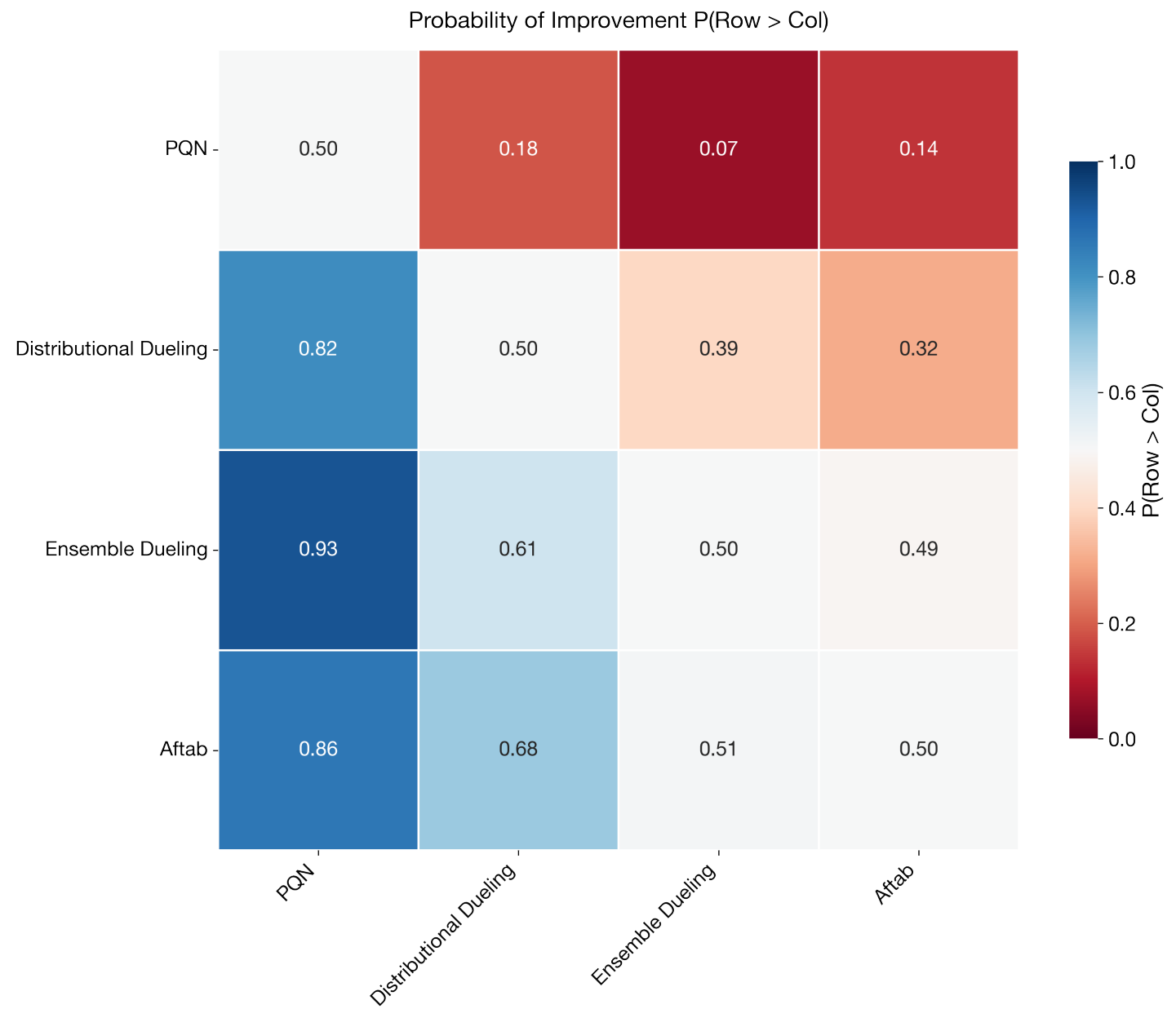


Figure S6: **Complete Probability of Improvement matrix for the currently reported Phase 3 family.** The matrix contains PQN, Categorical Dueling, Ensemble Dueling, and Aftab. Scalar Gamma-Hadamax-Valid is not part of this matrix.

### *S4.6. Availability of Full Statistical Outputs*

The released experimental artifacts include the game-level score tables and statistical-analysis outputs used to construct the reported comparisons. For

reproducibility, the released experimental artifacts provide machine-readable access to the following outputs for each experimental phase:

- the 57 paired game-level HNS values used in each comparison;
- the raw two-sided Wilcoxon signed-rank $p$-value matrix;
- the Bonferroni-adjusted $p$-value matrix;
- the complete Probability of Improvement matrix; and
- the analysis configuration required to reproduce the reported bootstrap confidence intervals.

The released statistical-analysis outputs contain the numerical matrices used for the reported comparisons and should be used as the machine-readable source for exact values beyond those summarized in the manuscript.

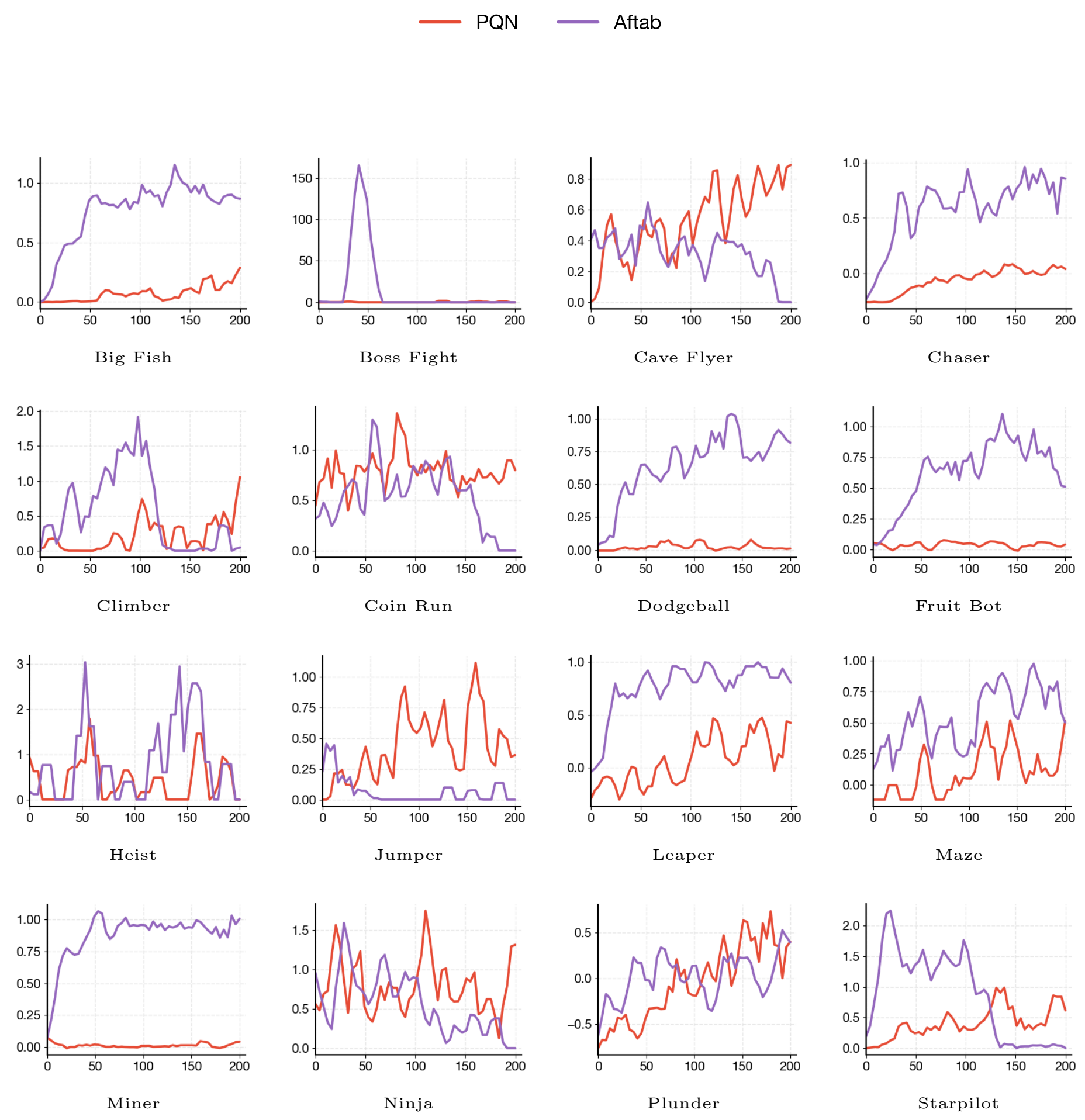


Figure S7: Procgen Hard Procedural Evaluation: unnormalized evaluation returns over 200 million training frames comparing the baseline PQN to the complete Aftab framework across the 16 individual Procgen Hard environments.

Table S20: Procgen Hard Raw Returns: Unnormalized evaluation returns over 200 million training frames comparing PQN and Aftab across the 16 Procgen Hard environments.

| | PQN | Aftab |
|---|---|---|
| Big Fish | 7.233 | **31.491** |
| Boss Fight | **0.023** | 0.000 |
| Cave Flyer | **5.313** | 0.000 |
| Chaser | 2.564 | **8.871** |
| Climber | **1.049** | 0.000 |
| Coin Run | **4.816** | 0.000 |
| Dodgeball | 0.192 | **12.596** |
| Fruit Bot | -0.415 | **13.647** |
| Heist | **0.418** | 0.347 |
| Jumper | **2.306** | 0.000 |
| Leaper | 5.500 | **9.520** |
| Maze | 1.449 | **4.561** |
| Miner | 0.358 | **10.052** |
| Ninja | **2.898** | 0.520 |
| Plunder | **4.983** | 4.672 |
| Starpilot | **4.508** | 0.223 |

Table S21: **Procgen Hard raw-score Area Under the Curve (AUC) by environment**: For each Procgen Hard environment, the AUC is computed over the complete 200-million-frame training trajectory using trapezoidal integration across the recorded evaluation checkpoints and then averaged across the four experimental seeds. Bold values indicate the higher AUC within each environment. Because raw reward scales differ substantially across Procgen tasks, these values are intended only for within-environment comparison between Aftab and PQN and are not aggregated directly across environments.

| Environment | Aftab | PQN |
|---|---:|---:|
| Big Fish | **148.354** | 13.218 |
| Boss Fight | **2961.430** | 298.998 |
| Cave Flyer | 70.669 | **113.820** |
| Chaser | **119.001** | -10.464 |
| Climber | **128.451** | 42.228 |
| Coin Run | 111.235 | **159.278** |
| Dodgeball | **131.911** | 4.117 |
| Fruit Bot | **130.779** | 10.437 |
| Heist | **96.673** | 68.645 |
| Jumper | 19.699 | **93.572** |
| Leaper | **148.856** | 14.215 |
| Maze | **97.860** | 20.906 |
| Miner | **175.375** | 1.638 |
| Ninja | 107.727 | **162.305** |
| Plunder | **19.860** | 3.061 |
| Star Pilot | **180.687** | 93.280 |

## S5. Reproducibility Details

This section provides the implementation and training details required to reproduce the experiments reported in the main manuscript. Unless explicitly stated otherwise, competing configurations within a benchmark use the same environment-interaction budget, optimization settings, and random-seed set. Architectural changes are therefore evaluated within a common training protocol rather than through configuration-specific hyperparameter tuning.

The implementation, experiment configurations, processed results, and experimental outputs are publicly available in the project repository:

`https://github.com/tahashieenavaz/aftab`

An archived snapshot associated with this study is permanently available through Zenodo:

`https://zenodo.org/records/22812154`

A mirror of the development repository is additionally maintained at

`https://github.com/lorisnanni/aftab`.

### *S5.1. Random Seeds and Paired Experimental Design*

All reported experiments use the same four fixed training seeds:

$$\{475284,\ 219842,\ 525975,\ 909314\}. \tag{S43}$$

The same seed set is reused across competing configurations. This preserves the paired structure of the game-level comparisons and avoids introducing different seed sets as an additional source of variation between methods.

For a training seed $s$, the corresponding test environments use seed

$$s_{\text{test}} = s + 1000. \tag{S44}$$

Eight test environments are maintained concurrently with training for both Atari-57 and Procgen Hard. Test-environment transitions are not inserted into

the training rollout batch and do not contribute to the training interaction budget.

*S5.2. Common Optimization Settings*

Table S22 summarizes the optimization settings shared across the reported experiments.

Table S22: **Common optimization settings.** Settings are held fixed across the relevant architectural comparisons unless the architecture itself requires a different output objective.

| **Setting** | **Value** |
|---|---|
| Optimizer | RAdam |
| Learning rate | $2.5 \times 10^{-4}$ |
| Optimizer $\epsilon$ | $1 \times 10^{-5}$ |
| $\beta_1$ | 0.9 |
| $\beta_2$ | 0.999 |
| Weight decay | 0 |
| Discount factor $\gamma$ | 0.99 |
| $\lambda$-return parameter | 0.65 |
| Optimization epochs per rollout | 2 |
| Gradient-norm clipping | 10.0 |
| Exploration schedule | Linear $\epsilon$-greedy |
| Initial $\epsilon$ | 1.0 |
| Final $\epsilon$ | 0.001 |
| Exploration annealing period | First 10% of training |
| Downstream embedding dimension | 512 |

The exploration probability decreases linearly from 1.0 to 0.001 during the first 10% of the training horizon and remains at 0.001 thereafter. Evaluation actions do not use $\epsilon$-greedy exploration.

Scalar PQN configurations, all scalar Phase 1 and Phase 2 configurations, and Ensemble Dueling minimize the half mean-squared error between the pre-

dicted action value and the bootstrapped $\lambda$-return. Categorical Dueling and Aftab instead transform the scalar $\lambda$-return into an HL-Gauss target and optimize cross-entropy.

The experiments follow the replay-free PQN training regime: no historical experience-replay buffer and no separate target network are used. The newly collected synchronous rollout is retained temporarily for $\lambda$-return construction and repeated mini-batch optimization and is then replaced by the next rollout.

*S5.3. Atari-57 Environment Protocol*

The primary design study uses all 57 Atari games with the corresponding EnvPool Atari `-v5` environments. Observations consist of four stacked grayscale frames resized to $84 \times 84$, producing input tensors of shape

$$4 \times 84 \times 84. \tag{S45}$$

Environment observations are represented as unsigned 8-bit image values and are divided by 255 inside the network before visual encoding.

The Atari-specific environment settings are summarized in Table S23.

Each rollout contains

$$128 \times 32 = 4096 \tag{S46}$$

training transitions. The rollout is divided into 32 mini-batches of 128 transitions and optimized for two epochs before the next rollout is collected.

With a frame skip of four, the nominal 200-million-observed-frame budget corresponds to 50 million training-environment action transitions. Because complete rollout batches are always collected, each run performs 12,208 updates and contains

$$12,208 \times 4096 = 50,003,968 \tag{S47}$$

training transitions, corresponding to

Table S23: **Atari-57 environment and rollout configuration.**

| **Setting** | **Value** |
|---|---|
| Observation shape | $4 \times 84 \times 84$ |
| Parallel training environments | 128 |
| Parallel test environments | 8 |
| Rollout steps per environment | 32 |
| Transitions per rollout | 4096 |
| Mini-batches per epoch | 32 |
| Mini-batch size | 128 |
| Optimization epochs | 2 |
| Frame skip | 4 |
| Frame stack | 4 |
| Maximum initial no-op actions | 30 |
| Action set | Minimal Atari action set |
| Repeat-action probability | 0 |
| Training episodic life | Enabled |
| Test episodic life | Disabled |
| Training reward clipping | Enabled |
| Test reward clipping | Enabled |
| Nominal training budget | 200M observed ALE frames |

$$200,015,872 \tag{S48}$$

observed ALE frames. This quantity is referred to throughout the manuscript as the 200-million-frame training budget.

*S5.3.1. Atari Reward Handling and Score Reporting*

Reward clipping is enabled in the Atari environments, but the reward used for learning is distinct from the score used for reporting. EnvPool clips the reward returned through the environment transition to its sign:

$$r_t^{\mathrm{clip}} = \begin{cases} 1, & r_t > 0, \\ 0, & r_t = 0, \\ -1, & r_t < 0. \end{cases} \tag{S49}$$

This clipped reward is stored in the current rollout and is therefore used in construction of the bootstrapped $\lambda$-return.

The original unclipped Atari reward remains available through the environment information returned by EnvPool. The reporting code accumulates this unclipped quantity. Consequently, all raw Atari scores reported in the study, and all Human-Normalized Scores derived from them, remain on the original ALE game-score scale rather than the clipped reward scale.

The implementation similarly distinguishes training episodic-life transitions from complete-game score reporting. During training, loss of a life can terminate an episodic-life transition. Reported game scores, however, continue to accumulate until the underlying Atari environment signals a true game termination. Life-loss pseudo-episodes are therefore not counted as separate reported episodes.

*S5.3.2. Atari Evaluation*

Eight Atari test environments are stepped concurrently with training. Their interactions are used only for evaluation and are excluded from the training rollout and interaction budget.

Single-head configurations act greedily during evaluation. Multi-head configurations use the head-level majority-vote procedure described in Section S5.5.

*S5.4. Procgen Hard Protocol*

Procgen Hard is used only after completion of architecture selection on Atari-57. It therefore serves as a secondary evaluation of the final Aftab configuration against the independently rerun PQN reference and does not contribute to selection of the architecture.

All 16 Procgen Hard environments are evaluated. Procgen observations retain their native RGB representation with shape

$$3 \times 64 \times 64. \tag{S50}$$

Image values are normalized by dividing by 255 before entering the visual encoder.

Table S24 gives the Procgen training protocol.

Table S24: **Procgen Hard environment and rollout configuration.**

| **Setting** | **Value** |
|---|---|
| Observation shape | $3 \times 64 \times 64$ RGB |
| Parallel training environments | 64 |
| Parallel test environments | 8 |
| Rollout steps per environment | 256 |
| Transitions per rollout | $16,384$ |
| Mini-batches per epoch | 32 |
| Mini-batch size | 512 |
| Optimization epochs | 2 |
| Nominal training budget | 200M interactions |

Each rollout therefore contains

$$64 \times 256 = 16,384 \tag{S51}$$

training transitions. These are divided into 32 mini-batches of 512 transitions and optimized for two epochs.

Procgen does not use Atari-style frame skipping in the reported experiments, so one training transition corresponds to one environment interaction. Because the implementation collects complete rollout batches, each run contains 12,208 updates and

$$12,208 \times 16,384 = 200,015,872 \tag{S52}$$

training interactions.

The environment-construction code queries the options supported by each EnvPool environment. Atari-specific options such as frame skipping, frame stacking, no-op initialization, reward clipping, and episodic-life termination are therefore not blindly transferred to Procgen environments.

### *S5.5. Phase 3 Value-Estimation Settings*

Phase 3 fixes the Gamma-Hadamax-Valid representation and evaluates complete downstream value-estimation configurations. The settings specific to the categorical and ensemble models are summarized in Table S25.

Table S25: **Phase 3 categorical and ensemble settings.**

| **Setting** | **Value** |
|---|---|
| Categorical support bins | 51 |
| Categorical support | $[-10, 10]$ |
| HL-Gauss sigma ratio | 0.75 |
| Derived $\sigma$ | $\approx 0.2941$ |
| Range clamping | Enabled |
| Distributional value-change clip | 0.0 |
| Ensemble heads $K$ | 10 |
| Bootstrap inclusion probability $p$ | 1.0 |

For the 51-bin support over $[-10, 10]$, the default HL-Gauss width used by the implementation is

$$\sigma = 0.75 \left( \frac{10 - (-10)}{51} \right) \approx 0.2941. \tag{S53}$$

The HL-Gauss implementation uses range clamping for scalar targets outside the fixed support. The optional value-change clipping mechanism is disabled (`distributional_value_clip=0`).

This objective is categorical regression of a scalar bootstrapped $\lambda$-return. It is not the categorical Bellman-distribution projection used by C51.

### *S5.5.1. Ensemble Training and Evaluation*

Ensemble Dueling and Aftab use $K = 10$ independently parameterized downstream heads on top of the shared visual representation.

At the beginning of training, one head is sampled uniformly for each parallel training environment. Actions in that environment are selected $\epsilon$-greedily with respect to the active head. The same head remains active until the environment produces a termination or truncation, after which a new head is sampled uniformly.

The bootstrap inclusion probability is

$$p = 1. \tag{S54}$$

Consequently, every head receives every collected transition. The method therefore does not create classical bootstrap diversity through distinct resampled training datasets. Each head nevertheless computes its own head-specific bootstrapped $\lambda$-return using its own next-state value estimate.

For Aftab, each head independently transforms its scalar bootstrapped return into the corresponding HL-Gauss categorical target.

During evaluation, Q-values are not averaged across ensemble heads. Instead, each head independently selects its greedy action. The $K$ resulting actions are treated as votes, and the action receiving the largest number of votes is executed.

### *S5.6. Phase 3 Activation Functions*

All Phase 3 downstream streams use a hidden embedding dimension of 512 and Layer Normalization after the first linear projection. The implemented activation differs across configurations:

- Categorical Dueling uses ReLU;

- Ensemble Dueling uses ReLU; and
- Aftab uses GELU.

Phase 3 should therefore be interpreted as a comparison between complete implemented value-estimation architectures rather than as an isolated factorial ablation of categorical regression, dueling decomposition, ensembling, or activation function.

*S5.7. Software Environment*

The software environment recorded for the study includes the following versions:

Table S26: **Principal software versions used by the experimental pipeline.**

| **Package** | **Version** |
|---|---|
| EnvPool | 1.2.0 |
| Gymnasium | 1.3.0 |
| PyTorch | 2.11.0 |
| hl-gauss-pytorch | 0.2.3 |

On CUDA devices, the training pipeline uses float16 automatic mixed precision with gradient scaling. Network tensors use channels-last memory format. Models are compiled using

```
torch.compile(mode="max-autotune")
```

with static shapes. Float32 matrix-multiplication precision is set to `"high"`.

*S5.8. Hardware*

The reported experiments were run on NVIDIA A40 GPUs with 48 GB of GPU memory. The most computationally demanding configuration required approximately 13 hours of wall-clock training time on this hardware.

The reported parameter counts and multiply-accumulate counts are architectural complexity measures and should not be interpreted as direct hardware-runtime measurements. Actual wall-clock performance can vary with software versions, compilation, mixed-precision behavior, device utilization, and hardware.

*S5.9. Reference Implementations*

The PQN and Hadamax reference configurations included in the experimental comparisons are independently executed within the same PyTorch experimental pipeline as the proposed configurations and use the same four study seeds.

Published numerical results from the original PQN and Hadamax studies are not combined with these runs as experimental samples and are not included in the pairwise statistical tests. Published results are used only as external reproduction references.

The Hadamax reference in this work is an independent PyTorch reimplementation rather than an exact execution of the original JAX code. The present Atari pipeline uses $4 \times 84 \times 84$ observations, whereas the published Hadamax implementation uses a $4 \times 64 \times 64$ input. The PyTorch implementation also uses explicit integer padding rather than the JAX-style `SAME` padding of the original implementation. These differences should be retained when interpreting numerical reproduction comparisons.

*S5.10. Released Reproducibility Artifacts*

The public repository provides the experimental material used to generate the reported results, including:

- model and network definitions;
- environment and optimizer configurations;
- the fixed random-seed list;

- processed Atari and Procgen score tables;
- learning-curve outputs;
- Wilcoxon signed-rank test matrices;
- Bonferroni-corrected statistical matrices;
- Probability of Improvement matrices; and
- Procgen environment-level AUC tables.

The version of the code and associated research artifacts archived for this study is permanently available through Zenodo at `https://zenodo.org/records/22812154`. The GitHub repositories remain available for development and access to the current implementation.

## References


[1] J. E. Kooi, Z. Yang, V. François-Lavet, Hadamax encoding: Elevating performance in model-free atari, arXiv preprint arXiv:2505.15345 (2025). URL: https://arxiv.org/abs/2505.15345. doi:10.48550/arXiv.2505.15345.

[2] Z. Wang, T. Schaul, M. Hessel, H. Hasselt, M. Lanctot, N. Freitas, Dueling network architectures for deep reinforcement learning, in: International Conference on Machine Learning (ICML), 2016, pp. 1995–2003.

[3] I. Osband, C. Blundell, A. Pritzel, B. Van Roy, Deep exploration via bootstrapped dqn, arXiv preprint arXiv:1602.04621 (2016).


\end{document}